\pdfoutput=1 
\documentclass[english,11pt]{article}
\usepackage[T1]{fontenc}
\usepackage[latin9]{inputenc}
\usepackage{geometry}
\usepackage{babel}
\usepackage{verbatim}
\usepackage{mathtools}
\usepackage{bm}
\usepackage{amsmath}
\usepackage{amssymb}
\usepackage[unicode=true,
 bookmarks=false,
 breaklinks=false,
 pdfborder={0 0 1},
 backref=section,
 colorlinks=false]{hyperref}

\makeatletter

\newcommand{\upstairs}[1]{\textsuperscript{#1}}
\newcommand{\affilone}{\dag}
\newcommand{\affiltwo}{\ddag}

\usepackage{bm}
\usepackage{enumitem}

\usepackage{babel}
\usepackage{subcaption}

\usepackage{algorithmic}

\usepackage{amsthm}
\usepackage{bbm}

\theoremstyle{plain}

\newtheorem{lemma}{\textbf{Lemma}}
\newtheorem{theorem}{\textbf{Theorem}}

\newtheorem{proposition}{\textbf{Proposition}}

\theoremstyle{definition}

\usepackage{color}
\definecolor{cm}{RGB}{0,0,200}
\definecolor{purple}{RGB}{200,0,200}

\makeatother

\newcommand{\Exp}{\ensuremath{\mathbb{E}}}

\newcommand{\Prob}{\ensuremath{\mathbb{P}}}

\newcommand{\numobs}{\ensuremath{n}}

\newcommand{\zeros}{\ensuremath{\bm{0}}}

\newcommand{\indicator}{\ensuremath{\mathbbm{1}}}

\newcommand{\ccon}{\ensuremath{c}}
\newcommand{\plaincon}{\ensuremath{c}}
\newcommand{\cprime}{\ensuremath{\ccon'}}

\newcommand{\Uvar}{\ensuremath{U}}
\newcommand{\Vvar}{\ensuremath{V}}
\newcommand{\Xvar}{\ensuremath{X}}
\newcommand{\xvar}{\ensuremath{x}}

\newcommand{\entropy}{\ensuremath{h}}

\newcommand{\ScaledCheby}{\ensuremath{P}}
\newcommand{\coeff}{\ensuremath{a}}

\newcommand{\Poi}{\ensuremath{\mathsf{Poi}}}

\newcommand{\Vhat}{\ensuremath{\widehat{V}}}
\newcommand{\VhatPlug}{\ensuremath{\Vhat_{\mathsf{plug}}}}
\newcommand{\VhatIS}{\ensuremath{\Vhat_{\mathsf{IS}}}}
\newcommand{\VhatSwitch}{\ensuremath{\Vhat_{\mathsf{switch}}}}
\newcommand{\VhatCheby}{\ensuremath{\Vhat_{\mathsf{C}}}}

\newcommand{\Vphi}{\ensuremath{V_{\RewardDistPlain}}}

\newcommand{\Risk}{\ensuremath{\mathcal{R}}}

\newcommand{\Var}{\ensuremath{\mathsf{Var}}}
\newcommand{\Bias}{\ensuremath{\mathsf{Bias}}}

\newcommand{\Rmax}{\ensuremath{r_{\scriptscriptstyle \max}}}
\newcommand{\pibehave}{\ensuremath{\pi_{\mathsf{b}}}}

\newcommand{\pitarget}{\ensuremath{\pi_{\mathsf{t}}}}
\newcommand{\bpitarget}{\ensuremath{\bm{\pi}_{\mathsf{t}}}}

\newcommand{\numaction}{\ensuremath{k}}

\newcommand{\Action}{\ensuremath{A}}
\newcommand{\action}{\ensuremath{a}}
\newcommand{\ActionSpace}{\ensuremath{[\numaction]}}
\newcommand{\GeneralActionSpace}{\ensuremath{\mathcal{A}}}

\newcommand{\ActionSubset}{\ensuremath{S}}

\newcommand{\Reward}{\ensuremath{R}}
\newcommand{\reward}{\ensuremath{r_{\RewardDistPl}}}
\newcommand{\rhat}{\ensuremath{\widehat{r}}}

\newcommand{\RewardDistPlain}{\ensuremath{f}}
\newcommand{\RewardDistPl}{\RewardDistPlain}
\newcommand{\RewardDist}[1]{\ensuremath{\RewardDistPlain(\,\cdot \mid
    #1)}}

\newcommand{\RewardFamily}{\ensuremath{\mathcal{F}}}

\newcommand{\real}{\ensuremath{\mathbb{R}}}

\newcommand{\mysupp}{\ensuremath{\operatorname{supp}}}

\newcommand{\Set}{\ensuremath{S}}
\newcommand{\SetStar}{\ensuremath{\Set^{\star}}}
\newcommand{\SetStarComp}{\ensuremath{(\SetStar)^c}}

\newcommand{\Target}{\ensuremath{T}}
\newcommand{\TargetHat}{\ensuremath{\widehat{T}}}

\newcommand{\LocalPara}{\ensuremath{\theta}}
\newcommand{\LocalParaSpace}{\ensuremath{\Theta}}
\newcommand{\LocalObs}{\ensuremath{\bm{Z}}}

\newcommand{\Prior}[1]{\ensuremath{\Xi_{#1}}}
\newcommand{\LocalMarginal}[1]{\ensuremath{F_{#1}}}

\newcommand{\likerat}{\ensuremath{\rho}}

\newcommand{\plower}{\ensuremath{\nu}}

\newcommand{\vb}{\ensuremath{\bm{v}}}

\newcommand{\thetabm}{\ensuremath{\bm{\theta}}}

\newcommand{\TVdist}{\ensuremath{\mathsf{TV}}}

\newcommand{\PiFamily}{\ensuremath{\Pi}}
\newcommand{\PiFam}{\PiFamily}

\newcommand{\Kull}[2]{\ensuremath{\mathsf{D}_{\scriptscriptstyle{\operatorname{KL}}}(#1 \, |\!| \, #2)}}

\newcommand{\Term}{\ensuremath{T}}

\newcommand{\ChebPoly}{\ensuremath{Q_{\cdeg}}}

\newcommand{\cdeg}{\ensuremath{L}}

\newcommand{\myleft}{\ensuremath{\ell}}
\newcommand{\myright}{\ensuremath{r}}

\newcommand{\pstar}{\ensuremath{p^*}}

\newcommand{\CompRat}{\ensuremath{\mathcal{C}}}

\newcommand{\delpar}{\ensuremath{\delta}}

\newcommand{\LagFun}{\ensuremath{\mathcal{L}}}

\newcommand{\betapar}{\ensuremath{\beta}}

\newcommand{\PosCount}{\ensuremath{P}}
\newcommand{\NegCount}{\ensuremath{N}}

\newcommand{\plowerPos}{\ensuremath{\plower_{\mathsf{p}}}}
\newcommand{\plowerNeg}{\ensuremath{\plower_{\mathsf{n}}}}

\newcommand{\HelperPrior}[1]{\ensuremath{\Gamma}_{#1}}
\newcommand{\LocalSet}{\ensuremath{E}}
\newcommand{\TermAlpha}{\ensuremath{\alpha}}

\newcommand{\numobsnew}{\ensuremath{m}}
\newcommand{\BayesPrior}{\ensuremath{\Omega}}

\newcommand{\ratio}{\ensuremath{\delta}}
\newcommand{\LocalBeta}{\ensuremath{\beta}}
\newcommand{\LocalAlpha}{\ensuremath{\alpha}}
\newcommand{\LocalLambda}{\ensuremath{\lambda}}

\newcommand{\PolyApproxError}{\ensuremath{E}}
\newcommand{\LocalDist}{\ensuremath{\mu}}

\newcommand{\decaycon}{\ensuremath{\gamma}}

\newcommand{\Event}{\ensuremath{\mathcal{E}}}

\newcommand{\slope}{\ensuremath{\widehat{\beta}}}

\newenvironment{carlist}
 {\begin{list}{$\bullet$}
 {\setlength{\topsep}{0in} \setlength{\partopsep}{0in}
  \setlength{\parsep}{0in} \setlength{\itemsep}{\parskip}
  \setlength{\leftmargin}{0.07in} \setlength{\rightmargin}{0.08in}
  \setlength{\listparindent}{0in} \setlength{\labelwidth}{0.08in}
  \setlength{\labelsep}{0.1in} \setlength{\itemindent}{0in}}}
 {\end{list}}

\newcommand{\bcar}{\begin{carlist}}
\newcommand{\ecar}{\end{carlist}}

\makeatletter
\long\def\@makecaption#1#2{
        \vskip 0.8ex
        \setbox\@tempboxa\hbox{\small {\bf #1:} #2}
        \parindent 1.5em  
        \dimen0=\hsize
        \advance\dimen0 by -3em
        \ifdim \wd\@tempboxa >\dimen0
                \hbox to \hsize{
                        \parindent 0em
                        \hfil 
                        \parbox{\dimen0}{\def\baselinestretch{0.96}\small
                                {\bf #1.} #2
                                } 
                        \hfil}
        \else \hbox to \hsize{\hfil \box\@tempboxa \hfil}
        \fi
        }
\makeatother

\begin{document}


\begin{center}

  {\bf{\Large Minimax Off-Policy Evaluation for Multi-Armed Bandits}} \\

  \vspace*{.2in}
  
  \begin{tabular}{cccc}
    Cong Ma\upstairs{\affilone, \affiltwo}\quad

    Banghua Zhu\upstairs{\affilone}\quad

    Jiantao Jiao\upstairs{\affilone, \affiltwo}\quad
    
    Martin J.\ Wainwright\upstairs{\affilone, \affiltwo} 
    \vspace*{.1in} \\

    \upstairs{\affilone} Department of Electrical Engineering and Computer Sciences, UC Berkeley \\
    \upstairs{\affiltwo} Department of Statistics, UC Berkeley
  \end{tabular}
  
  \vspace*{.2in}

  \begin{abstract}
    We study the problem of off-policy evaluation in the multi-armed
    bandit model with bounded rewards, and develop minimax
    rate-optimal procedures under three settings. First, when the
    behavior policy is known, we show that the Switch estimator, a
    method that alternates between the plug-in and importance sampling
    estimators, is minimax rate-optimal for all sample sizes. Second,
    when the behavior policy is unknown, we analyze performance in
    terms of the competitive ratio, thereby revealing a fundamental
    gap between the settings of known and unknown behavior policies.
    When the behavior policy is unknown, any estimator must have
    mean-squared error larger---relative to the oracle estimator
    equipped with the knowledge of the behavior policy--- by a
    multiplicative factor proportional to the support size of the
    target policy.  Moreover, we demonstrate that the plug-in approach
    achieves this worst-case competitive ratio up to a logarithmic
    factor. Third, we initiate the study of the partial knowledge
    setting in which it is assumed that the minimum probability taken
    by the behavior policy is known. We show that the plug-in
    estimator is optimal for relatively large values of the minimum
    probability, but is sub-optimal when the minimum probability is
    low. In order to remedy this gap, we propose a new estimator based
    on approximation by Chebyshev polynomials that provably achieves
    the optimal estimation error.  Numerical
    experiments on both simulated and real data corroborate our
    theoretical findings.
  \end{abstract}
\end{center}

\section{Introduction}
\label{sec:intro}

Various forms of sequential decision-making, including multi-armed
bandits~\cite{lattimore2020bandit}, contextual
bandits~\cite{auer2002nonstochastic,tewari2017ads}, and Markov
decision
processes~\cite{sutton2018reinforcement,szepesvari2010algorithms}, are
characterized in terms of policies that prescribe actions to be taken.
A central problem in all of these settings is that of policy
evaluation---that is, estimating the performance of a given target
policy.  As a concrete example, given a new policy for deciding
between treatments for cancer patients, one would be interested in
assessing the effect on mortality when it is applied to a certain
population of patients.

Perhaps the most natural idea is to deploy the target policy in an
actual system, thereby collecting a dataset of samples, and use them
to construct an estimate of the performance.  Such an
approach is known as on-policy evaluation, since the policy is
evaluated using data that were collected under the same target policy.
However, on-policy evaluation may not feasible; in certain
applications, it can be costly, dangerous and/or unethical, such as in
clinical trials and autonomous driving.  In light of these concerns, a
plausible work-around is to evaluate the target policy using
historical data collected under a different behavior policy; doing so
obviates the need for any further interactions with the real
environment.  This alternative approach is known as \emph{off-policy
  evaluation}, or OPE for short.  Methods for off-policy evaluation
have various applications, among them news
recommendation~\cite{li2011unbiased}, online
advertising~\cite{theocharous2015personalized},
robotics~\cite{irpan2019off}, to name just a few.  Although OPE is
appealing in not requiring collection of additional data, it also
presents statistical challenges, in that the target policy to be
evaluated is usually different from the behavioral policy that
generates the data.


\subsection{Gaps in current statistical understanding of OPE}
\label{subsec:gaps}

Recent years have witnessed considerable progress in the development
and analysis of methods for OPE.  Nonetheless, there remain a number
of salient gaps in our current statistical understanding of off-policy
evaluation, and these gaps motivate our work.

\paragraph{Non-asymptotic analysis of OPE.} 

The classical analysis of OPE relies upon asymptotics in which the
size of historical dataset, call it $\numobs$, increases to infinity
with all other aspects of the problem set-up held fixed.  Such
analysis shows that a simple plug-in estimator, to be described in the
sequel, is asymptotically efficient for the OPE problem in certain
settings~\cite{hirano2003efficient}. However, such classical analysis
fails to capture the modern practice of OPE, in which the sample size
$\numobs$ may be of the same order as other problem parameters, such
as the number of actions $\numaction$.  Thus, it is of considerable
interest to obtain non-asymptotic guarantees on the performance of
different methods, along with explicit dependence on different problem
parameters.  Li~et~al.~\cite{li2015toward} and
Wang~et~al.~\cite{wang2017optimal} went beyond the asymptotic setting
and studied the OPE problem for multi-armed bandits and contextual
bandits, respectively, from a non-asymptotic perspective. However, as
we discuss in the sequel, their analyses and results are applicable
only when the sample size $\numobs$ is sufficiently large.  In this
large sample regime, a number of estimators, including the plug-in,
importance sampling and Switch estimators to be discussed in this
paper, are all minimax rate-optimal. Thus, analysis of this type falls
short of differentiating between different estimators.  In particular,
are they all rate-optimal for the full range of sample sizes, or is
one estimator better than others?

\paragraph{Known vs.~unknown behavior policies.} 

In practice, the behavior policy generating the historical data might
be known or unknown to the statistician, depending on the application
at hand. This difference in available knowledge raises a natural
question: is there any fundamental difference between OPE problems
with known or unknown behavior policies?  This question, though
natural, appears to have been less explored in the literature. As we
noted above from an asymptotic point of view, the plug-in
estimator---which requires no information about the behavior
policy---is optimal. In other words, asymptotically speaking, knowing
the behavior policy brings no extra benefits to solving the OPE
problem.  Does this remarkable property continue to hold in the finite
sample setting?

\paragraph{OPE with partial knowledge of the behavior policy.}

The known and unknown cases form two extremes of a continuum: in
practice, one often has partial knowledge about the behavior policy.
For instance, one might have a rough idea on how well the behavior
policy covers/approximates the target policy, as measured in terms of
likelihood ratios defined by the two policies.  Alternatively, there
might be a guarantee on the overall exploration level of the behavior
policy, as measured by the minimum probability of observing each
state/action under the behavior policy. How does such extra knowledge
alter the statistical nature of the OPE problem?  Can one develop
estimators that fully exploit this information and yield improvements
over the case of a fully unknown behavior policy?

\subsection{Contributions and organizations}

In this paper, we focus on the off-policy evaluation problem under the
\emph{multi-armed bandit} model with bounded rewards.  This setting,
while seemingly simple, is rich enough to reveal some non-trivial
issues in developing optimal methods for OPE.

More concretely, consider a bandit model with a total of $\numaction$
possible actions to take, also known as arms.  Any (possibly
randomized) policy $\pi$ can be thought of as a probability
distribution over the action space $\ActionSpace \coloneqq \{1,2,
\ldots, \numaction\}$.  Given a target policy $\pitarget$ and a
collection of action-reward pairs $\{(\Action_{i},
\Reward_{i})\}_{i=1}^{\numobs}$ generated i.i.d.\ from the behavior
policy $\pibehave$ and the reward distributions
$\{\RewardDist{\action}\}_{\action \in \ActionSpace}$, the goal of OPE
is to estimate the value function $\Vphi(\pitarget)$ of the target
policy $\pitarget$, given by
\begin{align*}
\Vphi(\pitarget) \coloneqq \sum_{\action \in \ActionSpace}
\pitarget(\action) \reward(\action).
\end{align*}
Here the quantity $\reward(\action) \coloneqq \Exp_{\Reward \sim
  \RewardDist{\action}}[ \Reward]$ denotes the mean reward of the arm
$\action$.  Our goal is to provide a sharp non-asymptotic
characterization of the statistical limits of the OPE problem in three
different settings: (i) when the behavior policy $\pibehave$ is known;
(ii) when $\pibehave$ is unknown; and (iii) when we have partial
knowledge about $\pibehave$.  Along the way, we also develop
computationally efficient procedures that achieve the minimax rates,
up to a universal constant, for all sample sizes. The detailed
statements of our main results are deferred to
Section~\ref{sec:main-results}, but let us highlight here our
contributions that we make in each of the three settings.

\paragraph{Known behavior policy.}

First, when the behavior policy $\pibehave$ is known to the
statistician, we sharply characterize the minimax risk of estimating
the target value function $\Vphi(\pitarget)$ in
Theorem~\ref{thm:main}.  Notably, this bound holds for all sample
sizes, in contrast to previous previous statistical analysis of OPE,
which are either asymptotic or valid only when the sample size is
sufficiently large.  In addition, we show in
Proposition~\ref{prop:switch-general} that the so-called Switch
estimator achieves this optimal risk.  The family of Switch estimators
interpolate between two base estimators: a direct method based on the
plug-in principle applied to actions in some set $S$, and an
importance sampling estimate applied to its complement $S^c$. Our
theory identifies a simple convex program that specifies the optimal
choice of subset: solving this program specifies a threshold level of
the likelihood ratio at which to switch between the two base
estimators.  We prove that this choice yields a minimax-optimal
estimator, one that reduces the variance of the importance sampling
estimator alone.

\paragraph{Unknown behavior policy.}

Moving onto the case when the behavior policy $\pibehave$ is
completely unknown, we first argue that the global minimax risk is no
longer a sensible criterion to measure the performance of different
estimators. Instead, we propose a different metric, namely the
\emph{minimax competitive ratio}, that measures the performance of an
estimator against the best achievable via an oracle---in this setting,
an oracle with the knowledge of the behavior policy. With this new
metric in place, we uncover a fundamental statistical gap between the
known and unknown behavior policy cases in
Theorem~\ref{thm:ratio-lower-bound}. More specifically, when
evaluating a target policy $\pitarget$ that can take at most $s$
actions (for some $s \in \{1, 2, \ldots, \numaction \}$), any
estimator without the knowledge of the behavior policy must pay
multiplicative factor of $s$ (modulo a log factor) compared to the
oracle Switch estimator given knowledge of the behavior policy. We
further demonstrate that the plug-in estimator alone achieves this
optimal worst-case competitive ratio (up to a log factor),
illustrating its near-optimality in the unknown $\pibehave$
case~(cf.~Theorem~\ref{thm:reg-competitive-ratio}).

\paragraph{Partially known behavior policy.}

In the third part of the paper, we initiate the study of the middle
ground between the previous two extreme cases: what if we have some
partial knowledge regarding the behavior policy? More concretely, we
assume the knowledge of the minimum probability $\min_{\action
  \in \ActionSpace} \pibehave(\action)$ that is taken by the behavior
policy in Section~\ref{subsec:partial-knowledge}. Under such
circumstance, we first show that the plug-in estimator is sub-optimal
when the behavior policy is less exploratory---that is, in the regime
$\min_{\action \in \ActionSpace} \pibehave(\action) \ll 
( \log \numaction) / \numobs$.  We then propose a new estimator based on
approximation by Chebyshev polynomials and show that it is optimal in
estimating a large family of target policies. It is worth pointing out
that this optimality is established under a different but closely
related Poisson sampling model---instead of the usual multinomial
sampling one---with the benefit of simplifying the analysis.

\subsection{Related work}

Off-policy evaluation has been extensively studied in the past decades
and by now there has been an immense body of literature on this topic.
Here we limit ourselves to discussion of work directly related to the
current paper.

\paragraph{Various estimators for OPE.}

There exist two classical approaches to the OPE problem. The first is
a direct method based on the plug-in principle: it estimates the value
of the target policy using the reward and/or the transition dynamics
estimated from the data.  In the multi-armed bandit setting, the
direct method uses the data to estimate the mean rewards, and plugs
these estimates into the expression for the target value function. The
other approach is based on importance
sampling~\cite{horvitz1952generalization}, also known as inverse
propensity scoring (IPS) in the causal inference literature.  It
reweights the observed rewards according to the likelihood ratios
between the target and the behavior policies. Both methods are widely
used in practice; we refer interested readers to the recent empirical
study~\cite{paine2020hyperparameter} for various forms of these
estimators.  A number of authors~\cite{thomas2016data,wang2017optimal}
have proposed hybrid estimators that involve a combination of these
two approaches, a line of work that inspired our analysis of the
Switch estimator.  In this context, our novel contribution is to
specify a particular set for switching between the two estimators, and
showing that the resulting Switch estimator is minimax-optimal for any
sample size.

\paragraph{Statistical analysis of OPE. }

Statistical analysis of OPE can be separated into two categories:
asymptotic and non-asymptotic.  On one hand, the asymptotic properties
of the OPE estimators are quite well-understood, with plug-in methods
known to be asymptotically efficient~\cite{hirano2003efficient}, and
asymptotically minimax optimal in multi-armed
bandits~\cite{li2015toward}. Moving beyond bandits, a Cram\'er-Rao
lower bound was recently provided for tabular Markov decision
processes~\cite{jiang2016doubly}, and approaches based on the plug-in
principle were shown to approach this limit
asymptotically~\cite{yin2020asymptotically,duan2020minimax}.

Relative to such asymptotic analysis, there are fewer non-asymptotic
guarantees for OPE; of particular relevance are the two
papers~\cite{li2015toward,wang2017optimal}.
Li~et~al.~\cite{li2015toward} also studied the OPE problem under the
multi-armed bandit model, but under different assumptions on the
reward distributions than this paper.  They proved a minimax lower
bound that holds when the sample size is large enough, but did not
give matching upper bounds in this regime.
Wang~et~al.\cite{wang2017optimal}~extended this line of analysis to
the contextual bandit setting with uncountably many contexts.  They
provided matching upper and lower bounds, but again ones that only
hold when the sample size is sufficiently large.  Notably, in this
large sample regime and under the bounded reward condition of this
paper, all three estimators (plug-in, importance sampling and Switch)
are minimax optimal up to constant factors.  Thus, restricting
attention to this particular regime fails to uncover the benefits of
the Switch estimator. This paper provides a complete picture of the
non-asymptotic behavior of these estimators for the OPE problem,
showing that only the Switch estimator is minimax-optimal for all
sample sizes.

\paragraph{Estimation of nonsmooth functionals via function approximation.} 

The OPE problem with an unknown behavior policy is intimately
connected to the problem of estimating nonsmooth functionals. Portions
of our analysis and the development of the Chebyshev estimator exploit
this connection. The use of function approximation in functional
estimation was pioneered by Ibragimov~et~al.~\cite{ibragimov1987some},
and was later generalized to nonsmooth functionals by
Lepski~et~al.~\cite{lepski1999estimation} and Cai and
Low~\cite{cai2011testing}.  The underlying techniques have been used
to devise optimal estimators for a variety of nonsmooth functionals,
including Shannon
entropy~\cite{valiant2010clt,wu2016minimax,jiao2015minimax}, KL
divergence~\cite{han2016minimax}, support
size~\cite{valiant2011estimating,valiant2017estimating,wu2019chebyshev},
among others. Our development of the Chebyshev estimator is largely
inspired by the paper~\cite{wu2019chebyshev} on estimating the support
size, which can be viewed as a special case of OPE.

\paragraph{Notation:} 
For the reader's convenience, let us summarize the notation used
throughout the remainder of the paper. We reserve boldfaced symbols
for vectors. For instance, the symbol $\zeros$ denotes the all-zeros
vector, whose dimension can be inferred from the context.  For a
positive integer $\numaction$, we refer to $[\numaction]$ as the set
$\{1, 2, \ldots, \numaction\}$. For a finite set $\Set$, we use
$|\Set|$ to denote its cardinality. We denote by $\indicator\{ \Event
\}$ the indicator of the event $\Event$. For any distribution $\mu$ on
$\real$, we denote by $\mysupp(\mu)$ its support. For any distribution
$\pi$ on $\ActionSpace$ and any subset $\Set \subseteq \ActionSpace$,
we define $\pi(\Set) \coloneqq \sum_{\action \in \Set}
\pi(\action)$. We follow the convention that $0 / 0 = 0$.


\section{Background and problem formulation}
\label{sec:setup}

In this section, we introduce the multi-armed bandit model with
stochastic rewards, and then formally define the off-policy evaluation
(OPE) problem in this bandit setting.  We also introduce two existing
estimators---the plug-in and the importance sampling estimators---for
the OPE problem.

\subsection{Multi-armed bandits and value functions}

A multi-armed bandit (MAB) model is specified by an action space
$\GeneralActionSpace$ and a collection of reward distributions
$\RewardDistPlain \coloneqq \{\RewardDist{a}\}_{\action
  \in \GeneralActionSpace}$, where $\RewardDist{a}$ is the reward
distribution associated with the action or arm~$\action$.  Throughout
the paper, we focus on the MAB model with $\numaction$ possible
actions, and we index the action space $\GeneralActionSpace$ by
$\ActionSpace = \{1,2, \ldots, \numaction\}$.  In addition, we assume
that the collection of reward distributions $\RewardDistPlain$ belongs
to the family of distributions with bounded support---that is,
\begin{align}
\label{eq:reward-family}  
\RewardFamily(\Rmax) \coloneqq \left\{ \RewardDistPlain \mid
\mbox{$\mysupp (\RewardDist{\action}) \subseteq [0, \Rmax]$ for each
  $\action \in [\numaction]$} \right \}.
\end{align}
When the maximum reward $\Rmax$ is understood from the context, we
adopt the shorthand $\RewardFamily$ for this class of distributions.

A (randomized) policy $\pi$ is simply a distribution over the action
space $\ActionSpace$, where $\pi(\action)$ specifies the probability
of selecting the action $\action$.  Correspondingly, we can define the
value function $\Vphi(\pi)$ of the policy $\pi$ to be
\begin{align}
\label{def:value-function}  
\Vphi(\pi) \coloneqq \sum_{\action \in \ActionSpace} \pi(\action)
\reward(\action),
\end{align}
where $\reward(\action) \coloneqq \Exp_{\RewardDistPl}[\Reward \mid
  \Action = \action ]$ denotes the mean reward under $\RewardDistPl$
given that action $\action$ is taken.  Here $\Reward$ denotes a reward
random variable distributed according to $\RewardDist{\action}$.


\subsection{Observation model and off-policy evaluation}

Suppose that we have collected a collection of pairs $\{ (
\Action_{i}, \Reward_{i}) \}_{i=1}^\numobs$, where the action
$\Action_{i}$ is randomly drawn from the behavior policy
$\pibehave$, whereas the reward $\Reward_{i}$ is distributed according
to the reward distribution $\RewardDist{\Action_i}$.  Given a target
policy $\pitarget$, the goal of \emph{off-policy evaluation} (OPE) is
to evaluate the value function of the target policy, given by
\begin{align}
\label{eq:OPE-obj}  
\Vphi(\pitarget) & = \sum_{\action \in \ActionSpace}
\pitarget(\action) \reward(\action).
\end{align}
Note that this problem is non-trivial because the data $\{ (
\Action_{i}, \Reward_{i}) \}_{i=1}^\numobs$ is collected under the
behavior policy~$\pibehave$, which is typically distinct from the
target policy~$\pitarget$.

\subsection{Plug-in and importance sampling estimators} 

A variety of estimators have been designed to estimate the value
function $\Vphi(\pitarget)$. Here we introduce two important ones most
relevant to our development, namely the plug-in estimator and the
importance sampling estimator.  We note that in some of the
literature, the plug-in estimator is also known as the regression
estimator.


\paragraph{Plug-in estimator.} 

Perhaps the simplest method is based on applying the usual plug-in
principle.  Observe that the only unknown quantities in the
definition~\eqref{eq:OPE-obj} of the value function are the mean
rewards $\{\reward(\action)\}$.  These unknown quantities can be
estimated by their empirical counterparts
\begin{align}
\label{eq:empirical-reward}
\rhat(\action) \coloneqq
\begin{cases}
  \frac{1}{\numobs(\action)} \sum_{i=1}^{\numobs} \Reward_{i} \indicator
  \{\Action_{i}=a\}, & \mbox{if $\numobs(\action) \geq 1$, and} \\
  0 & \mbox{otherwise,}
\end{cases}
\end{align}
where $\numobs(\action) \coloneqq \sum_{i=1}^{\numobs} \indicator
\{ \Action_{i} = \action \}$ denotes the number of times that action
$\action$ is observed in the data set.  Substituting these empirical
estimates into the definition of the value function yields the
plug-in estimator
\begin{align}
\label{eq:plug-in-estimator}
\VhatPlug \coloneqq \sum_{\action \in \ActionSpace}
\pitarget(\action) \rhat(\action),
\end{align}
Observe that this estimator is fully agnostic to the behavior policy.
Thus, it can also be used when the behavior policy $\pibehave$ is
unknown, a setting that we also study in the sequel.


\paragraph{Importance sampling estimator.} 

An alternative estimator, one which does require knowledge of the
behavior policy, is based on the idea of importance sampling.  More
precisely, let \mbox{$\likerat(\action) \coloneqq \pitarget(\action) /
  \pibehave(\action)$} denote the likelihood ratio associated with the
action $\action$. The importance sampling (IS) estimator is given by
\begin{align}
\label{eq:IS-estimator}
\VhatIS & \coloneqq \frac{1}{\numobs} \sum_{i=1}^{\numobs}
\likerat(\Action_{i}) \Reward_{i}.
\end{align}
In words, it weighs the observed reward $\Reward_{i}$ based on the
corresponding likelihood ratio $\likerat(\Action_{i})$.  As long as
$\likerat(\action) < \infty$ for all $\action \in \ActionSpace$, the
importance sampling estimator $\VhatIS$ is an unbiased estimate of
$\Vphi(\pitarget)$.  Note that the IS estimate relies on knowledge of
the behavior policy $\pibehave$ via its use of the likelihood ratio.


\section{Main results}
\label{sec:main-results}

We now move onto the main results of this paper.  We begin in
Section~\ref{subsec:known-pi-b} with results in the case when the
behavior policy $\pibehave$ is known \emph{a priori}.  In
Section~\ref{SecCompetitiveRatio}, we provide guarantees when the
behavior policy is completely unknown, whereas
Section~\ref{subsec:partial-knowledge} is devoted to the setting where
certain partial knowledge about the behavior policy, say the minimum
value $\min_{\action \in \ActionSpace} \pibehave(\action)$, is known.

\subsection{Switch estimator with known $\pibehave$}
\label{subsec:known-pi-b}

When the behavior policy is known, both the plug-in estimator and the
importance sampling estimator are applicable.  In fact, they belong to
the family of \emph{Switch} estimators, as introduced in
past\footnote{To be clear, the paper~\cite{wang2017optimal} considers
  a restricted version of this family, in which the subset $\Set$ is
  restricted to be of the form $\{\action \in \ActionSpace \mid
  \likerat(\action) \geq \tau\}$ for some threshold $\tau \geq 0$,
  whereas we define the estimator for any set.}
work~\cite{wang2017optimal}. For any subset $\ActionSubset
\subseteq \ActionSpace$, we define the Switch estimator associated
with $\ActionSubset$ as
\begin{align}
\label{eq:switch-general}
\VhatSwitch(\ActionSubset) & \coloneqq \sum_{\action \in
  \ActionSubset} \pitarget(\action) \rhat(\action) + 
  \frac{1}{\numobs} \sum_{i=1}^{\numobs} \likerat(\Action_{i}) 
  \Reward_{i} \indicator \{\Action_{i} \notin \ActionSubset\},
\end{align}
where $\rhat(\action)$ is the empirical mean reward defined in
equation~\eqref{eq:empirical-reward}.  By making the choices
$\ActionSubset = \ActionSpace$ or $\ActionSubset = \emptyset$,
respectively, the Switch estimator $\VhatSwitch(\ActionSubset)$ reduces either to the plug-in estimator~\eqref{eq:plug-in-estimator} or to the IS estimator~\eqref{eq:IS-estimator}.  Choices of 
$\ActionSubset$ intermediate between these two extremes allow us to interpolate (or switch) between the plug-in estimator and the IS estimator.

The following proposition, whose proof is relatively elementary,
provides a unified performance guarantee for the family of Switch
estimators. 

\begin{proposition}
\label{prop:switch-general}
For any subset $\ActionSubset \subseteq \ActionSpace$, we have
\begin{align}
\label{eq:switch-general-bound}  
\Exp_{\pibehave \otimes \RewardDistPl} [(\VhatSwitch(\ActionSubset) -
  \Vphi(\pitarget))^{2}] & \leq 3 \Rmax^{2} \left \{
\pitarget^{2}(\ActionSubset) + \frac{\sum_{\action \notin
    \ActionSubset} \pibehave(\action) \likerat^{2}(\action)}{\numobs}
\right \}.
\end{align}
\end{proposition}
\noindent
See Section~\ref{SecProof:prop:switch-general} for the proof of this claim.

\bigskip
Given the family of Switch estimators $\{\VhatSwitch(\Set)\}_{\Set \subseteq \ActionSpace}$, it is natural to ask: how to choose the subset $\ActionSubset$ among all possible subsets of the
action space?  The unified upper bounds established in 
Proposition~\ref{prop:switch-general} offer us a reasonable guideline: one should select a subset $\ActionSubset$ to minimize the error 
bound~\eqref{eq:switch-general-bound}, i.e.,
\begin{align}
\label{eq:minimize-switch}
\min_{\ActionSubset \subseteq \ActionSpace} \left \{
\pitarget^{2}(\ActionSubset) + \frac{\sum_{\action \notin
    \ActionSubset} \pibehave(\action) \likerat^{2}(\action)}{\numobs}
\right\}.
\end{align}
At first glance, the minimization problem~\eqref{eq:minimize-switch}
is combinatorial in nature, which indicates the possible computational
hardness in solving it. Fortunately, it turns out that such an
``ambitious'' goal can instead be achieved via solving a tractable
convex program.  To make this claim precise, let us consider the
following convex program
\begin{align}
\label{eq:key-opt}  
\min_{\vb \in \real^{\numaction}} \left \{ \sqrt{\frac{1}{8 \numobs}
  \sum_{\action \in \ActionSpace}
  \frac{[\pitarget(\action)-v(\action)]^{2}}{\pibehave(\action)}} +
\frac{1}{2}\sum_{\action \in \ActionSpace} | v(\action)| \right \},
\end{align}
where $\vb = (v(1), v(2), \ldots, v(\numaction))^{\top}$ is a vector
of decision variables. Let $\vb^{\star}$ be a minimizer of this
optimization problem~\eqref{eq:key-opt}, whose existence is guaranteed
by the coerciveness of the objective function.  Correspondingly, we
define
\begin{align}
\label{eq:defn-S-star}  
\SetStar \coloneqq \{a\mid v^{\star}(\action)\neq0\}
\end{align}
to be the support of $\vb^{\star}$. It turns out that the choice
$\ActionSubset = \SetStar$ solves the best subset selection
problem~\eqref{eq:minimize-switch} up to a constant factor.  We
summarize in the following:

\begin{proposition}
\label{prop:optimal-subset}
There exists a universal constant $\plaincon > 0$ such that
\begin{align}
\label{eq:min-upper-bound}  
\min_{\ActionSubset \subseteq \ActionSpace} \left \{
\pitarget^{2}(\ActionSubset) + {\frac{\sum_{\action \notin
      \ActionSubset} \pibehave(\action)
    \likerat^{2}(\action)}{\numobs}} \right \} & \geq \plaincon \left
\{\pitarget^{2}(\SetStar) + {\frac{\sum_{\action \notin \SetStar}
    \pibehave(\action) \likerat^{2}(\action)}{\numobs}} \right \}.
\end{align}
\end{proposition}
\noindent
See Section~\ref{sec:proof-of-optimal-subset} for the proof of the optimality of $\SetStar$.

\bigskip
Thus, we conclude that the among the family of Switch estimators, the
optimal estimator is given by
\begin{subequations}
\begin{align}
\label{eq:switch-optimal}  
\VhatSwitch(\SetStar) \coloneqq \sum_{\action \in \SetStar} \pitarget(\action)
\rhat(\action) + \frac{1}{\numobs} \sum_{i=1}^{\numobs}
\likerat(\Action_{i}) \Reward_{i} \indicator \{\Action_{i} \notin
\SetStar\}.
\end{align}
In view of Proposition~\ref{prop:switch-general}, it enjoys the
following performance guarantee
\begin{align}\label{eq:optimal-upper-bound}
\Exp_{\pibehave \otimes
  \RewardDistPl}[(\VhatSwitch(\SetStar)-\Vphi(\pitarget))^{2}] \leq 3
\Rmax^{2}\left\{ \pitarget^{2}(\SetStar)+\frac{\sum_{a \notin
    \mathcal{S}^{\star}}
  \pibehave(\action)\likerat^{2}(\action)}{\numobs}\right\} .
\end{align}
\end{subequations}
From now on, we shall refer to $\VhatSwitch(\SetStar)$ as the Switch
estimator.


\subsubsection{Is the Switch estimator optimal?}
\label{subsec:optimality-switch}

The above discussion establishes the optimality of the Switch
estimator $\VhatSwitch(\SetStar)$ among the family of
estimators~\eqref{eq:switch-general} parameterized by a choice of
subset $\Set$.  However, does the Switch estimator continue to be
optimal in a larger context?  This question can be assessed by
determining whether it achieves, say up to a constant factor, the
\emph{minimax risk} given by
\begin{align}
\label{eq:defn-R-n-star}
\Risk_{\numobs}^{\star}(\pitarget; \pibehave)
\coloneqq \inf_{\Vhat}
\sup_{\RewardDistPl \in \RewardFamily} \Exp_{\pibehave \otimes
  \RewardDistPl} [(\Vhat - \Vphi(\pitarget))^{2}],
\end{align}
Here the infimum ranges over all measurable functions $\Vhat$ of the
data $\{(\Action_{i},\Reward_{i})\}_{i=1}^\numobs$, whereas the
supremum is taken over all reward distributions $\RewardDistPl$
belonging to our family $\RewardFamily$ of bounded mean distributions.
The following theorem provides a lower bound on this minimax risk:
\begin{theorem}
\label{thm:main}
There exists a universal positive constant $\plaincon$ such that for
all pairs $(\pibehave, \pitarget)$, we have
\begin{align*}
  \Risk_{\numobs}^{\star}(\pitarget; \pibehave) & \geq \plaincon \:
  \Rmax^{2}\left\{ \pitarget^{2}(\SetStar) + \frac{\sum_{\action
      \notin \SetStar} \pibehave(\action)
    \likerat^{2}(\action)}{\numobs} \right\} .
\end{align*}
\end{theorem}
\noindent
See Section~\ref{subsec:proof-of-theorem-switch} for the proof of this lower bound. 

\bigskip

By combining Theorem~\ref{thm:main} and the upper
bound~\eqref{eq:optimal-upper-bound} on the mean-squared error of the
Switch estimator $\VhatSwitch(\SetStar)$, we obtain a finite-sample
characterization of the minimax risk up to universal
constants---namely
\begin{align}
\label{eq:minimax-rate-known-pi}
\Risk_{\numobs}^{\star}(\pitarget; \pibehave) \asymp \Rmax^{2} \left\{
\pitarget^{2}(\SetStar) + \frac{\sum_{\action \notin \SetStar}
  \pibehave(\action) \likerat^{2}(\action) }{\numobs} \right\}.
\end{align}
Consequently, we see that the Switch estimator $\VhatSwitch(\SetStar)$
is optimal among all estimators in a minimax sense.

In order to gain intuition for this optimality result, it is helpful
to consider some special cases.

\bigskip

\paragraph{Degenerate case of on-policy evaluation:}
First, consider the degenerate setting $\pitarget = \pibehave$, so
that our OPE problem actually reduces to a standard on-policy
evaluation problem.  In this case, the IS estimator reduces to the
standard Monte Carlo estimate
\begin{align*}
\VhatIS = \frac{1}{\numobs} \sum_{i=1}^{\numobs}
\likerat(\Action_{i}) \Reward_{i} = \frac{1}{\numobs} 
\sum_{i=1}^{\numobs} \Reward_{i}.
\end{align*}
A straightforward calculation shows that it has mean-squared error
$\Rmax^2 / \numobs$, which we claim is order-optimal.  To reach this
conclusion from our expression~\eqref{eq:minimax-rate-known-pi} for
the minimax risk, it suffices to check that $\vb^{\star}=\zeros$ is a
minimizer of the optimization problem~\eqref{eq:key-opt}.  This fact
can be certified by showing that the all-zeros vector $\zeros$ obeys
the first-order optimality condition associated with the convex
program~\eqref{eq:key-opt}.  More precisely, for all actions $\action
\in \ActionSpace$, we have
\begin{align}
\label{eq:opt-condition-main-text}
\sqrt{\frac{1}{8 \numobs}}
\frac{\pitarget(\action)/\pibehave(\action)}{\sqrt{\sum_{\action \in
      \ActionSpace}
    \frac{[\pitarget(\action)]^{2}}{\pibehave(\action)}}} =
\sqrt{\frac{1}{8 \numobs}} \leq \frac{1}{2}.
\end{align}

\paragraph{Large-sample regime:}

Returning to the general off-policy case ($\pitarget \neq \pibehave$),
suppose that the sample size $\numobs$ satisfies a lower bound of the
form
\begin{align}
\label{eq:sample-size-requirement}
\numobs & \geq c \, \frac{\max_{\action \in \ActionSpace}
  \likerat^{2}(\action)}{\sum_{\action \in \ActionSpace}
  \pibehave(\action) \likerat^{2}(\action)}
\end{align}
for a sufficiently large constant $c$.  In this case, one can again
verify that the all-zeros vector $\zeros$ is optimal for the convex
program~\eqref{eq:key-opt} by verifying the first-order optimality
condition~\eqref{eq:opt-condition-main-text}.  As a consequence, we
conclude that the Switch estimator reduces to the IS estimator in the
\emph{large-sample regime} defined by the lower
bound~\eqref{eq:sample-size-requirement}.  In this regime, the IS
estimator achieves mean-squared error $\Rmax^2 \cdot \sum_{\action
  \in \ActionSpace} \pibehave(\action) \likerat^{2}(\action)/\numobs$.
Under the bounded reward condition, this result recovers the rate
provided by Li~et~al.~\cite{li2015toward} in the large sample
regime~\eqref{eq:sample-size-requirement} up to a constant factor; see
Theorem 1 in their paper.
\bigskip

It is worthwhile elaborating further on the connections with the paper
of Li~et~al.~\cite{li2015toward}: they studied classes of reward
distributions that are parameterized by bounds on their means (as
implied by our bounded rewards) and variances. In this sense, their
analysis is finer-grained than our study of bounded rewards only.
However, when their results are specialized to bounded reward
distributions~\eqref{eq:reward-family}, their minimax risk result
(cf.~equation (2) in their paper) applies only in the large sample
regime defined by the lower bound~\eqref{eq:sample-size-requirement}.
As we have discussed, when this lower bound holds, the IS estimator
itself is order optimal, so analysis restricted to this regime fails
to reveal the tradeoff between the two terms in the minimax
rate~\eqref{eq:minimax-rate-known-pi}, and in particular, the
potential sub-optimality of the IS estimator (as reflected by the
presence of the additional term $\pitarget^2(\SetStar)$ in the minimax
bound).


\subsubsection{A closer look at the Switch estimator}

In this subsection, we take a closer look at some properties of the
Switch estimator, and in particular its connection to truncation of
the likelihood ratio.

\paragraph{Link to likelihood truncation.}
We begin by investigating the nature of the best subset $\SetStar$, as
defined in equation~\eqref{eq:defn-S-star}.  Let us assume without
loss of generality that the actions are ordered according to the
likelihood ratios---viz.
\begin{align}
  \label{eq:likelihood-order}
 \likerat(1) \leq \likerat(2) \leq \cdots \leq \likerat(\numaction).
 \end{align} 
Under this condition, unraveling the proof of
Proposition~\ref{prop:optimal-subset} shows that the optimal subset
$\SetStar$ takes the form
\begin{align}
  \label{EqnTruncate}
\SetStar & = \{s, s + 1, \ldots, \numaction\} \quad \mbox{for some
  integer $s \in [\numaction]$.}
\end{align}
Here it should be understood that the choice $s = \numaction$
corresponds to $\SetStar = \emptyset$.  Thus, the significance of the
optimization problem~\eqref{eq:key-opt} is that it specifies the
optimal threshold at which to truncate the likelihood ratio.

As noted previously, Wang~et~al.~\cite{wang2017optimal} studied the
sub-family of Switch estimators obtained by varying the truncation
thresholds of the likelihood ratios. Similar to
Li~et~al.~\cite{li2015toward}, they studied the large sample regime in
which the IS estimator without any truncation is already minimax
optimal up to constant factors. This fails to explain the benefits of
truncating large likelihood ratios and the associated Switch
estimator. In contrast, the key optimization
problem~\eqref{eq:key-opt} informs us of the optimal subset $\SetStar$
and hence an optimal truncation threshold, which allows the Switch
estimator $\VhatSwitch(\SetStar)$ to optimally estimate the target
value function for all sample sizes.  This result is especially
relevant for smaller sample sizes in which the problem is challenging,
and the IS estimator can exhibit rather poor behavior.

\paragraph{Role of the plug-in component.} 

The Switch estimator~\eqref{eq:switch-optimal} is based on applying
the plug-in principle to the actions in $\SetStar$ with large
likelihood ratios.  However, doing so is not actually necessary to
achieve the optimal rate of
convergence~\eqref{eq:minimax-rate-known-pi}.  In fact, if we simply
estimate the mean reward by zero for any action in $\SetStar$, then we
obtain the estimate
\begin{align}
\label{eq:truncated-IS}
\Vhat & \coloneqq \frac{1}{\numobs} \sum_{i=1}^{\numobs}
\likerat(\Action_{i}) \Reward_{i} \indicator \{\Action_{i} \notin
\SetStar \},
\end{align}
which is also minimax-optimal up to a constant factor.  The intuition
is that for actions on the support $\SetStar$, the likelihood ratios
are so large that the off-policy data is essentially useless, and can
be ignored.  It suffices to use the zero estimate, yielding a squared
bias of the order $\pitarget^2(\SetStar)$.  On the other hand, for
actions in the complement $\SetStarComp$, the likelihood ratios are
comparatively small, so that the off-policy data should be exploited.

We note that truncated IS estimators of the
type~\eqref{eq:truncated-IS} have been explored in empirical work on
counterfactual reasoning~\cite{bottou2013counterfactual} and
reinforcement learning~\cite{wawrzynski2007truncated}; our work
appears to be the first to establish their optimality for general
likelihood ratios.  Also noteworthy is the paper by
Ionides~\cite{ionides2008truncated}, who analyzed the rate at which
the truncation level should decay, assuming that the likelihood ratios
decay at a polynomial rate.  Our theory, while focused on finite
action spaces, instead works for any configuration of the likelihood
ratios, and in addition provides a precise truncation level instead of
only a rate.

\subsubsection{Numerical experiments} 

In this section, we report the results of some simple numerical
experiments on simulated data that serve to illustrate the possible
differences between the three methods: Switch, plug-in and IS
estimators.  importance sampling estimators.  We performed experiments
with the uniform target policy (i.e., $\pitarget(\action) = 1 /
\numaction$ for all actions $\action \in \ActionSpace$), and for each
action $\action$, we defined the reward distribution
$\RewardDist{\action}$ to be an equi-probable Bernoulli distribution
over $\{0, 1\}$, so that $\Rmax = 1$.

For each choice of $\numaction$, we constructed a behavior policy of
the following form
\begin{align*}
\pibehave(1) = \pibehave(2) = \cdots = \pibehave(\sqrt{\numaction}) =
\frac{1}{\numaction ^ 2},\quad \text{and}\quad
\pibehave(\sqrt{\numaction} + 1) = \pibehave(\sqrt{\numaction} + 2) =
\cdots = \pibehave(\numaction) = \frac{1 - \frac{1}{\numaction^{3/2}
}} {\numaction - \sqrt{\numaction}}.
\end{align*}
In words, we set the first $\sqrt{\numaction}$ actions with a low
probability $\frac{1}{\numaction^2}$, whereas for the remaining
$\numaction - \sqrt{\numaction}$ actions, the behavior probabilities
are relatively large, which is close to $\frac{1}{\numaction}$.  As we
will see momentarily, this choice allows us to demonstrate interesting
differences between the three estimators.

As is standard in high-dimensional statistics~\cite{Wai19}, we study a
sequence of such problems indexed by the pair $(\numobs, \numaction)$;
in order to obtain an interesting slice of this two-dimensional space,
we set $\numobs = 1.5 \numaction$.  For such a sequence of problems,
we can explicitly compute that the mean-squared errors of the three
estimators scale as follows:
\begin{subequations}
\label{EqnTheory}
  \begin{align}
\Exp_{\pibehave \otimes \RewardDistPl} [(\VhatPlug -
  \Vphi(\pitarget))^{2}] & \asymp 1, \\ \Exp_{\pibehave \otimes
  \RewardDistPl} [(\VhatIS - \Vphi(\pitarget))^{2}] & \asymp
\numobs^{-1/2}, \quad \text{and} \\
\Exp_{\pibehave \otimes \RewardDistPl} [(\VhatSwitch(\SetStar) -
  \Vphi(\pitarget))^{2}] & \asymp \numobs^{-1}.
\end{align}
\end{subequations}
The purpose of our numerical experiments is to illustrate this
theoretically predicted scaling.

\begin{figure}  
	\centering
        \includegraphics[scale=0.5]{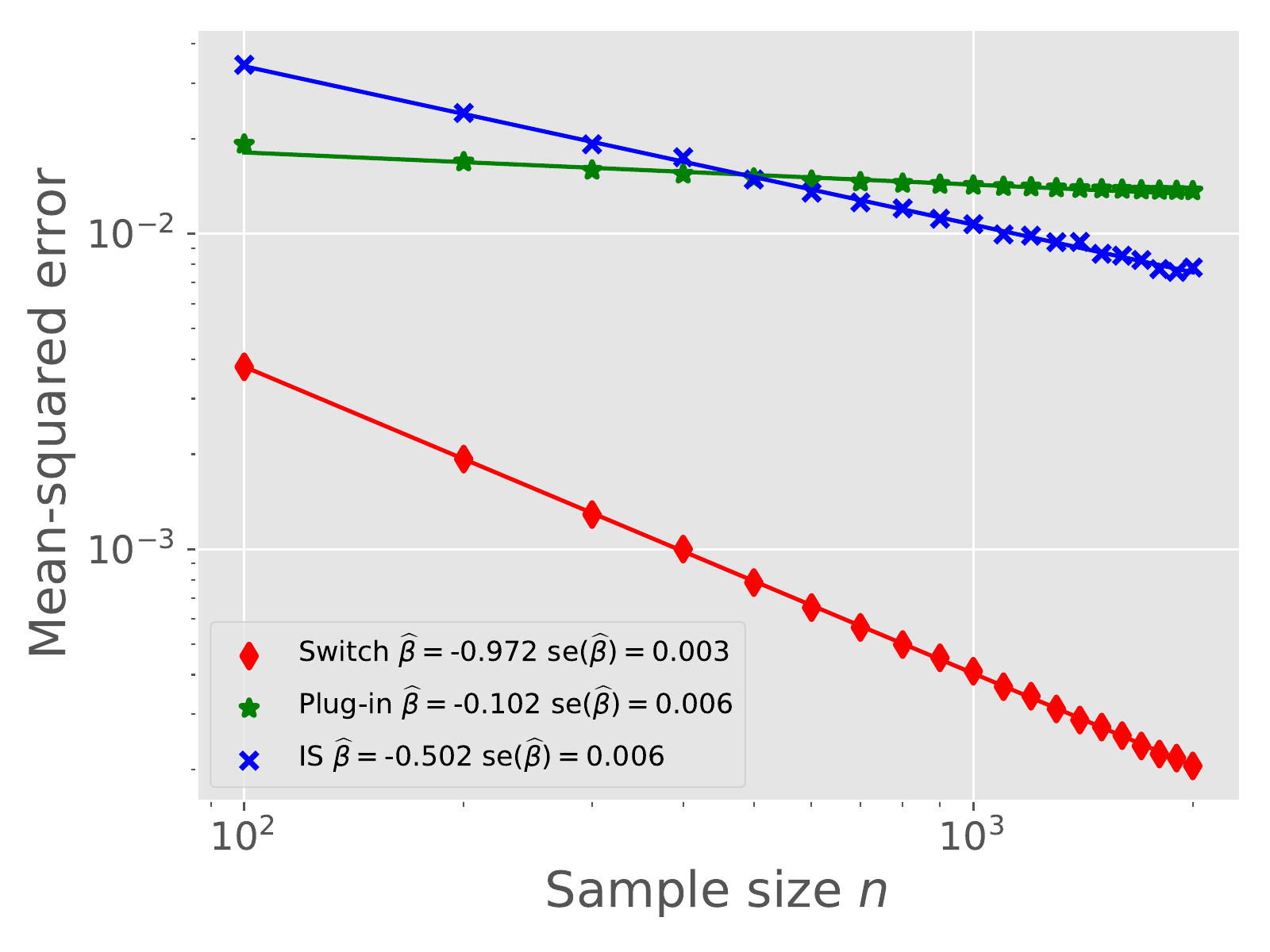}
	\caption{Log-log plot of the estimation errors vs.~the sample
          size. All results reported are averaged over $10^5$ random
          trials. The legends also contain the estimated slopes and
          their standard errors obtained by performing linear
          regressions on the logarithms of the error and the sample
          size. }
        \label{fig:switch-simulated}
\end{figure}

Figure~\ref{fig:switch-simulated} shows the mean-squared errors of
these three estimators versus the sample size $\numobs$, plotted on a
log-log scale. The results are averaged over $10^5$ random trials. As
can be seen from Figure~\ref{fig:switch-simulated}, the Switch
estimator performs better than the two competitors uniformly across
different sample sizes.  Note that our theory~\eqref{EqnTheory}
predicts that the mean-squared errors should scale as
$\numobs^{-\beta^*}$, where $\beta^* \in \{0, 1/2, 1 \}$ for the
plug-in, IS, and SWITCH estimators respectively.  In order to assess
these theoretical predictions, we performed a linear regression of the
log MSE on $\log(\numobs)$, thereby obtaining an estimated exponent
$\slope$ for each estimator.  These estimates and their standard
errors are shown in the legend of
Figure~\ref{fig:switch-simulated}. Clearly, the estimated slopes are
quite close to the theoretical predictions~\eqref{EqnTheory}.

\subsection{OPE when $\pibehave$ is unknown: competitive ratio}
\label{SecCompetitiveRatio}

Our analysis thus far has taken the behavior policy $\pibehave$ to be
known.  This condition, while realistic in some settings, is
unrealistic in others.  Accordingly, we now turn to the version of the
OPE problem in which the only knowledge provided are the action-reward
pairs $\{(\Action_{i}, \Reward_{i})\}_{i=1}^\numobs$.  Note that the
importance sampling estimator $\VhatIS$ and Switch estimators
$\VhatSwitch(\Set)$ are no longer applicable, since they require
knowledge of the behavior policy.  Consequently, we are led to the
natural question: what is an optimal estimator when $\pibehave$ is
unknown?  Before answering this question, one needs to first settle
upon a suitable notion of optimality.


\subsubsection{Optimality via the minimax competitive ratio}

The first important observation is that when the behavior policy is
unknown, the global minimax risk is no longer a suitable metric for
assessing optimality.  Indeed, for any target policy $\pitarget$, one
can construct a ``nasty'' behavior policy $\pibehave$ such that for any
estimator $\Vhat$, we have a lower bound of the form
\begin{align*}
\sup_{\RewardDistPl \in \RewardFamily} \Exp_{\pibehave \otimes
  \RewardDistPl}[(\Vhat - \Vphi(\pitarget))^{2}] & \geq c \, \Rmax^2,
\end{align*}
for some universal constant $c > 0$.  For this reason, if we measure
optimality according to the global minimax risk, then the trivial
``always return zero'' estimator $\Vhat \equiv 0$ is optimal, and hence
 the global minimax risk is not a sensible criterion in
this setting.

This pathology arises from the fact that the adversary has too much
power: it is allowed to choose an arbitrarily bad behavior policy
while suffering no consequences for doing so.  In order to mitigate
this deficiency, it is natural to consider the notion of a
\emph{competitive ratio}, as is standard in the literature on online
learning~\cite{fiat1998online}.  An analysis in terms of the
competitive ratio measures the performance of an estimator against the
best achievable by some oracle---in this case, an oracle equipped with
the knowledge of $\pibehave$.

For a given target policy $\pitarget$ and behavior policy $\pibehave$,
recall the definition~\eqref{eq:defn-R-n-star} of the minimax risk
$\Risk_{\numobs}^{\star}(\pitarget; \pibehave)$; it corresponds to
smallest mean-squared error that can be guaranteed, uniformly over a
class of reward distributions $\RewardFamily$, by any method equipped
with the oracle knowledge of $\pibehave$.  Given an estimator $\Vhat$
and a reward distribution $\RewardDistPl$, we can measure its
performance relative to this oracle lower bound via the
\emph{competitive ratio}
\begin{align}
\label{def:competitive-ratio}
\CompRat \big(\Vhat; \pitarget, \pibehave, \RewardDistPl \big)
\coloneqq \frac{\Exp_{\pibehave \otimes \RewardDistPl}[(\Vhat -
    \Vphi(\pitarget))^{2}]}{\Risk_{\numobs}^{\star}(\pitarget;
  \pibehave)}.
\end{align}
An estimator $\Vhat$ with a small competitive ratio---that is, close
to $1$---is guaranteed to perform almost as well as the oracle that
knows the behavior policy $\pibehave$. On the other hand, a large
competitive ratio indicates poor performance relative to the oracle.

As one concrete example, the ``always return zero'' estimator $\Vhat
\equiv 0$ is far from ideal when considered in terms of the
competitive ratio~\eqref{def:competitive-ratio}.  Indeed, suppose that
$\pibehave = \pitarget$ and $\reward(\action) = \Rmax/2$; we then have
\begin{align}
\label{eq:competitive-ratio-zero-estimator}
\sup_{\pibehave, \RewardDistPl \in \RewardFamily}
\frac{\Exp_{\pibehave \otimes \RewardDistPl}[(\Vhat -
    \Vphi(\pitarget))^{2}]} {\Risk_{\numobs}^{\star}(\pitarget;
  \pibehave)} & \geq \frac{\Exp_{\pitarget \otimes
    \RewardDistPl}[(\Vhat - \Vphi(\pitarget))^{2}]}
     {\Risk_{\numobs}^{\star}(\pitarget;\pibehave)}
     \stackrel{(i)}{\asymp} \frac{\Exp_{\pitarget \otimes
         \RewardDistPl} [(\Vphi(\pi))^{2}]}{ \Rmax^{2} / \numobs}
     \stackrel{(ii)}{\asymp} \numobs.
\end{align}
Here step (i) follows from the fact that $\Vhat \equiv 0$ by definition,
along with the scaling
\mbox{$\Risk_{\numobs}^{\star}(\pitarget;\pitarget) \asymp
  \tfrac{\Rmax^{2}}{\numobs}$} established in
Section~\ref{subsec:optimality-switch}.  Step (ii) follows from the
assumption that $\reward(\action)=\Rmax/2$, which implies that
$\Exp_{\pitarget \otimes \RewardDistPl} [(\Vphi(\pi))^{2}] = \Rmax^2 /
4$.  Thus, we see that the ``always return zero'' estimator $\Vhat
\equiv 0$ performs extremely badly relative to the oracle, and its
competitive ratio further degrades as the sample size $\numobs$
increases.


\subsubsection{Competitive ratio of the plug-in estimator}

As we have emphasized earlier, the plug-in approach is applicable even
if the behavior policy $\pibehave$ is unknown. The following theorem
provides a guarantee on its behavior in terms of the competitive
ratio:
\begin{theorem}
\label{thm:reg-competitive-ratio}
There exists a universal constant $\plaincon> 0$ such that for any
target policy $\pitarget$, the plug-in estimator $\VhatPlug$ satisfies
the bound
\begin{align}
  \sup_{\pibehave,\RewardDistPl \in \RewardFamily} \CompRat
  \big(\Vhat; \pitarget, \pibehave, \RewardDistPl \big) & \leq
  \plaincon \, |\mysupp(\pitarget)|.
\end{align}
\end{theorem}
\noindent See Section~\ref{sec:Proof-of-Theorem-reg-competitive-ratio}
for the proof of this theorem.

\begin{figure}	
  \centering
  \includegraphics[scale = 0.5]{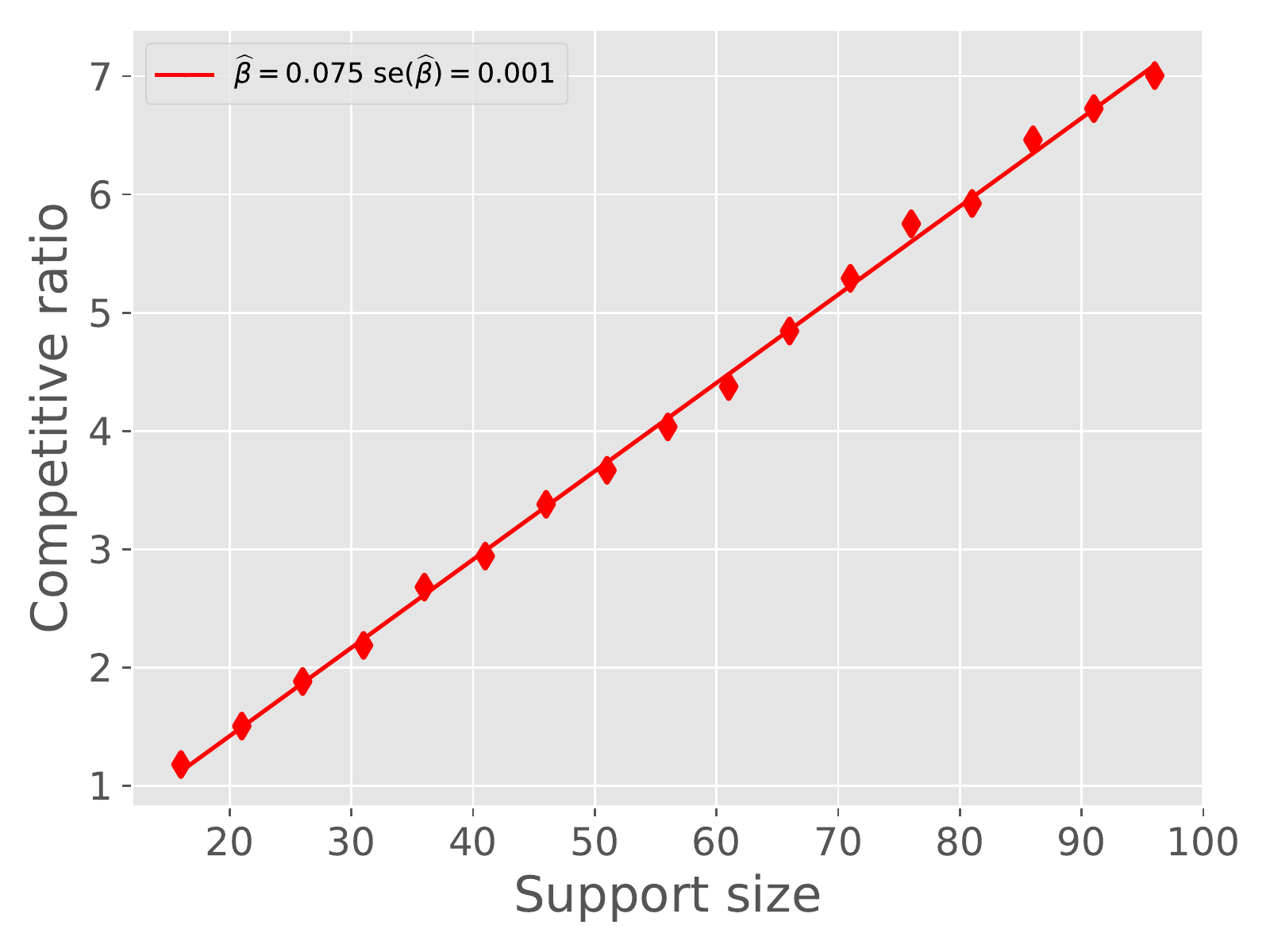}
  \caption{Illustration of the competitive ratio the plug-in estimator
    vs.~the support size of the target policy. Throughout the
    experiments, we set $\numaction = 100, \numobs = 2
    \numaction$. The behavior policy obeys $\pibehave(\action) =
    (\numobs \log \numaction)^{-1}$ for $\action \in [\numaction -1]$,
    and $\pibehave(\numaction) = 1 - (\numaction - 1) / (\numobs \log
    \numaction)$. For each support size $s$, we take the target policy
    $\pitarget$ to be the uniform distribution over $[s]$. Since
    $\Risk_{\numobs}^{\star}(\pitarget;\pibehave)$ is not known
    precisely, we use the mean-squared error of the Switch estimator
    as a surrogate, which is correct up to a constant. The results
    reported are averaged over $10^4$ Monte Carlo
    trials. } \label{fig:competitive-plug-in}
\end{figure}

\bigskip
Several remarks are in order. Note that the upper bound on the
competitive ratio is at most $c \numaction$, achieved for a target
distribution that places mass on all $\numaction$ actions.  Comparing
this worst-case guarantee with that of the ``always return zero''
estimator~\eqref{eq:competitive-ratio-zero-estimator} shows that
plug-in estimator is strictly better as soon as the sample size
$\numobs$ exceeds a multiple of $\numaction$.  Note that this is a
relatively mild condition on the sample size.  In addition,
Theorem~\ref{thm:reg-competitive-ratio} guarantees worst-case
competitive ratio of the plug-in estimator scales linearly with the
support size of $\pitarget$. This showcases the automatic adaptivity
of the plug-in estimator to the target policy under consideration. See
Figure~\ref{fig:competitive-plug-in} for a numerical illustration of
this phenomenon.

We note that Li~et~al.~\cite{li2015toward} established a similar
guarantee (see Theorem 3 in their paper~\cite{li2015toward}).  One
importance difference is that their guarantee only holds in the large
sample regime (cf.~the
restriction~\eqref{eq:sample-size-requirement}), whereas ours covers
the full spectrum of the sample size.  Moreover, their upper bound is
proportional to $\numaction$ for any target policy, and so does not
reveal the adaptivity of the plug-in estimator to the support size.


\subsubsection{Is the plug-in estimator optimal?}

A natural follow-up question is to investigate the optimality of 
the plug-in approach---in the sense of the worst-case competitive
ratio---in the unknown $\pibehave$ case.
It turns out that, the plug-in estimator is close to optimal,
 as demonstrated by the following theorem.

\begin{theorem}
  \label{thm:ratio-lower-bound}
  Suppose that the sample size is lower bounded as $\numobs \geq
  \plaincon \frac{\numaction}{\log \numaction}$ for a positive
  constant $c$. Then for each $s \in \{ 1, 2, \ldots, \numaction \}$,
  there exists a target policy $\pitarget$ supported on $s$ actions
  and
  \begin{align}
    \label{EqnRatioLower}
\inf_{\Vhat} \sup_{\pibehave,\RewardDistPl \in \RewardFamily}
\CompRat
  \big(\Vhat; \pitarget, \pibehave, \RewardDistPl \big)
   & \geq \plaincon' \, \max \left\{ \frac{s}{\log
  \numaction}, 1 \right\},
  \end{align}
where $\plaincon' > 0$ is a universal constant.
\end{theorem}
\noindent See
Section~\ref{sec:Proof-of-Theorem-lower-bound-competitive-ratio} for
the proof of Theorem~\ref{thm:ratio-lower-bound}.

\bigskip
As shown in the proof of Theorem~\ref{thm:ratio-lower-bound}, for each
given integer $s \in [\numaction]$, the lower
bound~\eqref{EqnRatioLower} is met by taking the target policy that
chooses actions uniformly from the set $\{1, 2, \ldots, s\}$.  This
lower bound shows that when evaluating a policy with support size $s$,
the gap---between performance when knowing the behavior policy
$\pibehave$ relative to not knowing it---scales as $s$ up to a
logarithmic factor; thus, these two settings are very different in
terms of their statistical difficulty.  In addition, comparing the
lower bound in Theorem~\ref{thm:ratio-lower-bound} with the upper
bound provided in Theorem~\ref{thm:reg-competitive-ratio}, one can see
that the plug-in estimator $\VhatPlug$ is optimal up to a logarithmic
factor, measured by the worst-case competitive ratio.


\subsection{OPE with lower bounds on the minimum exploration probability}
\label{subsec:partial-knowledge}

The preceding subsections consider two extreme cases in which the
behavior policy is either known or completely unknown.  This leaves us
with an interesting middle ground: what if we have some partial
knowledge regarding the behavior policy?  How can such information be
properly exploited by estimators?

In this section, we initiate the investigation of these questions by
focusing on a particular type of partial knowledge---namely, the
\emph{minimum exploration probability} $\min_{\action
  \in \ActionSpace} \pibehave(\action)$.  More precisely, for a given
scalar $\plower \geq 0$, consider the collection of distributions
\begin{align*}
\PiFamily(\plower ) & \coloneqq \big \{ \pi \mid \min_{\action
  \in \ActionSpace} \pi(\action) \geq \plower \}.
\end{align*}
Given that any randomized policy $\pi$ must sum to one, i.e.,
$\sum_{\action \in \ActionSpace} \pi(\action) = 1$, this family is
non-empty only when $\plower \in [0, 1/\numaction]$.  Our goal in this
section is to characterize the difficulty of the OPE problem when it
is known that $\pibehave \in \PiFamily(\plower)$ for some choice of
$\plower$.  We first analyze the plug-in estimator, which does
\emph{not} require knowledge of $\plower$.  We then derive a minimax
lower bound, which shows that the plug-in estimator is
\mbox{sub-optimal} for certain choices of $\plower$.  In the end, we
design an alternative estimator, based on approximation by Chebyshev
polynomials, that has optimality guarantees for a large family of
target policies, albeit under a different but closely related Poisson
sampling model.

\subsubsection{Performance of the plug-in estimator}

We begin with establishing a performance guarantee for the plug-in
estimator.

\begin{theorem}
\label{thm:reg-eta}
There exist universal constants $\plaincon, \cprime > 0$ such that
\begin{subequations}
\begin{align}
  \label{eq:reg-eta-case-1}
  \sup_{(\pibehave, \RewardDistPl) \in \PiFamily(\plower) \times \RewardFamily}
\Exp_{\pibehave \otimes
  \RewardDistPl}[(\VhatPlug-\Vphi(\pitarget))^{2}] \leq \plaincon
\left \{ \Rmax^{2}\cdot\exp(- 2 \numobs \plower ) +
\Risk_{\numobs}^{\star}(\pitarget; \pibehave) \right \}.
\end{align}
In addition, if $\plower \geq \tfrac{\log \numaction}{\numobs}$, then
we have
\begin{align}
  \label{eq:reg-eta-case-2}
  \sup_{(\pibehave, \RewardDistPl) \in \PiFamily(\plower) \times
    \RewardFamily} \Exp_{\pibehave \otimes
    \RewardDistPl}[(\VhatPlug-\Vphi(\pitarget))^{2}] \leq \cprime
  \Risk_{\numobs}^{\star}(\pitarget; \pibehave).
\end{align}
\end{subequations}
\end{theorem}
\noindent See Section~\ref{sec:Proof-of-Theorem-reg-eta} for the proof of these two claims.

\bigskip
Two interesting observations are worth making.  First, if the behavior
policy is \emph{sufficiently exploratory}, in the sense that it
belongs to the family $\PiFamily(\plower)$ for some $\plower \geq
\tfrac{\log \numaction}{\numobs}$, then the plug-in estimator
$\VhatPlug$ achieves the optimal estimation error
$\Risk_{\numobs}^{\star}(\pitarget; \pibehave)$ up to a constant
factor.  In other words, the side condition $\min_{\action
  \in \ActionSpace} \pibehave(\action) \geq \tfrac{\log
  \numaction}{\numobs}$ is sufficient for the plug-in approach to
perform optimally.

On the other hand, when the behavior policy is less
exploratory---meaning that $\plower < \tfrac{\log
  \numaction}{\numobs}$---its mean-squared error involves the
additional term $\Rmax^{2}\cdot\exp(- 2 \numobs \plower)$.  As shown
in the proof of the upper bound~\eqref{eq:reg-eta-case-1}, this extra
price stems from bias of the plug-in estimator: if we fail to observe
rewards for some action $\action$, then the plug-in estimator has no
avenue for estimating the mean reward $\reward(\action)$; any estimate
that it makes incurs a bias of the order $\pitarget^2(\action) \cdot
\Rmax^2$. When $\pibehave(\action) = \plower$, such an event takes
place with probability on the order of $\exp(- \numobs \plower )$.

\subsubsection{Is the plug-in estimator optimal under partial knowledge?}
Is the extra price $\Rmax^{2}\cdot\exp(- 2 \numobs \plower )$
necessary for all estimators?  In order to answer this question, we
need to characterize the constrained minimax risk
\begin{align}
\label{eq:minimax-risk-eta}  
\Risk_{\mathsf{M}}^{\star} \left(\pitarget, \numobs, \plower \right)
\coloneqq \inf_{\Vhat} \sup_{(\pibehave, \RewardDistPl) \in
  \PiFamily(\plower ) \times \RewardFamily} \Exp_{\pibehave \otimes
  \RewardDistPl}[(\Vhat - \Vphi(\pitarget))^{2}],
\end{align}
where the supremum is taken over all possible behavior policies in
$\PiFamily(\plower)$, and reward distributions in $\RewardFamily$. In
view of our guarantee~\eqref{eq:reg-eta-case-2} for the plug-in
approach, it can be seen that when $\plower \geq \tfrac{\log
  \numaction}{\numobs}$, then
\begin{align*}
\Risk_{\mathsf{M}}^{\star} \left(\pitarget, \numobs, \plower \right) 
\asymp \sup_{\pibehave\in\PiFamily(\plower )}
\Risk_{\numobs}^{\star}(\pi; \pibehave),
\end{align*}
Consequently, the plug-in estimator is optimal when the behavior
policy is sufficiently exploratory.

As a result, in the remainder of this section, we concentrate on the
regime $\plower < \tfrac{\log \numaction}{\numobs}$.  We begin by
stating a minimax lower bound in this regime:
\begin{theorem}
\label{thm:lower-bound-eta} 
Consider the case $\plower < \tfrac{\log \numaction}{\numobs} $.  If
$\plower$ further satisfies $\plower \leq \frac{1}{2 \numaction}$ and
$\plower \geq \ccon \frac{1}{\numobs \log \numaction}$ for some
sufficiently large constant $\ccon > 0$, then there exists another
universal positive constant $\cprime$ such that
  \begin{align}
\label{eq:lower-bound-eta}    
\Risk_{\mathsf{M}}^{\star} \left(\pitarget, \numobs, \plower \right)
\geq \cprime \left \{ \sup_{\pibehave\in\PiFamily(\plower )}
\Risk_{\numobs}^{\star}(\pi; \pibehave) + \Rmax^{2} \cdot \exp(-200
\sqrt{\numobs \plower \log \numaction}) \right \}.
\end{align}
\end{theorem}

\noindent See Section~\ref{sec:Proof-of-Theorem-lower-bound-eta} for
the proof of this theorem. 

\bigskip
Note that if $\plower \lesssim \frac{1}{\numobs \log \numaction}$, the
worst-case risk is lower bounded as $\Omega(\Rmax^2)$.  Combining the
lower bound in Theorem~\ref{thm:lower-bound-eta} with the upper bound
shown in Theorem~\ref{thm:reg-eta}, we conclude that the plug-in
approach is minimax optimal up to constants once $\plower \gtrsim \log
\numaction / \numobs$. However, observe that there remains a gap
between the upper and lower bounds when the behavior policy is known
to be less exploratory---that is, in the regime $\frac{1}{\numobs \log
  \numaction} \lesssim \plower \ll \frac{\log \numaction}{\numobs}$.


\subsubsection{Optimal estimators via Chebyshev polynomials in the Poisson model}

In this section, we devote ourselves to the design of optimal
estimators when the behavior policy $\pibehave$ is less exploratory,
meaning that $\pibehave \in \PiFamily(\plower )$ for some
$\frac{1}{\numobs \log \numaction} \lesssim \plower \ll \frac{\log
  \numaction}{\numobs}$.

\paragraph{The Poisson model.}

In order to bring the key issues to the fore, we analyze the estimator
under the Poissonized sampling model that is standard in the
functional estimation literature. Recall that in the multinomial
observation model, the action counts $ \{ \numobs(\action), \action
\in [\numaction] \}$ follow a multinomial distribution with parameters
$\numobs$ and $\pibehave$. In the alternative Poisson model, the total
number of samples is assumed to be random, distributed according to a
Poisson distribution with parameter $\numobs$.  As a result, the
action counts obey $\numobs(\action) \overset{\text{ind.}}{\sim}
\Poi(\numobs \pibehave(\action))$ for $\action \in \ActionSpace$.
Correspondingly, we can define the minimax risk under the Poisson
model as
\begin{align}
\Risk_{\mathsf{P}}^{\star} (\pitarget, \numobs, \plower ) \coloneqq
\inf_{\Vhat} \sup_{(\pibehave, \RewardDistPl) \in \PiFamily(\plower )
  \times \RewardFamily} \Exp_{\pibehave \otimes \RewardDistPl}[(\Vhat
  - \Vphi(\pitarget))^{2}] ,
\end{align}
where the expectation is taken under the Poisson model. 

Although the two sampling models differ, the difference is not
actually essential in terms of characterizing minimax risks.  In
particular, the corresponding risks ${\Risk_{\mathsf{P}}^{\star}
  (\pitarget,n,\plower )}$ and ${\Risk_{\mathsf{M}}^{\star}
  (\pitarget,n,\plower )}$ are closely related, as demonstrated by the
following lemma.

\begin{lemma}
\label{lemma:risk-upper-bound-using-Poisson}
For any $\betapar \in (0,1)$, we have
\begin{align*}
\Risk_{\mathsf{M}}^{\star}(\pitarget, \numobs, \plower ) \leq
\frac{\Risk_{\mathsf{P}}^{\star} (\pitarget,(1-\betapar) \numobs,
  \plower )}{1 - \exp(-\numobs \betapar^{2}/2)}.
\end{align*}
\end{lemma}

\noindent
See Appendix~\ref{sec:proof-risk-upper-bound-Poisson} for the proof of
this bound.

\bigskip
Setting $\betapar = 1/2$ in the above lemma reveals that
$\Risk_{\mathsf{M}}^{\star}(\pitarget, \numobs, \plower ) \lesssim
\Risk_{\mathsf{P}}^{\star} (\pitarget,\tfrac{1}{2} \numobs,\plower)$.
Consequently, in order to obtain an upper bound on the risk under
multinomial sampling, it suffices to control the risk under the
Poisson model.

\paragraph{The Chebyshev estimator.}

Now we turn to the construction of the optimal estimator under the
Poisson model. It turns out that Chebyshev polynomials play a central
role in such a construction.  Recall that the Chebyshev polynomial
with degree $\cdeg$ is given by
\begin{align}
\label{def:chebyshev-poly}
\ChebPoly(x) & \coloneqq \cos(\cdeg \arccos x)
\end{align}
for $x \in [-1, 1]$.  Correspondingly, for any pair of scalars such
that $\myright > \myleft > 0$, we can define a shifted and scaled
polynomial
\begin{align*}
\ScaledCheby_{\cdeg}(x) & = - \frac{\ChebPoly \left(\frac{2x-\myright
    - \myleft}{\myright
    -\myleft}\right)}{\ChebPoly\left(\frac{-\myright-\myleft}{\myright
    - \myleft}\right)} \eqqcolon\sum_{d=0}^{\cdeg} \coeff_{d} x^{d},
\end{align*}
where $\coeff_{d}$ denotes the coefficient of $x^{d}$.  Using the
coefficients of this polynomial as a building block, we then define a
function, with domain the set of nonnegative integers, given by
\begin{align*}
g_{\cdeg}(j) & = \begin{cases} \coeff_{j} \frac{j!}{\numobs^{j}}+1, &
  \mbox{for $j = 0, 1, \ldots, \cdeg$, and} \\
  1 & \mbox{ if $j > \cdeg$.}
\end{cases}
\end{align*}
In terms of these quantities, the \emph{Chebyshev estimator} takes the
form
\begin{align}
\label{def:chebyshev-estimator}
\VhatCheby & \coloneqq \sum_{\action \in \ActionSpace}
\pitarget(\action)\rhat(\action) g_{\cdeg}(\numobs(\action)),
\end{align}
where $\rhat(\action)$ is the empirical mean reward defined in
equation~\eqref{eq:empirical-reward}.  In words, when the action count
$\numobs(\action)$ is larger than the degree $\cdeg$, one uses the
usual sample mean reward $\rhat(\action)$. On the other hand, when the
action count $\numobs(\action)$ is below this threshold, the Chebyshev
estimator rescales the empirical mean reward by the value
$g_{\cdeg}(\numobs(\action))$.  The goal of this rescaling is to
reduce the bias of the plug-in estimate.

A little calculation helps to provide intuition regarding this
bias-reduction effect.  Under the Poisson sampling model, the biases of the
Chebyshev estimator and the plug-in estimator are given by
\begin{align*}
\Exp [ \VhatCheby ] - \Vphi(\pitarget) &= \sum_{\action
  \in \ActionSpace} \pitarget(\action) \reward(\action) e^{- \numobs
  \pibehave(\action)} \ScaledCheby_{\cdeg} (\pibehave(\action)), \quad \mbox{and}
\\
\Exp [ \VhatPlug ] - \Vphi(\pitarget) &= \sum_{\action
  \in \ActionSpace} \pitarget(\action) \reward(\action) e^{- \numobs
  \pibehave(\action)},
\end{align*}
respectively. For the plug-in estimator, if we allow the behavior
policy $\pibehave$ to range over the family $\PiFamily(\plower)$, then
the bias can be as large as $\Rmax \cdot \exp(-\numobs \plower)$.

By construction, the Chebyshev polynomial $\ScaledCheby_{\cdeg}$ is
the unique degree-$\cdeg$ polynomial such that
$\ScaledCheby_{\cdeg}(0) = -1$, and that is closest in sup norm to the
all-zeros function on the interval $[\myleft, \myright]$; see
Exercise 2.13.14 in the book~\cite{timan2014theory}. By suitable
choices of the triple $(\myleft, \myright, \cdeg)$, we can shape the
additional modulation factor $\ScaledCheby_{\cdeg}
(\pibehave(\action))$ so as to reduce the bias of the plug-in
estimator.  The following theorem makes this intuition precise:
\begin{theorem}
\label{thm:upper-bound-eta}
Suppose that the target policy satisfies the bound
\begin{align}
\label{eq:target-constraint}
\sum_{\action \in \ActionSpace}\pitarget^{2}(\action) \leq
\frac{1}{\numaction^\decaycon} \qquad \mbox{for some scalar $\decaycon
  > 0$,}
\end{align}
and that we implement the Chebyshev estimator with
\begin{align*}
\myleft \coloneqq \plower,\quad \myright \coloneqq \ccon_{1} \log
\numaction / \numobs, \quad \mbox{and} \quad \cdeg = \ccon_{0}\log
\numaction,
\end{align*}
for some sufficiently large constant $\ccon_{1} > 0$ and some constant
$\plaincon_{0}\leq \decaycon / 7$. Then under the Poisson sampling
model, there exists a pair of positive constants $(\ccon, \cprime)$
such that
\begin{align}
\label{eq:risk-chebyshev}
\sup_{(\pibehave, \RewardDistPl) \in \PiFamily(\plower ) \times
  \RewardFamily} \Exp \Big[ \big( \VhatCheby - \Vphi(\pitarget)
  \big)^{2} \Big] & \leq \ccon \left \{ \Rmax^2 \exp (- \cprime
\sqrt{\numobs \plower \log \numaction }) +
\sup_{\pibehave\in\PiFamily(\plower )}
\Risk_{\numobs}^{\star}(\pitarget; \pibehave) \right \}.
\end{align}
\end{theorem}
\noindent See Section~\ref{sec:Proof-of-Theorem-upper-bound-eta} for
the proof of this upper bound.

\bigskip
Several comments are in order. First, when $\plower < \log \numaction
/ \numobs$, the worst-case risk~\eqref{eq:risk-chebyshev} of the
Chebyshev estimator~\eqref{def:chebyshev-estimator} matches the lower
bound~\eqref{eq:lower-bound-eta} derived in
Theorem~\ref{thm:lower-bound-eta}, which showcases the optimality of
the Chebyshev estimator when the partial knowledge $\plower$ is
available.

Second, the restriction~\eqref{eq:target-constraint} on the evaluation
policy is worth emphasizing. In words, the
constraint~\eqref{eq:target-constraint} requires the target policy to
be somewhat ``de-localized''---that is, there is no action $\action$
that has an extremely large probability mass. As an example, the
uniform policy on $\ActionSpace$ satisfies such a constraint.

Third, it should be noted that the Chebyshev estimator requires
knowledge of the minimum exploration probability $\plower$.  This
property makes it less practically applicable a priori, and how to
design an estimator that can adapt to the nested family of behavior
policies $\{ \PiFamily(\plower) \}_{0 < \plower \leq 1/\numaction}$ is
an interesting question for future work.

\subsubsection{Numerical experiments}
\label{SecExperimentsforchebyshev}

We conclude this section with experiments on both simulated data and
real data to assess the performance of the Chebyshev estimator
relative to other choices.

\begin{figure}[h]
  \centering \includegraphics[scale = 0.7]{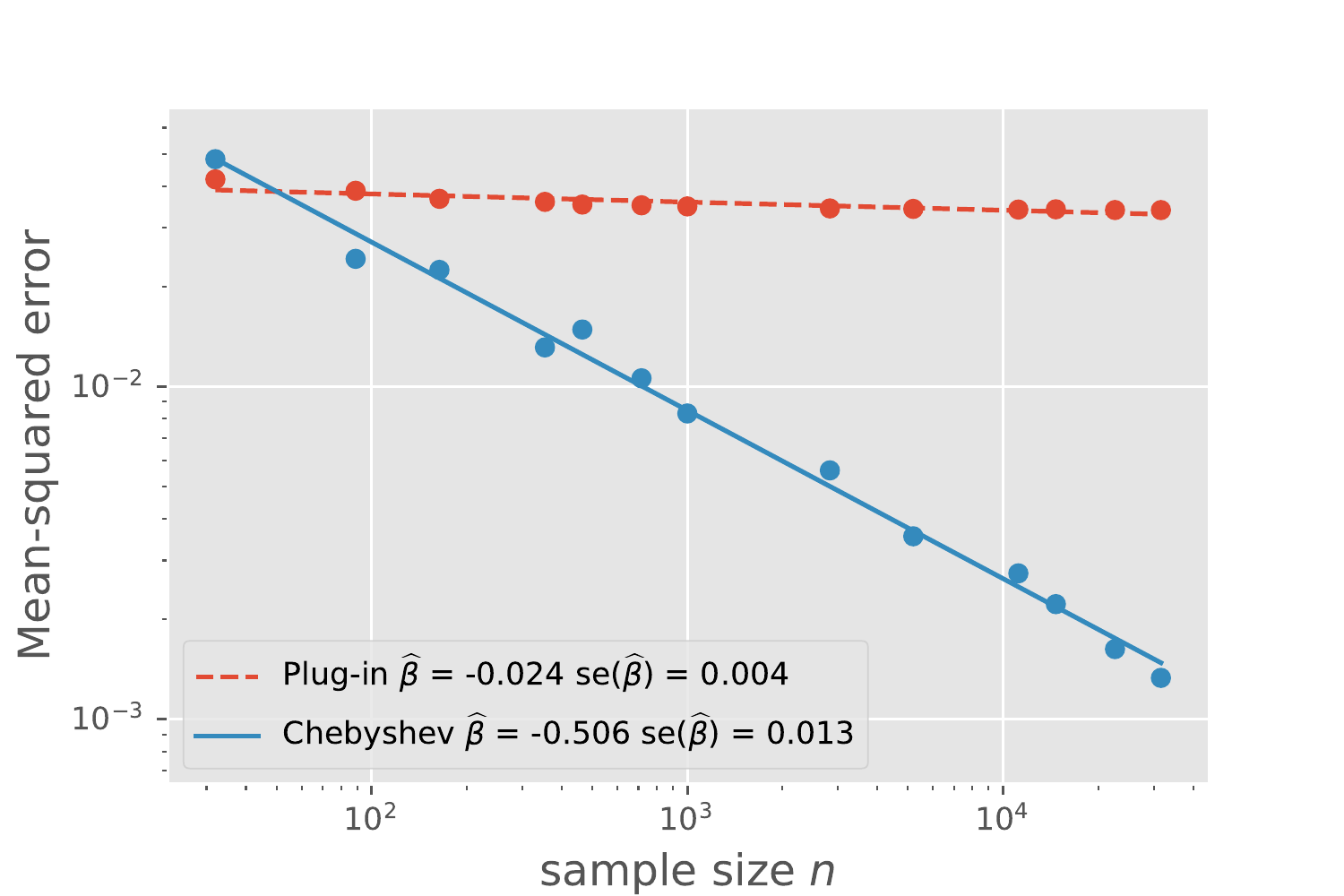}
  \caption{Log-log plot of the estimation errors vs.~the sample
    size. All results reported are averaged over $10^4$ random
    trials. The legends also contain the estimated slopes obtained by
    performing a linear regression of the log error on the log sample
    size.}
      \label{fig:chebyshev-simulated}
\end{figure}

\paragraph{Simulated data.}

We begin with some experiments on simulated data.  As in our previous
simulations, we fix the target policy to be uniform over
$\ActionSpace$, and for each action $\action \in \ActionSpace$, we
choose the reward distribution $\RewardDist{\action}$ to be an
equi-probable Bernoulli distribution over $\{0, 1\}$, so that $\Rmax =
1$.  For each $\numaction$, we define the behavior policy
\begin{align*}
\pibehave(1) = 1 - \frac{\numaction-1}{\numaction^{1.5}},\quad
\text{and}\quad \pibehave(2) = \pibehave(3) = \cdots =
\pibehave(\numaction) = \frac{1}{\numaction^{1.5}}.
\end{align*}

Again, we consider a particular scaling of the pair $(\numobs,
\numaction)$ that highlights interesting differences.  In particular,
when the sample size $\numobs$ scales as $\numobs = \numaction^{1.5}$,
then our theory predicts that the plug-in and Chebyshev estimators
should have mean-squared error scaling as
\begin{align*}
\Exp_{\pibehave \otimes \RewardDistPl} [(\VhatPlug -
  \Vphi(\pitarget))^{2}] & \asymp 1, \qquad \text{and} \\
\Exp_{\pibehave \otimes \RewardDistPl} [(\VhatCheby -
  \Vphi(\pitarget))^{2}] & \asymp \numobs^{-\delta} \quad \mbox{for some
  $\delta > 0$},
\end{align*}
respectively.  Figure~\ref{fig:chebyshev-simulated} plots the
mean-squared errors of the two estimators vs.~the sample size
$\numobs$ on a log-log scale. The results are averaged over $10^4$
random trials. It is clear from Figure~\ref{fig:chebyshev-simulated}
that the Chebyshev estimator performs better than the one based on the
plug-in principle. Based on the estimated slopes (as shown in the
legend), the mean-squared error of the Chebyshev estimator decays as
$\numobs^{-1/2}$, while consistent with our theory, that of the
plug-in estimator nearly plateaus.

\paragraph{Real data.}

We now turn to some experiments with the MovieLens 25M data
set~\cite{harper2015movielens}.  In order to form a bandit problem, we
extracted a random subset of $500$ movies that each have at least $10$
ratings.  This subset of movies defines an action space $\ActionSpace$
with $\numaction = 500$. For each movie, we average its rating over
all samples in order to define the mean reward $\reward(\action)$
associated with the movie $\action$.  This is the ground truth that
defines our problem instance.  Setting the target policy to be
uniform, our goal is to estimate the mean rating of these $500$
movies---that is, the quantity $\sum_{\action \in [\numaction]}
\reward(\action)/\numaction$.
  
In order to evaluate our methods, we need to generate an off-policy
dataset.  In order to do so, we uniformly subsample $\numobs$ ratings
from the set of all ratings on our subset of $500$ movies.  This
procedure implicitly defines a behavior policy that is very different
from the uniform target policy, because the number of ratings for each
movie vary drastically.  Given such an off-policy dataset, we evaluate
the mean-squared errors of four different estimators---the plug-in
estimator, the IS estimator, the Switch estimator as well as the
Chebyshev estimator.  We repeat this procedure for a total of $10^4$
trials for a range of sample sizes $\numobs$.

\begin{figure}[t]
    \centering
    \includegraphics[scale = 0.7]{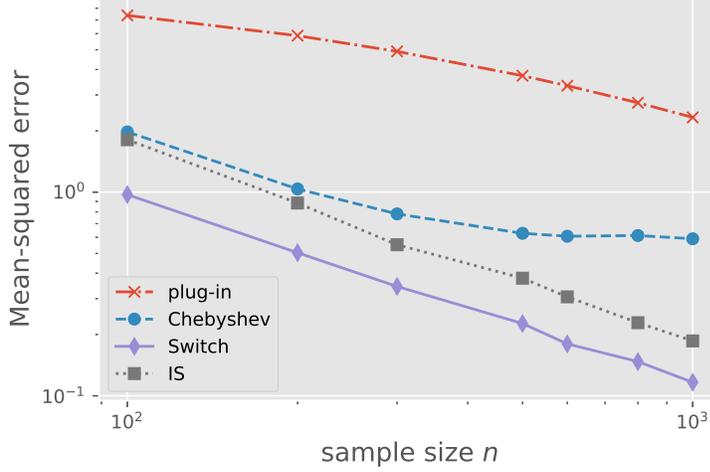}
    \caption{Mean-squared errors of four different estimators vs.~the
      sample size $\numobs$ on the MovieLens 25M data set. The results
      are averaged over $10^4$ trials.  See the text for further
      details on the experimental set-up.}
    \label{fig:sim_real}
\end{figure}

Figure~\ref{fig:sim_real} plots the mean-squared error (averaged over
the trials) versus the sample size $\numobs$ for the four estimators.
To be clear, the Switch estimator and the IS estimator have the luxury
of knowing the behavior policy whereas the Chebyshev estimator is
given minimum exploration probability.  The plug-in estimator requires
no side information.  Given the oracle knowledge of the behavior
policy, the Switch estimator always outperforms other estimators,
including the IS estimator with the same knowledge. In addition, the
Chebyshev estimator outperforms plug-in estimator, especially in the
small sample regime. These qualitative changes are consistent with our
theoretical predictions.

%

\section{Proofs}

We now turn to the proofs of the main results presented in
Section~\ref{sec:main-results}.  We begin in
Section~\ref{SecProof:prop:switch-general} with the proof of
Proposition~\ref{prop:switch-general}, followed by the proof of
Proposition~\ref{prop:optimal-subset} in
Section~\ref{sec:proof-of-optimal-subset}. Sections~\ref{subsec:proof-of-theorem-switch}
through~\ref{sec:Proof-of-Theorem-upper-bound-eta} are devoted to the
proofs of Theorems~\ref{thm:main} through \ref{thm:upper-bound-eta}.


\subsection{Proof of Proposition~\ref{prop:switch-general}}
\label{SecProof:prop:switch-general}

We begin with the standard bias-variance decomposition
\begin{align}
  \label{EqnBiasVariance}
\Exp_{\pibehave \otimes \RewardDistPl}[(\VhatSwitch(\ActionSubset) -
  \Vphi(\pitarget))^{2}] & = \left(\Exp_{\pibehave \otimes
  \RewardDistPl} [\VhatSwitch(\ActionSubset) ]-
\Vphi(\pitarget)\right)^{2} + \Var
\left(\VhatSwitch(\ActionSubset)\right).
\end{align}
Our proof involves establishing the following two bounds
\begin{subequations}
  \begin{align}
    \label{EqnBiasBound}
\left(\Exp_{\pibehave \otimes
  \RewardDistPl}[\VhatSwitch(\ActionSubset)]-\Vphi(\pitarget)\right)^{2} &
\leq \Rmax^{2}\pitarget^{2}(\ActionSubset), \quad \mbox{and} \\
    \label{EqnVarBound}
\Var \left(\VhatSwitch(\ActionSubset)\right) & \leq 2 \Rmax^{2} \left
\{ \pitarget^{2}(\ActionSubset) + \frac{\sum_{\action \notin
    \ActionSubset}\pibehave(\action)\likerat^{2}(\action)}{\numobs}
\right \}.
\end{align}
\end{subequations}
The claim of the proposition follows by substituting these bounds into the
bias-variance decomposition~\eqref{EqnBiasVariance}.

\paragraph{Proof of the bias bound~\eqref{EqnBiasBound}:}
Using the shorthand $\Exp$ for the expectation over $\pibehave \otimes
\RewardDistPl$, we have
\begin{align*}
\Exp [\VhatSwitch(\ActionSubset)] &
= \sum_{\action \in \ActionSubset} \pitarget(\action)
\Exp[\rhat(\action)]+\Exp[\likerat(\Action_{i})\Reward_{i} \indicator
  \{\Action_{i} \notin \ActionSubset\}] \\
& = \sum_{\action \in \ActionSubset}
\pitarget(\action)\Exp[\rhat(\action)]+\sum_{\action \notin
  \ActionSubset} \pitarget(\action) \reward(\action).
\end{align*}
Recalling the definition~\eqref{eq:OPE-obj} of $\Vphi(\pitarget)$, we have
\begin{align*}
\left(\Exp [\VhatSwitch(\ActionSubset)]-\Vphi(\pitarget)\right)^{2} =
\left(\sum_{\action \in \ActionSubset}\pitarget(\action)\left\{
\Exp[\rhat(\action)]- \reward(\action)\right\} \right)^{2}\leq
\Rmax^{2}\pitarget^{2}(\ActionSubset),
\end{align*}
where the final inequality follows from the bound $\big|
\Exp[\rhat(\action)]- \reward(\action) \big| \leq \Rmax$.

\paragraph{Proof of the variance bound~\eqref{EqnVarBound}:}
Using the inequality \mbox{$\Var(X+Y) \leq 2 \Var(X) +
  2\Var(Y)$,} we have
\begin{align*}
\Var \left(\VhatSwitch(\ActionSubset)\right) & \leq 2 \Var
\left(\sum_{\action \in \ActionSubset} \pitarget(\action)
\rhat(\action) \right) + 2 \Var \left(\frac{1}{\numobs}
\sum_{i=1}^{\numobs}\likerat(\Action_{i}) \Reward_{i} \indicator
\{\Action_{i} \notin \ActionSubset\}\right).
\end{align*}
The first term can be bounded as
\begin{align*}
\Var \left(\sum_{\action \in
  \ActionSubset}\pitarget(\action)\rhat(\action)\right)\leq\Exp\left[\left(\sum_{\action
    \in \ActionSubset}\pitarget(\action)\rhat(\action)\right)^{2}
  \right]\leq \Rmax^{2}\left(\sum_{\action \in
  \ActionSubset}\pitarget(\action) \right)^{2} = \Rmax^{2}
\pitarget^{2}(\ActionSubset),
\end{align*}
where the penultimate relation arises from the fact that
$|\rhat(\action)|\leq \Rmax$.  With regards to the variance
brought by importance sampling, one has
\begin{align*}
\Var \left(\frac{1}{\numobs}
\sum_{i=1}^{\numobs} \likerat(\Action_{i}) \Reward_{i} \indicator
\{\Action_{i} \notin \ActionSubset\}\right) &
=\frac{1}{\numobs} \Var \left(\likerat(\Action_{i})\Reward_{i} \indicator
\{\Action_{i} \notin \ActionSubset\}\right) \\
& \leq \frac{1}{\numobs} \Exp[\likerat^{2}(\Action_{i})
  \Reward_{i}^{2} \indicator \{\Action_{i} \notin \ActionSubset\}] \\ &
\leq \Rmax^{2}\frac{\Exp[\likerat^{2}(\Action_{i}) \indicator
    \{\Action_{i} \notin \ActionSubset\}]}{\numobs} \\ & = \Rmax^{2}
\frac{\sum_{\action \notin \ActionSubset}
  \pibehave(\action)\likerat^{2}(\action)}{\numobs},
\end{align*}
where the last inequality uses the fact that 
$|\Reward_{i}| \leq \Rmax$.
Combining the two terms yields the claimed variance bound~\eqref{EqnVarBound}.


\subsection{Proof of Proposition~\ref{prop:optimal-subset}}
\label{sec:proof-of-optimal-subset}

Recall that the subset $\SetStar$ corresponds to the support set of
the solution $\vb^{\star}$ to the convex program~\eqref{eq:key-opt}.
Here we state an important connection between the objective value of
the problem~\eqref{eq:key-opt} and this support set:
\begin{lemma}
\label{lemma:minimum-value}
We have
\begin{align}
\label{eq:intermediate-goal}
\min_{\vb \in \real^{\numaction}} \left \{ \sqrt{ \frac{1}{8 \numobs}
  \sum_{\action \in \ActionSpace} \frac{[\pitarget(\action) -
      v(\action)]^{2}}{\pibehave(\action)}} + \frac{1}{2}
\sum_{\action \in \ActionSpace} |v(\action)| \right \} \asymp
\pitarget(\SetStar) + \sqrt{\frac{\sum_{\action \notin \SetStar}
    \pibehave(\action) \likerat^{2}(\action)}{\numobs}},
\end{align}
where $\asymp$ denotes equality up to a universal constant.
\end{lemma}
\noindent See Appendix~\ref{sec:proof-of-lemma-min-value} for the
proof of this lemma.

In light of the equivalence~\eqref{eq:intermediate-goal} as well as
the sandwich bound $x^2 + y^2 \leq (x+y)^2 \leq 2x^2 + 2y^2$ for any
$x, y > 0$, proving the bound~\eqref{eq:min-upper-bound} reduces to
establishing the lower bound
\begin{align}
\label{eq:min-upper-bound-proof}  
\min_{\ActionSubset \subseteq \ActionSpace} \left \{
\frac{1}{2} \pitarget(\ActionSubset) + \sqrt{\frac{\sum_{\action \notin
      \ActionSubset} \pibehave(\action)
    \likerat^{2}(\action)}{ 8 \numobs}} \right \} & \geq  
    \min_{\vb \in \real^{\numaction}} \left \{ \sqrt{ \frac{1}{8 \numobs}
  \sum_{\action \in \ActionSpace} \frac{[\pitarget(\action) -
      v(\action)]^{2}}{\pibehave(\action)}} + \frac{1}{2}
\sum_{\action \in \ActionSpace} |v(\action)| \right \}.
\end{align}
Letting $\xi(\action) \in \{0,1 \}$ denote a binary indicator variable
for the event $\{ \action \in \ActionSubset \}$, the optimization
problem on the left hand side is equivalent to
\begin{align*}
\pstar & = \min_{ \bm{\xi} \in \real^{\numaction}} \left \{
\frac{1}{2} \sum_{\action \in \ActionSpace}
\pitarget(\action)\xi(\action)+\sqrt{\frac{\sum_{\action
      \in \ActionSpace} \pibehave(\action) \likerat^{2}(\action)
    (1-\xi(\action)) }{8 \numobs}} \right \} \quad
\mbox{s.t. $\xi(\action) \in \{0,1 \}$ for all $\action
  \in \ActionSpace$.}
\end{align*}
By relaxing to the requirement $\xi(\action) \in [0,1]$, we obtain a
convex lower bound
\begin{align*}
\pstar & \geq \min_{\bm{\xi} \in \real^{\numaction}} \left
\{\frac{1}{2} \sum_{\action
  \in \ActionSpace}\pitarget(\action)\xi(\action)+\sqrt{\frac{\sum_{\action
      \in \ActionSpace}\pibehave(\action)\likerat^{2}(\action)(1-\xi(\action))}{8
    \numobs}} \right \} \quad \mbox{s.t.  $\xi(\action) \in [0,1]$ for
  all $\action \in \ActionSpace$.}
\end{align*}
Since the the inclusion $\xi(\action) \in [0,1]$ guarantees that
$1-\xi(\action)\geq(1-\xi(\action))^{2}$, we can further relax
to obtain the lower bound
\begin{align*}
\pstar & \geq \min_{\bm{\xi} \in \real^{\numaction} } \left
\{\frac{1}{2} \sum_{\action
  \in \ActionSpace}\pitarget(\action)\xi(\action)+\sqrt{\frac{\sum_{\action
      \in \ActionSpace}\pibehave(\action)\likerat^{2}(\action)(1-\xi(\action))^{2}}{8
    \numobs}} \right \} \quad \mbox{s.t. $\xi(\action) \in [0,1]$ for
  all $\action \in \ActionSpace$.}
\end{align*}
Now we are ready to prove the claimed
bound~\eqref{eq:min-upper-bound-proof}.  Applying the change of
variables \mbox{$\xi(\action)=v(\action)/\pitarget(\action)$,} we can
transform the problem above into
\begin{align*}
  \min_{\vb} & \left \{ \frac{1}{2} \sum_{\action \in \ActionSpace}
  v(\action) + \sqrt{ \frac{1}{8 \numobs}\sum_{\action
      \in \ActionSpace} \frac{ [\pitarget(\action) -
        v(\action)]^{2}}{\pibehave(\action)} } \right \}.
\end{align*}
subject to the constraints $v(\action) \in [0, \pitarget(\action)]$
for all $\action \in \ActionSpace$.  In fact, following some simple
calculations, one can see that it is equivalent to the following
unconstrained problem
\begin{align}
\label{eq:key-opt-replicate}
\min_{\vb \in \real^{\numaction}} \left \{ \sqrt{ \frac{1}{8 \numobs}
  \sum_{\action \in \ActionSpace} \frac{[\pitarget(\action) -
      v(\action)]^{2}}{\pibehave(\action)}} + \frac{1}{2}
\sum_{\action \in \ActionSpace} |v(\action)| \right \}
\end{align}
Note that this is identical to the lower bound we aim at 
in equation~\eqref{eq:min-upper-bound-proof}, and hence the proof
is finished.


\subsection{Proof of Theorem~\ref{thm:main}}
\label{subsec:proof-of-theorem-switch}

Our proof is based on using Le~Cam's method to lower bound the minimax
risk.  In doing so, a key step is to construct two similar reward
distributions $\RewardDistPl_{1}, \RewardDistPl_{2} \in \RewardFamily$
such that the absolute distance $|V_{\RewardDistPl_{1}}(\pi) -
V_{\RewardDistPl_{2}}(\pi)|$ is large.

For each action $\action$, let $\RewardDistPl_{2}(\, \cdot \mid
\action)$ be a Bernoulli distribution on the set $\{0, \Rmax\}$ with
parameter $\tfrac{1}{2}$, let $\RewardDistPl_{1}(\,\cdot\mid a)$ be a
Bernoulli distribution over $\{0, \Rmax\}$ with parameter
$\tfrac{1}{2} + \delpar(\action)$ for some $\delpar(\action) \in [0,
  \tfrac{1}{2}]$.  From Le~Cam's inequality, we have the lower bound
\begin{align*}
\inf_{\Vhat} \sup_{\RewardDistPl \in \RewardFamily} \Exp_{\pibehave
  \otimes \RewardDistPl} [(\Vhat - \Vphi(\pitarget))^{2}] \geq
\frac{1}{8} (V_{\RewardDistPl_{1}}(\pi) - V_{\RewardDistPl_{2}}
(\pi))^{2} e^{-n \Kull{\pibehave \otimes \RewardDistPl_{1}}{\pibehave
    \otimes \RewardDistPl_{2}}}.
\end{align*}
With this choice of $\RewardDistPl_{1}$ and $\RewardDistPl_{2}$, one
can verify that
\begin{align*}
\Kull{\pibehave \otimes \RewardDistPl_{1}}{\pibehave \otimes
  \RewardDistPl_{2}} = \sum_{\action \in \ActionSpace}
\pibehave(\action) \Kull{\RewardDistPl_{1}}{\RewardDistPl_{2}} \leq 4
\sum_{\action \in \ActionSpace} \pibehave(\action)\delpar^{2}(\action),
\end{align*}
where the inequality arises from the relation
$\Kull{\mathsf{Bern}(\tfrac{1}{2}+\delpar(\action))}{\mathsf{Bern}(\tfrac{1}{2})}
\leq 4 \delpar^{2}(\action)$.  In addition, it is easily seen that
\begin{align*}
(V_{\RewardDistPl_{1}}(\pi)-V_{\RewardDistPl_{2}}(\pi))^{2} =
  \tfrac{1}{4} \Rmax^{2} \Big( \sum_{\action \in \ActionSpace}
  \pitarget(\action) \delpar(\action) \Big)^{2}.
\end{align*}
Therefore, we can obtain a lower bound on the minimax risk---one that
is optimal within this particular family---by solving the optimization
problem
\begin{align}
\label{eq:lower-bound-opt}   
\max_{\bm{\delpar} \in\real^{\numaction}} & \quad \Rmax^{2} \Big(
\sum_{\action \in \ActionSpace} \pitarget(\action) \delpar(\action)
\Big)^{2} \quad \mbox{subject to $4 \sum \limits_{\action
    \in \ActionSpace} \pibehave(\action) \delpar^{2}(\action) \leq
  \frac{1}{2 \numobs}$ and $\delpar(\action) \in [0, \tfrac{1}{2}]$
  for all $\action \in \ActionSpace$.}
\end{align}

First, we make the observation that the optimization
problem~\eqref{eq:lower-bound-opt} is equivalent to the following
optimization problem in the sense that they share the same minimizer
and the minimum values are in a one-to-one correspondence
\begin{align}
\label{eq:lower-bound-opt-primal}   
\max_{\bm{\delpar} \in \real^{\numaction}} \left \{ \sum_{\action
  \in \ActionSpace} \pitarget(\action) \delpar(\action) \right \}
\quad \mbox{such that $\sum \limits_{\action \in \ActionSpace}
  \pibehave(\action) \delpar^{2}(\action) \leq \tfrac{1}{8 \numobs}$
  and $\delpar(\action) \in [0, \tfrac{1}{2}]$ for all $\action
  \in \ActionSpace$.}
\end{align}

Note that this is a convex problem with quadratic constraints. We find
it easier to look at its dual formulation, which is supplied in the
following lemma.

\begin{lemma}
\label{lemma:duality-lower-bound}
The Fenchel dual problem of the optimization
program~\eqref{eq:lower-bound-opt-primal} is given by
\begin{align}
\label{eq:lower-bound-opt-dual}
\min_{\vb \in \real^{\numaction}} \quad \left \{ \sqrt{\frac{1}{8
    \numobs} \sum_{\action \in \ActionSpace} \frac{ [
      \pitarget(\action) - v(\action)]^{2}}{\pibehave(\action)}} +
\frac{1}{2} \sum_{\action \in \ActionSpace} |v(\action)| \right \},
\end{align}
which shares the same optimal objective value as that of the
problem~\eqref{eq:lower-bound-opt-primal}.
\end{lemma}
\noindent See Appendix~\ref{sec:proof-duality-lower-bound} for the
proof of this result. \\

Fortunately, the minimum value of the dual
program~\eqref{eq:lower-bound-opt-dual} has been characterized in
Lemma~\ref{lemma:minimum-value}, namely
\begin{align*}
\min_{\vb \in \real^{\numaction}} \left \{ \sqrt{ \frac{1}{8 \numobs}
  \sum_{\action \in \ActionSpace} \frac{[\pitarget(\action) -
      v(\action)]^{2}}{\pibehave(\action)}} + \frac{1}{2}
\sum_{\action \in \ActionSpace} |v(\action)| \right \} \asymp
\pitarget(\SetStar) + \sqrt{\frac{\sum_{\action \notin \SetStar}
    \pibehave(\action) \likerat^{2}(\action)}{\numobs}},
\end{align*}
where $\asymp$ denotes equality up to universal constants. 

Taking the preceding equivalence relationships collectively and use
the elementary relation $(x + y)^2 \asymp x^2 +y^2$, we can arrive at
the desired conclusion
\begin{align*}
\inf_{\Vhat} \sup_{\RewardDistPl \in \RewardFamily} \Exp_{\pibehave
  \otimes \RewardDistPl}[(\Vhat - \Vphi(\pitarget))^{2}] & \geq
\plaincon \: \Rmax^{2}\left\{ \pitarget^{2}(\SetStar) +
\frac{\sum_{\action \notin \SetStar} \pibehave(\action)
  \likerat^{2}(\action)}{\numobs} \right\}
\end{align*} 
with $c$ a universal positive constant.


\subsection{Proof of Theorem~\ref{thm:reg-competitive-ratio}}
\label{sec:Proof-of-Theorem-reg-competitive-ratio}

The mean-squared error of $\VhatPlug$ can be decomposed as
\begin{align}
\Exp_{\pibehave \otimes
  \RewardDistPl}[(\VhatPlug-\Vphi(\pitarget))^{2}] &
=\Big(\sum_{\action \in \ActionSpace}\pitarget(\action)
\reward(\action)(1-\pibehave(\action))^{\numobs}
\Big)^{2}+\sum_{\action
  \in \ActionSpace}\pitarget^{2}(\action)\sigma_{\RewardDistPl}^{2}(\action)\Exp\left[
  \tfrac{ \indicator \{\numobs(\action)>0\}}{\numobs(\action)}\right] \nonumber\\ &
\quad + \Var \Big(\sum_{\action \in \ActionSpace}\pitarget(\action)
\reward(\action) \indicator \{\numobs(\action)>0\}
\Big). \label{eq:MSE-plug-in}
\end{align}
See Appendix A.6 of the paper~\cite{li2015toward} for the calculations
underlying this decomposition.  Here the first term represents the
squared bias of $\VhatPlug$, while the remaining two correspond to the
variance of $\VhatPlug$.

Fix any behavior policy $\pibehave$, and let $\SetStar$ be defined as
in equation~\eqref{eq:defn-S-star}. Our proof consists of upper
bounding the squared bias and variance in terms of functions of
$\Risk_{\numobs}^{\star}(\pitarget; \pibehave)$.  More precisely, we
prove the following two bounds
\begin{subequations}
\label{subeq:MSE-plug-in}
\begin{align}
\label{eq:bias-plug-in}  
\Big(\sum_{\action \in \ActionSpace}\pitarget(\action)
\reward(\action)(1-\pibehave(\action))^{\numobs} \Big)^{2}  \leq
\plaincon \; |\mysupp(\pitarget)|\, \Risk_{\numobs}^{\star}(\pitarget;
\pibehave), \quad \mbox{and} \\
\label{eq:variance-plug-in}
\sum_{\action \in \ActionSpace}\pitarget^{2}
(\action)\sigma_{\RewardDistPl}^{2}(\action)\Exp\left[\tfrac{ \indicator
    \{\numobs(\action)>0\}}{\numobs(\action)}\right] + \Var
\Big(\sum_{\action \in \ActionSpace} \pitarget(\action)
\reward(\action) \indicator \{\numobs(\action)>0\} \Big) \leq \plaincon'
\; \Risk_{\numobs}^{\star}(\pitarget; \pibehave),
\end{align}
\end{subequations} 
for some universal constants $(\plaincon, \plaincon')$.  Since the
bounds~\eqref{subeq:MSE-plug-in} hold for any $\pibehave$, the desired
conclusion follows by combining the above two bounds.  In the sequel,
we focus on establishing the bounds~\eqref{subeq:MSE-plug-in}.


\paragraph{Proof of the bias bound~\eqref{eq:bias-plug-in}:}
Beginning with the squared bias, we have
\begin{align}
\Big( \sum_{\action \in \ActionSpace} \pitarget(\action)
\reward(\action) (1 - \pibehave(\action))^{\numobs} \Big)^{2}
& \overset{\text{(i)}}{\leq} \Rmax^{2} \Big( 
 \sum_{\action \in \ActionSpace}
\pitarget(\action) (1 - \pibehave(\action))^{\numobs} \Big)^{2} \nonumber \\ 
& \overset{\text{(ii)}}{\leq}
2 \Rmax^{2} \Big \{ \Big(\sum_{\action \in \SetStar}
\pitarget(\action) (1 - \pibehave(\action))^{\numobs} \Big)^{2} + 
\Big( \sum_{\action \in \SetStarComp}
\pitarget(\action) (1 - \pibehave(\action))^{\numobs} \Big)^{2} \Big\} \nonumber \\
 & \overset{\text{(iii)}}{\leq} 2 \Rmax^{2} 
\Big\{ \pitarget^{2}(\SetStar) + \Big(\sum_{\action
  \in \SetStarComp}
\pitarget(\action) (1 - \pibehave(\action))^{\numobs} \Big)^{2} \Big\}
. \label{eq:bias-plug-in-reuse}
\end{align}
Here the first inequality (i) arises from the assumption 
$|\reward(\action)| \leq \Rmax$, the second relation (ii) applies the inequality
$(a+b)^2 \leq 2(a^2 + b^2)$, and the last one (iii) uses the fact
$(1-\pibehave(\action))^{\numobs} \leq 1$.

Applying the Cauchy--Schwarz inequality yields
\begin{align*}
\big(\sum_{\action \in \SetStarComp} \pitarget(\action)
(1-\pibehave(\action))^{\numobs} \big)^{2} & = \big( \sum_{\action \in
  \SetStarComp \cap \mysupp(\pitarget)} \pitarget(\action)
(1-\pibehave(\action))^{\numobs} \big)^{2} \\
& \leq | \SetStarComp \cap \mysupp(\pitarget) | \left(\sum_{\action
  \in \SetStarComp}
\pitarget^{2}(\action)(1-\pibehave(\action))^{2\numobs} \right) \\ &
\leq | \mysupp(\pitarget) | \sum_{\action \in \SetStarComp}
\pitarget^{2}(\action) (1-\pibehave(\action))^{\numobs} \\
& \leq | \mysupp(\pitarget) | \sum_{\action \in
  \SetStarComp}\pitarget^{2}(\action)
\frac{1}{\numobs\pibehave(\action)},
\end{align*}
where the last inequality follows from the bound $(1 -
\pibehave(\action))^{\numobs} \leq 1/(\numobs \pibehave(\action))$.
Combining the preceding two bounds with our previous
expression~\eqref{eq:minimax-rate-known-pi} for
$\Risk_{\numobs}^{\star}(\pitarget; \pibehave)$ yields the claimed
bound~\eqref{eq:bias-plug-in}.


\paragraph{Proof of the variance bound~\eqref{eq:variance-plug-in}:}

We now move onto the two variance terms.  Since
$\sigma_{\RewardDistPl}^{2}(\action)\leq \Rmax^{2}$, we can write
\begin{align*}
\sum_{\action \in \ActionSpace} \pitarget^{2}(\action)
\sigma_{\RewardDistPl}^{2}(\action) \Exp \left[ \tfrac{ \indicator \{
    \numobs(1) > 0\}}{\numobs(\action)} \right] & \leq \Rmax^{2} \left
\{ \sum_{\action \in \SetStar} \pitarget^{2}(\action) \Exp
\left[\tfrac{ \indicator \{\numobs(\action) > 0 \}}{
    \numobs(\action)}\right] + \sum_{\action \notin \SetStar}
\pitarget^{2}(\action) \Exp \left[ \tfrac{ \indicator \{\numobs(\action)
    > 0 \} }{\numobs(\action)} \right] \right\}
\end{align*}
Since $\Exp \left[ \frac{ \indicator \{\numobs(\action) > 0\}}{
    \numobs(\action)} \right] \leq 1$, the first term on the
right-hand side can be upper bounded as
\begin{align*}
\sum_{\action \in \SetStar} \pitarget^{2}(\action) \Exp
\left[\tfrac{ \indicator \{\numobs(\action) > 0 \}}{
    \numobs(\action)}\right] & \leq
\sum_{\action \in \SetStar} \pitarget^{2}(\action).
\end{align*}
From Lemma 1 of the paper~\cite{li2015toward}, we have the bound $\Exp
\left[ \frac{ \indicator \{ \numobs(\action) > 0 \}}{\numobs(\action)}
  \right] \leq \frac{5}{\numobs \pibehave(\action)}$, which implies
the second term is upper bounded as
\begin{align*}
\sum_{\action \notin \SetStar} \pitarget^{2}(\action) \Exp \left[
  \tfrac{ \indicator \{\numobs(\action) > 0 \} }{\numobs(\action)}
  \right] & \leq \frac{5}{\numobs} \sum_{\action \notin \SetStar}
\pitarget^{2}(\action) \frac{1}{\pibehave(\action)}.
\end{align*}
By combining these two inequalities, we conclude that
\begin{align}
  \label{EqnHanaSleep}
\sum_{\action \in \ActionSpace} \pitarget^{2}(\action)
\sigma_{\RewardDistPl}^{2}(\action) \Exp \left[ \tfrac{ \indicator \{
    \numobs(1) > 0\}}{\numobs(\action)} \right] & \leq \plaincon \;
\Risk_{\numobs}^{\star}(\pitarget; \pibehave),
\end{align}
for a universal constant $\plaincon$.

Turning to the second quantity in the variance, we have
\begin{align*}
\Var \left( \sum_{\action \in \ActionSpace}\pitarget(\action)
\reward(\action) \indicator \{\numobs(\action) > 0\} \right) & \leq
\sum_{\action \in \ActionSpace} \pitarget^{2}(\action)
\reward^{2}(\action) (1 - \pibehave(\action))^{\numobs} \\
& \leq \Rmax^{2} \left\{ \sum_{\action \in \SetStar}
\pitarget^{2}(\action) (1-\pibehave(\action))^{\numobs} +
\sum_{\action \notin \SetStar} \pitarget^{2}(\action)
(1-\pibehave(\action))^{\numobs} \right\} \\
& \stackrel{(i)}{\leq} \Rmax^{2} \left\{ \sum_{\action \in \SetStar}
\pitarget^{2}(\action) + \sum_{\action \notin \SetStar}
\pitarget^{2}(\action) \frac{1}{n\pibehave(\action)} \right\} \\
& \leq \plaincon \Risk_{\numobs}^{\star} (\pitarget; \pibehave),
\end{align*}
where step (i) follows from the elementary inequalities $(1 -
\pibehave(\action))^{\numobs} \leq 1$ and $(1 -
\pibehave(\action))^{\numobs} \leq 1/( \numobs \pibehave(\action))$.
Combining this bound with our earlier inequality~\eqref{EqnHanaSleep}
yields the claimed bound~\eqref{eq:variance-plug-in}.


\subsection{Proof of Theorem~\ref{thm:ratio-lower-bound}}
\label{sec:Proof-of-Theorem-lower-bound-competitive-ratio}

We first show that the competitive ratio is lower bounded by $1$.
From the elementary inequality $\inf \sup \geq \sup \inf$, we see that
\begin{align*}
\inf_{\Vhat} \sup_{\pibehave,\RewardDistPl \in \RewardFamily} \frac{
  \Exp_{\pibehave \otimes \RewardDistPl} [(\Vhat -
    \Vphi(\pitarget))^{2}]} {\Risk_{\numobs}^{\star}(\pitarget;
  \pibehave)} \geq \sup_{\pibehave} \inf_{\Vhat} \sup_{\RewardDistPl
  \in \RewardFamily} \frac{ \Exp_{\pibehave \otimes \RewardDistPl}
  [(\Vhat - \Vphi(\pitarget))^{2}]}
    {\Risk_{\numobs}^{\star}(\pitarget; \pibehave)} \stackrel{(i)}{=}
    1,
\end{align*}
where equality (i) follows from the
definition~\eqref{eq:defn-R-n-star} of
$\Risk_{\numobs}^{\star}(\pitarget; \pibehave)$.

The remainder of our analysis is to prove a lower bound in
terms of $s/\log \numaction$.  Throughout this analysis, we consider a
target distribution $\pitarget$ that is uniform over $\{1,2,\ldots,
s\}$, and we consider the set
\begin{align*}
\PiFam(\plower) \coloneqq \{ \pibehave \, \mid \, \min_{\action \in
  \ActionSpace} \pibehave(\action) \geq \plower \}, \quad \mbox{where
  $\plower \coloneqq 1 / (\numobs \log \numaction)$.}
\end{align*}
Since $\PiFam(\plower)$ is a subset of all behavior policies, we have
\begin{align}
\inf_{\Vhat} \sup_{\pibehave,\RewardDistPl \in \RewardFamily} \frac{
  \Exp_{\pibehave \otimes \RewardDistPl} [(\Vhat -
    \Vphi(\pitarget))^{2}]} {\Risk_{\numobs}^{\star}(\pitarget;
  \pibehave)} & \geq \inf_{\Vhat} \sup_{\pibehave \in \PiFam(\plower
  )} \frac{1}{\Risk_{\numobs}^{\star} (\pitarget; \pibehave)}
\sup_{\RewardDistPl \in \RewardFamily} \Exp_{\pibehave \otimes
  \RewardDistPl} [(\Vhat - \Vphi(\pitarget))^{2}] \nonumber \\
\label{eq:lower-bound-ratio}
& \geq \frac{\inf_{\Vhat} \sup_{\pibehave \in \PiFam(\plower )}
  \sup_{\RewardDistPl \in \RewardFamily} \Exp_{\pibehave \otimes
    \RewardDistPl} [(\Vhat-\Vphi(\pitarget))^{2}]} {\sup_{\pibehave
    \in \PiFam(\plower )} \Risk_{\numobs}^{\star}(\pitarget;
  \pibehave)},
\end{align}
where in the second line we use the trivial upper bound
$\Risk_{\numobs}^{\star}(\pitarget; \pibehave) \leq
\sup_{\pibehave\in\PiFam(\plower )}\Risk_{\numobs}^{\star}(\pitarget;
\pibehave)$ for any $\pibehave\in\PiFam(\plower )$.

In view of the lower bound~\eqref{eq:lower-bound-ratio}, the proof can
be decomposed into two steps: (1) Provide a lower bound on the
numerator; and (2) Establish an upper bound on the denominator.

\paragraph{Step 1:}
The required lower bound on the numerator in
equation~\eqref{eq:lower-bound-ratio} can be obtained by applying
Theorem~\ref{thm:lower-bound-eta}.  More specifically, doing so with
$\plower = 1/(n\log \numaction)$ yields
\begin{align}
\inf_{\Vhat}\sup_{\pibehave\in\PiFam(\plower )}\sup_{\RewardDistPl \in
  \RewardFamily}
\Exp_{\pibehave \otimes \RewardDistPl}[(\Vhat - \Vphi(\pitarget))^{2}] \gtrsim
\Rmax^{2}.\label{eq:ratio-proof-lower-bound}
\end{align}

\paragraph{Step 2:}
We now turn to the upper bound on the denominator of
equation~\eqref{eq:lower-bound-ratio}.  By combining
Lemma~\ref{lemma:minimum-value} with the
characterization~\eqref{eq:minimax-rate-known-pi} of
$\Risk_{\numobs}^{\star}(\pitarget; \pibehave)$, we find that
\begin{align*}
\left[ \Risk_{\numobs}^{\star}(\pitarget;
  \pibehave)/\Rmax^{2}\right]^{1/2} \asymp \underset{\vb \in
  \real^{\numaction}}{\inf} \quad \left \{ \sqrt{\frac{1}{8
    \numobs}\sum_{\action \in \ActionSpace} \frac{[\pitarget(\action)
      - v(\action)]^{2}}{\pibehave(\action)}} + \frac{1}{2}
\sum_{\action \in \ActionSpace} | v(\action) | \right \},
\end{align*}
which further implies 
\begin{align*}
\sup_{\pibehave \in \PiFam(\plower )} \left[ 
\Risk_{\numobs}^{\star}(\pitarget; \pibehave) / \Rmax^{2} \right]^{1/2} & 
\asymp \sup_{\pibehave \in \PiFam(\plower )} 
\underset{\vb \in \real^{\numaction}}{\inf} \quad \left \{ 
\sqrt{\frac{1}{8 \numobs}\sum_{\action \in \ActionSpace}
\frac{[\pitarget(\action) - v(\action)]^{2}}{\pibehave(\action)}} 
+ \frac{1}{2} \sum_{\action \in \ActionSpace} | v(\action) | \right \} \\
 & \leq \underset{\vb \in \real^{\numaction}}{\inf}
 \sup_{\pibehave \in \PiFam(\plower )} \quad \left \{ 
 \sqrt{\frac{1}{8 \numobs} \sum_{\action \in \ActionSpace}
 \frac{[\pitarget(\action)-v(\action)]^{2}}{\pibehave(\action)}} 
 + \frac{1}{2}\sum_{\action \in \ActionSpace}|v(\action)| \right \}.
\end{align*}
Here, the last inequality arises from the elementary fact
$\sup\inf\leq\inf\sup$.  Focusing on the inner maximization problem,
we can easily see that
\begin{align*}
\sup_{\pibehave \in \PiFam(\plower)} \quad \sqrt{\frac{1}{8 \numobs}
  \sum_{\action \in \ActionSpace} \frac{[\pitarget(\action) -
      v(\action)]^{2}}{\pibehave(\action)}} + \frac{1}{2}
\sum_{\action \in \ActionSpace} |v(\action)| & \leq \sqrt{\frac{\log
    \numaction}{8} \sum_{\action \in \ActionSpace} [\pitarget(\action)
    - v(\action)]^{2}} + \frac{1}{2} \sum_{\action \in \ActionSpace}
|v(\action)|,
\end{align*}
where we substitute in the definition of $\plower$. Combine the
previous two bounds together to reach
\begin{align*}
\sup_{\pibehave \in \PiFam(\plower )}
\left[\Risk_{\numobs}^{\star}(\pitarget; \pibehave)/\Rmax^{2}
  \right]^{1/2} & \lesssim \underset{\vb \in \real^{\numaction}}{\inf}
\sup_{\pibehave \in \PiFam(\plower )} \left \{ \sqrt{\frac{\log
    \numaction}{8} \sum_{\action \in \ActionSpace}
  [\pitarget(\action)-v(\action)]^{2}}+\frac{1}{2}\sum_{\action
  \in \ActionSpace} |v(\action)| \right \} \\
& \leq \sqrt{\frac{\log \numaction}{8 s}},
\end{align*}
where the final inequality follows by setting $\vb = \zeros$ and using
the definition of $\pitarget$.  Consequently, we have established the
upper bound
\begin{align}
\label{eq:ratio-proof-upper-bound}  
\sup_{\pibehave \in \PiFam(\plower)}\Risk_{\numobs}^{\star}(\pitarget;
\pibehave)\lesssim\frac{\log \numaction}{s}\cdot \Rmax^{2}.
\end{align}
Combining equation~\eqref{eq:lower-bound-ratio} with
equations~\eqref{eq:ratio-proof-lower-bound},
and~\eqref{eq:ratio-proof-upper-bound} yields the desired conclusion.


\subsection{Proof of Theorem~\ref{thm:reg-eta}}
\label{sec:Proof-of-Theorem-reg-eta}

Recall the decomposition~\eqref{eq:MSE-plug-in} of the mean-squared
error incurred by the plug-in estimator:
\begin{subequations}
\begin{multline}
  \label{eq:MSE-plug-in-copy}
\Exp_{\pibehave \otimes \RewardDistPl}
    [(\VhatPlug-\Vphi(\pitarget))^{2}] =\Big( \sum_{\action
      \in \ActionSpace}\pitarget(\action) \reward(\action) (1 -
    \pibehave(\action))^{\numobs} \Big)^{2}+\sum_{\action
      \in \ActionSpace}\pitarget^{2}(\action)
    \sigma_{\RewardDistPl}^{2}(\action)\Exp\left[ \tfrac{ \indicator
        \{\numobs(\action)>0\}}{\numobs(\action)}\right] \\
    + \Var \Big(\sum_{\action
      \in \ActionSpace}\pitarget(\action) \reward(\action) \indicator
    \{\numobs(\action) > 0 \} \Big). 
\end{multline}
As shown in equation~\eqref{eq:variance-plug-in}, the variance
components are well-behaved in the sense that
\begin{align}
\label{eq:variance-plug-in-copy}
\sum_{\action \in \ActionSpace}\pitarget^{2}
(\action)\sigma_{\RewardDistPl}^{2}(\action)\Exp\left[\tfrac{
    \indicator \{\numobs(\action)>0\}}{\numobs(\action)}\right] + \Var
\Big(\sum_{\action \in \ActionSpace} \pitarget(\action)
\reward(\action) \indicator \{\numobs(\action)>0\} \Big) \leq
\plaincon' \; \Risk_{\numobs}^{\star}(\pitarget; \pibehave)
\end{align}
for some universal constant $\plaincon'$.  As a result, we only need
to focus on the squared bias term.  Corresponding to the statements in
Theorem~\ref{thm:reg-eta}, we split the proof into two cases: (1)
$\plower $ is arbitrary, and (2) $\plower \geq \log \numaction /
\numobs$.

\paragraph{Case 1:}  We begin with the case of an arbitrary $\plower$.
Since $\min_{\action} \pibehave(\action) \geq \plower$, one has $(1 -
\pibehave(\action))^{\numobs} \leq \exp(-\numobs \pibehave(\action))
\leq \exp(-\numobs \plower)$. This combined with the fact that
$|\reward(\action)| \leq \Rmax$ yields
\begin{align}
\label{eq:bias-plug-in-case-1}
\Big(\sum_{\action \in \ActionSpace}\pitarget(\action)
\reward(\action)(1-\pibehave(\action))^{\numobs}
\Big)^{2} \leq \Rmax^2 \Big ( \sum_{\action \in \ActionSpace} 
\pitarget(\action) \exp(-\numobs \plower) \Big )^{2} = 
\Rmax^2 \exp(-2 \numobs \plower),
\end{align}
\end{subequations}
where the last equality arises from the fact that $\sum_{\action
  \in \ActionSpace} \pitarget(\action) = 1$.  Combining the
bounds~\eqref{eq:MSE-plug-in-copy}, \eqref{eq:variance-plug-in-copy}
and~\eqref{eq:bias-plug-in-case-1} yields the claimed
bound~\eqref{eq:reg-eta-case-1}.

\paragraph{Case 2:} Now suppose that $\plower$ is lower bounded as
$\plower \geq \log \numaction / \numobs$.  By applying the bias
bound~\eqref{eq:bias-plug-in-reuse}, we find that
\begin{subequations}
\begin{align}
\label{eq:bias-plug-in-reuse-copy}
  \Big( \sum_{\action \in \ActionSpace} \pitarget(\action)
  \reward(\action) (1 - \pibehave(\action))^{\numobs} \Big)^{2} & \leq
  2 \Rmax^{2} \Big\{ \pitarget^{2}(\SetStar) + \Big(\sum_{\action \in
    \SetStarComp} \pitarget(\action) (1 -
  \pibehave(\action))^{\numobs} \Big)^{2} \Big\}.
\end{align}
We then apply the Cauchy--Schwarz inequality to obtain
\begin{align}
\Big(\sum_{\action \in \SetStarComp} \pitarget(\action) (1 -
\pibehave(\action))^{\numobs} \Big)^{2} & \leq \numaction \cdot
\sum_{\action \in \SetStarComp} \pitarget^2 (\action) (1 -
\pibehave(\action))^{2 \numobs} \nonumber \\ & \leq \sum_{\action \in
  \SetStarComp} \pitarget^2 (\action) (1 -
\pibehave(\action))^{\numobs} \nonumber \\ & \leq \sum_{\action \in
  \SetStarComp} \pitarget^2 (\action) \frac{1}{\numobs
  \pibehave(\action)}.
\label{eq:bias-plug-in-case-2}
\end{align}
\end{subequations}
Here, the middle line follows from $(1 - \pibehave(\action))^{\numobs} 
\leq \exp(-\numobs \pibehave(\action))
\leq \exp(-\numobs \plower) \leq 1/\numaction$, and the last 
inequality holds since $(1 - \pibehave(\action))^{\numobs} 
\leq \frac{1}{\numobs \pibehave(\action)}$. Combining the 
bounds~\eqref{eq:MSE-plug-in-copy}, \eqref{eq:variance-plug-in-copy}, 
\eqref{eq:bias-plug-in-reuse-copy}, \eqref{eq:bias-plug-in-case-2} 
with the expression~\eqref{eq:minimax-rate-known-pi} of 
$\Risk_{\numobs}^{\star}(\pitarget; \pibehave)$, we arrive 
at the desired bound~\eqref{eq:reg-eta-case-2}. 


\subsection{Proof of Theorem~\ref{thm:lower-bound-eta}}
\label{sec:Proof-of-Theorem-lower-bound-eta}

Without loss of generality, we may assume that the actions are ordered
such that $\pitarget(1) \geq \pitarget(2) \geq \cdots \geq
\pitarget(\numaction)$.  First, observe that
\begin{align}
\Risk_{\mathsf{M}}^{\star} \left(\pitarget, \numobs, \plower \right) & =
\inf_{\Vhat}\sup_{\pibehave\in\PiFam(\plower
  ),\RewardDistPl \in \RewardFamily}\Exp_{\pibehave \otimes
  \RewardDistPl}[(\Vhat - \Vphi(\pitarget))^{2}] \nonumber \\
& \stackrel{(\text{i})}{\geq} \sup_{\pibehave \in \PiFam(\plower
  )}\inf_{\Vhat}\sup_{\RewardDistPl \in \RewardFamily}\Exp_{\pibehave \otimes
  \RewardDistPl}[(\Vhat - \Vphi(\pitarget))^{2}]\nonumber \\
\label{eq:naive-lower-bound-eta}
& \stackrel{(\text{ii})}{=} \sup_{\pibehave \in \PiFam(\plower)}
\Risk_{\numobs}^{\star}(\pitarget; \pibehave),
\end{align}
where the inequality (i) follows from the fact that
$\inf\sup \geq \sup\inf$, whereas the equality (ii) uses the
definition~\eqref{eq:defn-R-n-star} of
$\Risk_{\numobs}^{\star}(\pitarget; \pibehave)$. This yields the first term in the lower bound~\eqref{eq:lower-bound-eta}. In particular, if
\begin{align}
\label{eq:case-1}
\pitarget^{2}(1) \exp (-2 \numobs \plower ) \geq\exp(-200\sqrt{
  \numobs \plower \log \numaction}),
\end{align}
then the bound~\eqref{eq:naive-lower-bound-eta} tells us that
\begin{align*}
\Risk_{\mathsf{M}}^{\star}\left(\pitarget,n,\plower \right) &
\geq\sup_{\pibehave\in\PiFam(\plower
  )}\Risk_{\numobs}^{\star}(\pitarget;
\pibehave)\overset{(\text{i})}{\geq}\frac{1}{4}\Rmax^{2}\pitarget^{2}(1)(1-\plower
)^{\numobs}\overset{(\text{ii})}{\geq}\frac{1}{4}\Rmax^{2}\pitarget^{2}(1)\exp(-2
\numobs\plower ) \\
&
\stackrel{(\text{iii})}{\geq}\frac{1}{4}\Rmax^{2}\exp(-200\sqrt{\numobs
  \plower \log \numaction}).
\end{align*}
Here, the first relation (i) uses Theorem~1 in the
paper~\cite{li2015toward}, the second inequality (ii) uses the
elementary relation $(1-\plower )^{\numobs}\geq e^{-2 \numobs\plower
}$ for $\plower \in [0, \tfrac{1}{2}]$, while the last one (iii)
arises from the condition~\eqref{eq:case-1}.  In words, when the
largest mass $\pitarget(1)$ in $\{\pitarget(\action)\}$ is
sufficiently large (cf.~the condition~\eqref{eq:case-1}), the term
$\sup_{\pibehave\in\PiFam(\plower )}\Risk_{\numobs}^{\star}(\pitarget;
\pibehave)$ dominates $\Rmax^{2}\cdot\exp(-200\sqrt{\numobs \plower
  \log \numaction})$, and hence the desired lower
bound~\eqref{eq:lower-bound-eta} follows.

Therefore, in the remaining part of this section, we concentrate on
establishing the lower bound for the case when
\begin{align}
\label{eq:case-2}
  \pitarget^{2}(1)\exp(-2 \numobs \plower) \leq \exp ( -200
  \sqrt{\numobs \plower \log \numaction}).
\end{align}
We record an immediate consequence of the relation~\eqref{eq:case-2} that will be useful later
\begin{align}
\label{eq:case-2-implication}  
\sum_{\action = 1}^{\numaction-1} \pitarget^{2}(\action) \leq
\pitarget(1) \leq \exp(\numobs \plower -100\sqrt{\numobs \plower \log
  \numaction}).
\end{align}
In order to prove lower bound in this case, it is convenient to
make use of a Poissonized model.

\paragraph{The Poissonized model.}
Recall that in the multinomial observation model, $(\numobs(1),
\numobs(2), \ldots, \numobs(\numaction))$ follows a multinomial
distribution with parameters $\numobs$ and $\pibehave$.  Here the
dependence among the counts $\{\numobs(\action)\}_{\action
  \in \ActionSpace}$ complicates the analysis. In order to sidestep
this dependence, we consider the Poisson model in which each action
$\action$ is taken with $\numobs(\action)$ times with
$\numobs(\action)\sim \Poi(\numobs \pibehave(\action))$ independently
across actions. Then conditional on the count $\numobs(\action)=t$, a
total of $t$ rewards are observed independently for each action
$a$. Note that in the Poissonized model, $\{\numobs(\action)\}$ are
mutually independent by design, which greatly facilitates the
analysis. Another difference that is worth pointing out is that in the
original multinomial model, $\sum_{\action
  \in \ActionSpace}\pibehave(\action)$ must sum to $1$, while in the
Poisson case, this restriction does not necessarily hold. To account
for this fact, we define the following $\varepsilon$-relaxed
probability simplex for a parameter $0 \leq \varepsilon \leq 1$:
\begin{align}
\label{eq:defn-relax-set}  
\LocalParaSpace(\plower ,\varepsilon) \coloneqq \Big\{
\pibehave \geq \zeros \mid \min_{\action
  \in \ActionSpace} \pibehave(\action) \geq \plower , \quad \Big
|\sum_{\action \in \ActionSpace} \pibehave(\action)-1\Big
|\leq\varepsilon \Big\}.
\end{align}
Correspondingly we can define the minimax risk over the relaxed
parameter set under the Poisson model as
\begin{align}
\label{eq:lower-bound-eta-Poisson-eps}  
\Risk_{\mathsf{P}}^{\star} \left(\pitarget, \numobs, \plower, \varepsilon
\right)\coloneqq\inf_{\Vhat} \sup_{\pibehave \in \Theta(\plower,
  \varepsilon), \RewardDistPl \in \RewardFamily} 
  \Exp_{\pibehave \otimes \RewardDistPl}
      [(\Vhat - \Vphi(\pitarget))^{2}],
\end{align}
where we note that the expectation $\Exp_{\pibehave \otimes
  \RewardDistPl}[\cdot]$ is taken under the Poissonized model. It
turns out that the risks under these two models, i.e., the multinomial
model and the Poisson model, are closely related, as shown in the
following lemma.
\begin{lemma}
\label{lemma:risk-equivalence-Poisson-1}
For any $\varepsilon \in (0, 1/4)$, the following relation holds:
\begin{align*}
\Risk_{\mathsf{P}}^{\star} \left( \pitarget, n, \plower, 
\varepsilon\right) \leq
\Risk_{\mathsf{M}}^{\star}\left(\pitarget,n/2,\plower
/(1+\varepsilon)\right)+e^{-3n/72}\cdot \Rmax^{2}.
\end{align*}
\end{lemma}
\noindent See Appendix~\ref{subsec:Proof-of-auxiliary-lower-bound-eta}
for the proof of this claim. \\

\vspace*{0.1in}

\noindent Lemma~\ref{lemma:risk-equivalence-Poisson-1} shows that it
suffices to establish a good lower bound on
$\Risk_{\mathsf{P}}^{\star} \left( \pitarget, \numobs, \plower,
\varepsilon \right)$, and the remainder of our analysis focuses on
this sub-problem.

\paragraph{The Bernoulli reward model. }

Recall that $\RewardDistPl$ can be any reward distribution supported
on $[0,\Rmax]$.  For the purpose of the lower bound, we can restrict
our attention to Bernoulli reward models, in which each action is
associated with a Bernoulli reward distribution over $\{0,\Rmax\}$
with the parameter $r(\action)/\Rmax$. Here $0\leq r(\action)\leq
\Rmax$ is a parameter associated with the action $\action$. This
Bernoulli reward model, in conjunction with the Poisson sampling model
yields two observations $\{\PosCount_{\action}, \NegCount_{\action}\}$
for each action $\action$
\begin{align*}
\PosCount_{\action} \sim \Poi( \numobs \pibehave(\action) \tfrac{r(\action)}{\Rmax}),
\qquad \text{and} \qquad
\NegCount_{\action} \sim \Poi( \numobs \pibehave(\action) (1 - \tfrac{r(\action)}{\Rmax})).
\end{align*}
Here, $\PosCount_{\action}$ denotes the number of positive rewards (i.e., the rewards with value $\Rmax$) obtained for arm $\action$, while $\NegCount_{\action}$ denotes the number of ``negative'' rewards---meaning those with value $0$. We naturally have
\begin{align}
\Risk_{\mathsf{P}}^{\star}\left(\pitarget,n,\plower ,\varepsilon\right) & =
\inf_{\Vhat} \sup_{\pibehave\in\LocalParaSpace(\plower
  ,\varepsilon),\RewardDistPl \in \RewardFamily} \Exp_{\pibehave \otimes
  \RewardDistPl}[(\Vhat - \Vphi(\pitarget))^{2}]\nonumber \\
\label{eq:lower-bound-bernoulli-reward}
& \geq \inf_{\Vhat}\sup_{\pibehave\in\Theta(\plower ,\varepsilon),
  \{r(\action)\}}\Exp_{\pibehave \otimes \RewardDistPl}
      [(\Vhat - \Vphi(\pitarget))^{2}],
\end{align}
where the set of numbers $\{r(\action)\}$ dictates the Bernoulli
reward model $\RewardDistPl$.

\paragraph{A useful reparameterization.}

Now we introduce a reparameterization of the models introduced above.
Let us denote
\begin{align}
\label{eq:defn-eta-p-n}  
\plowerPos(\action) \coloneqq
\pibehave(\action) \tfrac{r(\action)}{\Rmax},
\qquad \text{and} \qquad \plowerNeg(\action) \coloneqq
\pibehave(\action)(1 - \tfrac{r(\action)}{\Rmax}).
\end{align}
Using this notation, we can translate our target $\Vphi(\pitarget)$
into
\begin{align*}
\Target( \thetabm ) \coloneqq \Vphi(\pitarget) = 
\sum_{\action \in \ActionSpace} \pitarget(\action) r(\action)
=\Rmax \cdot \sum_{\action
  \in \ActionSpace} \pitarget(\action) \frac{
  \plowerPos(\action)} {\plowerPos(\action) + \plowerNeg(\action)},
\end{align*}
where we denote by $ \thetabm \in\real^{2K}$ the collection of
parameters under the new parametrization, i.e.,
\begin{align*}
 \thetabm \coloneqq \big( \plowerPos(1),\plowerNeg(1),\plowerPos(2),
 \plowerNeg(2), \cdots, \plowerPos(\numaction),\plowerNeg(\numaction)
 \big) ^{\top}.
\end{align*}
Note that there is a one-to-one mapping between the original
parameterization $\{\pibehave(\action), r(\action)\}$ and the new one
$ \thetabm $. Hence, with an abuse of notation, we shall denote
\begin{align*}
\LocalParaSpace(\plower ,\varepsilon) \coloneqq \Big\{
 \thetabm \geq \zeros \mid \Big |\sum_{\action \in
 \ActionSpace } \plowerPos(\action) + 
 \plowerNeg(\action) - 1 \Big | \leq \varepsilon, \quad {
 \plowerPos(\action) + \plowerNeg(\action)} \geq \plower
,\quad\text{for all } \action \in \ActionSpace \Big\}.
\end{align*}
With this set of notation in place, the lower bound in
equation~\eqref{eq:lower-bound-bernoulli-reward} can be equivalently
written as
\begin{align}
\label{eq:lower-bound-Bernoulli-Poisson-1}  
\inf_{\Vhat}\sup_{ \thetabm \in \LocalParaSpace(\plower
  ,\varepsilon)} \Exp_{ \thetabm }[(\Vhat(\LocalObs) - 
  \Target( \thetabm ))^{2}],
\end{align}
where $\LocalObs \coloneqq (\PosCount_{1}, \NegCount_{1},\cdots, 
\PosCount_{\numaction}, \NegCount_{\numaction})^{\top}
\in [\numobs]^{2 \numaction}$
denotes the observations following the Poisson sampling and the Bernoulli reward models.

We intend to invoke Lemma~\ref{lemma:fuzzy-tsybakov-1} to obtain
a good lower bound. It all boils down to constructing two prior distributions
$\Prior{0}, \Prior{1}$ over the parameter space $\LocalParaSpace(\plower ,\varepsilon)$
such that the functional values $\Target( \thetabm )$ are well separated under different priors, while at the same time one cannot differentiate those two distributions based on the data alone.

\paragraph{A construction of two priors $\Prior{0},\Prior{1}$ over 
$\LocalParaSpace(\plower ,\varepsilon)$. }

The construction of the two priors hinges on the existence of two
random variables $\Xvar, \Xvar'$, introduced in the following lemma.

\begin{lemma}\label{lemma:existence-X-X'}There exist two random
variables $\Xvar, \Xvar'$ supported on $[0,1]$ such that 
\begin{subequations}\label{subeq:properties-of-X-X'}
  \begin{align}
 \label{eq:property-gap}     
\Exp \left[\frac{\Xvar}{\Xvar + \frac{\numobs \plower }{8 \log
      \numaction}}\right] - \Exp\left[\frac{\Xvar'}{\Xvar' + \frac{\numobs
      \plower }{8 \log \numaction}} \right] & \geq \frac{1}{2} \exp
\left(-96\sqrt{\numobs \plower \log \numaction} \right) \\
\Exp \left[\Xvar \right] = \Exp \left[\Xvar' \right] & =\frac{\numobs \plower
}{8 \log \numaction};\label{eq:property-mean} \\
\label{eq:property-moment}
\Exp\left[\Xvar^{j}\right] & = \Exp\left[(\Xvar')^{j}\right],\qquad\text{for
  all } 1 \leq j\leq \left\lceil 48 \log \numaction \right \rceil.
\end{align}
\end{subequations}
\end{lemma}
\noindent See Appendix~\ref{subsec:Proof-of-auxiliary-lower-bound-eta}
for the proof of this claim.

Now we are ready to construct two ``helper'' priors $\HelperPrior{0},
\HelperPrior{1}$ on $\LocalParaSpace(\plower ,\varepsilon)$.  Under both priors, we always set $\plowerNeg(\action) = \plower$ for $1\leq \action \leq \numaction - 1$, and $\plowerNeg(\numaction)=0$. 
Under $\HelperPrior{0}$, we let
$\Xvar_{1}, \cdots, \Xvar_{\numaction-1}$ be i.i.d.~copies of $\Xvar$, and set $\plowerPos(\action) = \frac{8\log \numaction}{\numobs} \cdot 
\Xvar_{\action}$ for $1\leq \action \leq \numaction - 1$, 
$\plowerPos(\numaction) = 1 - (\numaction - 1) \plower - (\numaction -
1)\frac{8\log \numaction}{\numobs}\Exp[\Xvar].$ Similarly, under
$\HelperPrior{1}$, we let $\Xvar_{1}',\cdots,\Xvar_{\numaction-1}'$ be i.i.d.~copies
of $\Xvar'$, and set $\plowerPos(\action)=\frac{8\log
  \numaction}{\numobs}\cdot \Xvar'_{\action}$ for $1\leq \action \leq \numaction - 1$,
$\plowerPos(\numaction) = 1 - (\numaction - 1)\plower -(\numaction -
1)\Exp[\Xvar']$.  It is straightforward to check that under both priors
$\HelperPrior{0}$ and $\HelperPrior{1}$
\begin{align*}
\plowerPos(\numaction)=1-(\numaction - 1)\plower -(\numaction - 1)\frac{8\log \numaction}{\numobs}\cdot\frac{\numobs \plower }{8\log \numaction}=1-2(\numaction - 1)\plower \geq \plower ,
\end{align*}
as long as $\plower \leq \frac{1}{2 \numaction}$. 
Finally, for $i\in\{0,1\}$, we define the prior $\Prior{i}$ to
be the push-forward measure of the restriction of $\HelperPrior{i}$ to the
following set: 

\begin{align}
\label{eq:restriction-set}  
\LocalSet_{i} \coloneqq \LocalParaSpace(\plower ,1/5) \cap \left\{  \thetabm \mid \big|
\Target( \thetabm ) - \Exp_{ \thetabm \sim \HelperPrior{i}}[ \Target( \thetabm )] \big |
\leq \frac{\exp (-96 \sqrt{\numobs \plower  \log \numaction})}{8} \cdot \Rmax
\right\}.
\end{align}


\paragraph{Application of Le Cam's method:} 

Now we are positioned to invoke Le Cam's method, in the form of
Lemma~\ref{lemma:fuzzy-tsybakov-1}, with the choices
\begin{align*}
\xi \coloneqq \frac{ \Exp_{\Prior{0}} [\Target( \thetabm)] +
  \Exp_{\Prior{1}}[\Target( \thetabm)]}{2}, \quad s \coloneqq \frac{
  \exp( -96 \sqrt{\numobs \plower \log \numaction})}{16} \cdot \Rmax,
\quad \mbox{and} \quad
\LocalParaSpace \coloneqq \Theta(\plower, 1/5).
\end{align*}
From the constructions of the priors $\Prior{0}$ and
$\Prior{1}$, we have
\begin{align*}
\Exp_{\Prior{0}} [\Target( \thetabm)] - \Exp_{\Prior{1}} [\Target(
  \thetabm)] & = \Rmax \cdot \left\{ \Exp_{\Prior{0}} \left[
  \sum_{\action \in \ActionSpace} \pitarget(\action) \frac{
    \plowerPos(\action)}{\plowerPos(\action) + \plowerNeg(\action)}
  \right]- \Exp_{\Prior{1}} \left[ \sum_{\action \in \ActionSpace}
  \pitarget(\action) \frac{\plowerPos(\action)}{\plowerPos(\action)+
    \plowerNeg(\action)} \right] \right\} \\
& = \Rmax \left(\sum_{ 1\leq \action \leq \numaction -
  1}\pitarget(\action)\right)\left\{ \Exp \left[ \frac{\Xvar}{\Xvar +
    \frac{\numobs \plower }{8\log \numaction}} \right] - \Exp \left[
  \frac{\Xvar'}{\Xvar'+\frac{\numobs \plower }{8\log
      \numaction}}\right]\right\} \\
& \geq \frac{1}{4} \Rmax \exp \left(-96 \sqrt{\numobs \plower \log
  \numaction}\right) = 4s,
\end{align*}
where the last inequality arises from the property of $\Xvar,\Xvar'$
(cf.~the inequality~\eqref{eq:property-gap}) as well as the fact that
$\sum_{1 \leq \action \leq \numaction - 1} \pitarget( \action ) \geq
1/2$. Consequently, $\Target( \thetabm) \leq \xi-s$ almost surely for
$ \thetabm \sim\Prior{1}$, and $\Target( \thetabm) \geq \xi+s $ almost
surely for $ \thetabm \sim\Prior{0}$, which immediately implies
$\beta_{0} =\beta_{1}=0$ in Lemma~\ref{lemma:fuzzy-tsybakov-1}.

It remains to control the total variation distance between $\Prior{0}$
and $\Prior{1}$. To begin with, denoting $\HelperPrior{i}' = 
\HelperPrior{i} \circ ( \thetabm ^{\otimes \numobs})^{-1}$
for $i\in\{0,1\}$, the triangle inequality gives
\begin{align*}
\TVdist(\Prior{0}, \Prior{1}) & \leq 
\TVdist(\Prior{0}, \HelperPrior{0}') + 
\TVdist(\HelperPrior{1}', \Prior{1}) + 
\TVdist(\HelperPrior{0}', \HelperPrior{1}') \\
& \leq \Prior{0} (\LocalSet_{0}^{c}) + \Prior{1}(\LocalSet_{1}^{c})+
\TVdist(\HelperPrior{0}',\HelperPrior{1}').
\end{align*}
Regarding the last term $\TVdist(\HelperPrior{0}',\HelperPrior{1}')$, we invoke Lemma~\ref{lemma:Poisson-moments} to obtain 
\begin{align*}
\TVdist(\HelperPrior{0}', \HelperPrior{1}') & \leq (\numaction - 1) 
\TVdist ( \Exp_{\Xvar} [\Poi(8\log \numaction \cdot \Xvar)] 
-\Exp_{\Xvar'}[ \Poi (8\log \numaction\cdot \Xvar')]) \\
 & \leq \left(\frac{16e\log \numaction}{48\log \numaction}\right)^{48\log \numaction}
 \leq \frac{1}{10}
\end{align*}
as long as the number $\numaction$ of actions is sufficiently large. 
For the remaining two terms
$\Prior{0}(\LocalSet_{0}^{c})+\Prior{1}(\LocalSet_{1}^{c})$, we concentrate on the term
$\Prior{0}(\LocalSet_{0}^{c})$ and the same argument and bound apply to the term $\Prior{1}(\LocalSet_{1}^{c})$. By definition, one has 
\begin{align}
\Prior{0}(\LocalSet_{0}^{c}) \leq \Prob _{\Prior{0}} \left( \left | 
\sum_{\action \in \ActionSpace} \plowerPos(\action) + 
\plowerNeg(\action) - 1 \right |
\geq \frac{1}{5} \right ) + \Prob _{\Prior{0}} \left( 
\mid \Target( \thetabm)- \Exp_{\Prior{0}} [\Target( \thetabm)] \mid 
\geq \frac{ \exp(-96\sqrt{\numobs \plower \log \numobs }}{8} 
\cdot \Rmax \right).\label{eq:fail-prob}
\end{align}
In regard to the first term, one has 
\begin{align*}
\sum_{\action \in \ActionSpace}\plowerPos(\action) + 
\plowerNeg(\action) - 1 = \frac{ 8 \log \numaction}
{\numobs} \cdot(\sum_{\action = 1}^{\numaction-1} \Xvar_{\action} 
- \Exp\left[\Xvar_{\action} \right]),
\end{align*}
which together with the Chebyshev's inequality gives 
\begin{align*}
\Prob _{\Prior{0}} \left ( \left | \sum_{\action \in \ActionSpace} 
\plowerPos (\action) + \plowerNeg(\action) - 1 \right |
\geq \frac{1}{5} \right) & = \Prob \left(\left |
\frac{8\log \numaction}{\numobs} \cdot (\sum_{\action = 1}^{\numaction-1} 
\Xvar_{\action} - \Exp \left [ \Xvar_{\action} \right ] ) \right |
\geq \frac{1}{5} \right) \\
 & \leq\frac{\numaction40^{2}\log^{2} \numaction \Var(\Xvar)} {\numobs^{2}}\\
 & \leq\frac{\numaction40^{2}\log^{2} \numaction}{ \numobs^{2}}\leq\frac{1}{10}.
\end{align*}
Here the penultimate inequality follows from the fact that $\Xvar 
\in[0,1]$ and hence $\Var(\Xvar) \leq 1$, and the last relation holds as long as $\numaction \gg\sqrt{\numaction}\log \numaction$. 

Moving on to the second term in the equation~\eqref{eq:fail-prob}, we have via the Chebyshev's inequality that 
\begin{align*}
\Prob _{\Prior{0}}\left(\mid \Target( \thetabm)-\Exp_{\Prior{0}}[\Target( \thetabm)]
\mid\geq\frac{\exp(-96\sqrt{\numobs \plower \log \numaction}}{8}\cdot \Rmax\right) 
& \leq\frac{8^{2}\Var\left[\sum_{1\leq \action \leq \numaction - 1}
\pitarget(\action)\cdot\frac{\Xvar_{\action}}{\Xvar_{\action} + 
\frac{\numobs \plower }{8\log \numaction}} 
\right]}{{\exp(-96\sqrt{\numobs \plower \log \numaction})}}\\
 & \leq\frac{8^{2}\sum_{1\leq \action \leq \numaction - 1}
 \pitarget^{2}(\action)}{{\exp(-96\sqrt{\numobs \plower \log \numaction})}}.
\end{align*}
Recall that we are working under the assumption~\eqref{eq:case-2}
in which the restriction~\eqref{eq:case-2-implication} holds.
This observation leads to
\begin{align*}
\Prob _{\Prior{0}}\left(| \Target( \thetabm
)-\Exp_{\Prior{0}}[\Target (\thetabm)] | \geq\frac{\exp(-96\sqrt{\numobs \plower
    \log \numaction }}{8}\cdot \Rmax\right) &
\leq8^{2}\pitarget(1)\exp(96\sqrt{\numobs \plower \log \numaction}) \\
& \leq 8^{2} \exp(\numobs \plower -4\sqrt{\numobs \plower \log \numaction})
\\
& \leq 8^{2} \exp(-3\sqrt{\numobs \plower \log
  \numaction})\leq\frac{1}{10}.
\end{align*}
Here the last line holds under the assumption that $\plower \leq
\tfrac{\log \numaction}{\numobs}$ and $\numobs \plower \log \numaction
\gg 1$.

In all, we have arrived at the conclusion that
\begin{align*}
\TVdist \left(\Prior{0},\Prior{1}\right)\leq\frac{3}{10},
\end{align*}
which allows us to combine Lemma~\ref{lemma:fuzzy-tsybakov-1} and
Lemma~\ref{lemma:risk-equivalence-Poisson-1} to finish the proof.


\subsection{Proof of Theorem~\ref{thm:upper-bound-eta}}
\label{sec:Proof-of-Theorem-upper-bound-eta}

Recall that we work under the Poisson sampling model, and hence
throughout this section we use the shorthand notation $\Exp [\cdot]$
to denote expectation under the Poisson model.

Denoting by $\Bias(\VhatCheby) \coloneqq \Exp[\VhatCheby] -
\Vphi(\pitarget)$ the bias of the Chebyshev estimator, we have the
usual \mbox{bias-variance} decomposition---namely
\begin{align*}
  \Exp \left [ \big( \VhatCheby - \Vphi(\pitarget) \big )^2 \right] =
  \big( \Bias(\VhatCheby) \big )^2 + \Var(\VhatCheby).
\end{align*}
As before, we break the analysis into two parts, namely controlling
the bias and variance, and aim at proving the following two bounds:
\begin{subequations}
\label{subeq:cheby-MSE-bounds}
\begin{align}
\label{eq:cheby-bias}   
\big( \Bias(\VhatCheby) \big )^2 & \leq 16 \Rmax^2 \exp(- 2 \sqrt{
  \tfrac{\plaincon_{0}^{2}}{\plaincon_{1}} \numobs \plower \log
  \numaction}), \quad \mbox{and} \\
\label{eq:cheby-variance}
\Var(\VhatCheby) & \leq \cprime \{ 
\Risk_{\numobs}(\pitarget; \pibehave) + \plaincon_{0} \Rmax^2 \log
\numaction \cdot \numaction^{4\plaincon_{0} - \decaycon} \},
\end{align}
\end{subequations}
with $\cprime > 0$ a universal constant.  Taking
the above two bounds together, we can deduce that
\begin{align*}
\Exp \left [ \big( \VhatCheby - \Vphi(\pitarget) \big )^2 \right] &
\leq 16 \Rmax^2 \exp(- 2
\sqrt{\tfrac{\plaincon_{0}^{2}}{\plaincon_{1}} \numobs \plower \log
  \numaction}) + \cprime \Risk_{\numobs}(\pitarget; \pibehave) + \cprime \plaincon_{0}
\Rmax^2 \log \numaction \cdot \numaction^{4\plaincon_{0} - \decaycon} \\ &
\leq \ccon \left \{ \Rmax^2 \exp(-
2\sqrt{\tfrac{\plaincon_{0}^{2}}{\plaincon_{1}} \numobs \plower \log
  \numaction}) + \Risk_{\numobs}(\pitarget; \pibehave) \right \},
\end{align*}
as long as $\plaincon_{0} \leq \decaycon / 7$, and $\ccon > 0$ is an absolute constant.  Taking the supremum over
$\pibehave$ completes the proof of Theorem~\ref{thm:upper-bound-eta}.

The remaining two sections are devoted to establishing the
bounds~\eqref{subeq:cheby-MSE-bounds}.


\subsubsection{Proof of the bias bound~\eqref{eq:cheby-bias}}

It is easily seen from the definition~\eqref{def:chebyshev-estimator}
of the Chebyshev estimator that
\begin{align*}
\Exp[ \VhatCheby ] & = \sum_{\action \in \ActionSpace}
\pitarget(\action) \Exp \left[ \rhat(\action) g_{\cdeg} (\numobs(\action)) 
\right ] \\ 
& = \sum_{\action \in \ActionSpace}
\pitarget(\action) \sum_{j=0}^{\infty} \Exp \left [ 
\rhat(\action) g_{\cdeg} (\numobs(\action)) \mid \numobs(\action) = j
  \right] \Prob \left(\numobs(\action) = j \right)\\ 
  & = \sum_{\action \in \ActionSpace} \pitarget(\action) \reward(\action)
\sum_{j=0}^{\infty} g_{\cdeg}(j) \Prob \left(\numobs(\action) = j \right),
\end{align*}
where the last relation hinges on the fact that $g_{\cdeg}(0) = 0$.
Consequently, the bias of $\VhatCheby$ is given by
\begin{align}
\Bias(\VhatCheby) & = \sum_{\action \in \ActionSpace}
\pitarget(\action) \reward(\action) \left \{
\sum_{j=0}^{\infty} g_{\cdeg}(j) \Prob \left(\numobs(\action) = j \right) 
- 1 \right \} \nonumber \\
& = \sum_{\action \in \ActionSpace} \pitarget(\action) \reward(\action)
\left\{ \sum_{j=0}^{\infty} \left ( g_{\cdeg}(j) - 1 \right ) 
\Prob \left( \numobs(\action) = j \right) \right\} \nonumber\\
\label{eq:bias}
& = \sum_{\action \in \ActionSpace} \pitarget(\action) \reward(\action) 
\left\{ e^{- \numobs
  \pibehave(\action) } \sum_{j=0}^{\cdeg} \coeff_{j} ( \pibehave(\action) )^{j} 
  \right \}
= \sum_{\action \in \ActionSpace} \pitarget(\action) \reward(\action) 
e^{ - \numobs \pibehave(\action)}
\ScaledCheby_{\cdeg}(\pibehave(\action)).
\end{align}
Here the middle line uses the fact that $\sum_{j=0}^{\infty} \Prob
\left(\numobs(\action)=j\right)=1$, and the last one follows from the definitions
of $g_{\cdeg}(j)$ and $\ScaledCheby_{\cdeg} ( \pibehave(\action) )$.

In light of equation~\eqref{eq:bias}, the key in bounding the bias is
to control $e^{- \numobs \pibehave(\action)} 
\ScaledCheby_{\cdeg} ( \pibehave(\action) )$,
which is supplied in the following lemma. 
\begin{lemma}
\label{lemma:bound-chebyshev}
For any $\pibehave(\action) \geq \plower$, one has 
\begin{align*}
\left | e^{-\numobs \pibehave(\action)}
\ScaledCheby_{\cdeg} ( \pibehave(\action) )\right | \leq 4 
\exp \left ( - \cdeg \sqrt{ \myleft / \myright} \right ). 
\end{align*}
\end{lemma}
\noindent See Appendix~\ref{subseq:proof-of-lemma-cheby-bound} for the
proof of this claim. \\

In all, this leads us to conclude that
\begin{align*}
|\Bias(\VhatCheby)| \leq 4 \Rmax \sum_{\action \in \ActionSpace}
\pitarget(\action) \exp \left ( - \cdeg \sqrt{ \myleft / \myright}
\right ) \leq 4 \Rmax \exp \left (-
\sqrt{\tfrac{\plaincon_{0}^{2}}{\plaincon_{1}} \numobs \plower \log
  \numaction} \right ),
\end{align*}
where we have used the definitions $\myleft = \plower$, 
$\myright = \ccon_{1} \log \numaction / \numobs$ as well as the 
relation $|\reward(\action)| \leq \Rmax$. This establishes 
the bias upper bound~\eqref{eq:cheby-bias}.


\subsubsection{Proof of the variance bound~\eqref{eq:cheby-variance}}

Now we move on to the variance of the Chebyshev estimator
$\VhatCheby$.  Thanks to the independence brought by the Poisson
model, we have
\begin{align}
\Var (\VhatCheby) = \Var \Big ( \sum_{ \action \in \ActionSpace} 
\pitarget(\action) \rhat(\action) g_{\cdeg}(\numobs(\action))) \Big ) = 
\sum_{ \action \in \ActionSpace} \pitarget^{2}(\action) \Var 
\left ( \rhat(\action) g_{\cdeg}(\numobs(\action)) \right ). 
\label{eq:variance-independence}
\end{align}
Applying the law of total variance yields the decomposition
\begin{align}
\Var \left ( \rhat(\action) g_{\cdeg} (\numobs(\action)) \right) &
= \underbrace{ \Var \left ( \Exp \left [ \rhat(\action) 
g_{\cdeg} (\numobs(\action)) \mid \numobs(\action) \right] \right)}_{
\eqqcolon \TermAlpha_{1}} + \underbrace{ \Exp \left [ \Var 
\left ( \rhat(\action) g_{\cdeg} (\numobs(\action)) \mid \numobs(\action) 
\right ) \right ] }_{\eqqcolon \TermAlpha_{2}}.
\label{eq:variance-chebyshev-decomposition}
\end{align}
Suppose for the moment that the two terms $\TermAlpha_{1}$ and 
$\TermAlpha_{2}$ obey (whose proof are deferred to Appendix~\ref{sec:proof-of-bounds-alpha})
\begin{subequations}
\label{subeq:variance-alpha}
\begin{align}
\TermAlpha_{1} & \leq \Rmax^2 \left \{ 
2 e^{-\numobs \pibehave(\action) } + 
\frac{1}{2} \plaincon_{0} \log \numaction \cdot \numaction^{4\plaincon_{0}} 
\right \}, \quad \text{and} \label{eq:variance-alpha-1} \\
\TermAlpha_{2} & \leq \Rmax^2 \left \{ 2 e^{- \numobs \pibehave(\action)} + 
\frac{1}{2} \plaincon_{0} \log \numaction \cdot \numaction^{4\plaincon_{0}} + 
2 \min \left \{ 1, \frac{5}{\numobs \pibehave(\action)}\right \} \right \} . \label{eq:variance-alpha-2}  
\end{align}
\end{subequations}
Then combing the preceding bounds together yields
\begin{align*}
\Var(\VhatCheby) & \leq \Rmax^2 \sum_{\action \in \ActionSpace} 
\pitarget^{2}(\action) \left\{ 2 e^{ - \numobs \pibehave(\action) } 
+ \frac{1}{2} \plaincon_{0} \log \numaction \cdot \numaction^{4\plaincon_{0}} + 
2 \min \left \{ 1, \frac{5}{\numobs \pibehave(\action)}\right \} \right\} \\
& \leq \Rmax^2 \sum_{\action \in \ActionSpace} 
\pitarget^{2}(\action) \left\{ 
 \frac{1}{2} \plaincon_{0} \log \numaction \cdot \numaction^{4\plaincon_{0}} + 
4 \min \left \{ 1, \frac{5}{\numobs \pibehave(\action)}\right \} \right\},
\end{align*}
where the last inequality follows from the elementary bound $e^{-
  \numobs \pibehave(\action)} \leq \min \left \{ 1, \frac{1}{\numobs
  \pibehave(\action)} \right \}$.  Repeating the analysis of the
plug-in estimator, we find that
\begin{align*}
\Rmax^2 \sum_{\action \in \ActionSpace} \pitarget^{2}(\action) 
\min \left \{ 1, \frac{5}{\numobs \pibehave(\action)}\right \}
\leq \ccon \Risk_{\numobs}^{\star}(\pitarget; \pibehave),
\end{align*}
for some constant $\ccon > 0$.
This bound combined with the assumption $\sum_{\action}
\pitarget^2(\action) \leq \numaction^{- \decaycon}$ implies another 
positive constant $\cprime > 0$ such that 
\begin{align*}
\Var(\VhatCheby) \leq \cprime \{ \Risk_{\numobs}^{\star}(\pitarget;
\pibehave) + \plaincon_{0} \Rmax^2 \log \numaction \cdot
\numaction^{4\plaincon_{0} - \decaycon} \},
\end{align*}
which finishes the proof of the variance upper bound~\eqref{eq:cheby-variance}.


\section{Discussion}
\label{sec:discussion}

In this paper, we have studied the off-policy evaluation problem for
multi-armed bandits with bounded rewards in three different settings.
First, when the behavior policy is known, we showed that the Switch
estimator, which interpolates between the plug-in and importance
sampling estimator, is minimax optimal.  Second, when the behavior
policy is unknown, we analyzed performance in terms of a competitive
ratio, and showed that the plug-in estimator is near-optimal.  Third,
we took some initial steps into the intermediate regime, when partial
knowledge of the behavior policy is given in the form of the minimum
probability over all actions.  We showed that the plug-in approach,
while optimal in some regimes, can be sub-optimal, and we developed an
estimator based on Chebyshev polynomials that is provably optimal for
a large family of target distributions.

This paper focused purely on multi-armed bandits, and extending
non-asymptotic analysis of this type to contextual bandits and Markov
decision processes is certainly of interest. In addition to such
extensions, our study leaves a few interesting technical questions to
answer. Let us single out three of them to conclude.

\paragraph{Extension to other reward distributions.}

Our focus throughout the paper has been on the
family~\eqref{eq:reward-family} of reward distributions with bounded
support.  In practice, one might encounter distributions with possibly
unbounded support but controlled moments (e.g., sub-Gaussian and
sub-exponential distributions), or bounds on variance or other
moments.  In these more general settings, it is not \emph{a priori}
clear that a linear\footnote{To be clear, the Switch estimator is
  linear with respect to the observed rewards.}  procedure, such as
the Switch estimator, need be optimal. However, we believe that the
underlying idea of truncating the likelihood ratio should be useful in
general.  From a technical perspective, the set of bounded reward
distributions is convex, which allows us demonstrate that Bernoulli
rewards are the hardest instances within this family; see the proof of
the lower bound in Theorem~\ref{thm:main}.  If we move beyond bounded
rewards, the set of reward distributions can be non-convex in general,
which introduces new challenges.  


\paragraph{Known and unknown $\pibehave$ cases.}

Our current characterization of the gap between these two cases relies
on the support size of the target policy, which allows us to
demonstrate the near-optimality of the plug-in estimator in the
unknown $\pibehave$ case.  However, the support size is a
discontinuous function of the target distribution, which makes it
sensitive to small perturbations. Is it possible to characterize the
gap using a smooth function of the target distribution? 


\paragraph{Adaptivity to the minimum exploration probability.}

The Chebyshev estimator proposed in this paper requires the knowledge
of the minimum exploration probability $\min_{\action
  \in \ActionSpace} \pibehave(\action)$.  In practice, this minimum
probability may not be known.  A natural question, then, is whether it
is possible devise an estimator that adapts to this minimum
probability---that is, exhibits the same optimal behavior without
knowing the minimum probability in advance.  If not, what is the price
for adaptivity? 


\subsection*{Acknowledgements}

Jiantao Jiao and Banghua Zhu were partially supported by NSF Grants
IIS-1901252, and CCF-1909499.  Cong Ma and Martin Wainwright were
partially supported by NSF grant DMS-2015454 and Office of Naval
Research grant DOD-ONR-N00014-18-1-2640.

\bibliographystyle{alpha}
\bibliography{OPE}


\appendix


\section{Proof of Lemma~\ref{lemma:risk-upper-bound-using-Poisson}}
\label{sec:proof-risk-upper-bound-Poisson}

We note that the proof of this result follows that of Lemma 1 in the
paper~\cite{wu2019chebyshev}; we include the details here for
completeness.  The minimax risk $\Risk_{\mathsf{P}} (\pitarget, (1-
\betapar) \numobs, \plower) $ can be rewritten as
\begin{align*}
\Risk_{\mathsf{P}} (\pitarget, (1- \betapar) \numobs, \plower ) = 
\inf_{ \{ \Vhat_{\numobsnew} \} } \sup_{\pibehave \in \PiFamily( \plower)
  ,\RewardDistPl \in \RewardFamily} \Exp_{\pibehave \otimes
  \RewardDistPl} [ (\Vhat_{\numobs'} - \Vphi(\pitarget))^{2} ],
\end{align*}
where $\{ \Vhat_{\numobsnew} \}_{\numobsnew \geq 0}$ denotes a family
of estimators corresponding to the sample size $m$, and
\mbox{$\numobs' \sim \Poi ( (1-\betapar) \numobs )$.}  Using the Bayes
risk as a lower bound of the minimax risk, we have
\begin{align*}
\Risk_{\mathsf{P}} (\pitarget, (1- \betapar) \numobs, \plower ) 
\geq \sup_{\BayesPrior} \inf_{\{ \Vhat_{\numobsnew} \}}
\Exp_{\pibehave \otimes \RewardDistPl} [(\Vhat_{\numobs'} - 
\Vphi(\pitarget)) ^{2}],
\end{align*}
where $\BayesPrior$ is a prior on the parameter space
$\PiFamily(\plower) \times \RewardFamily$.  Note that for any sequence
of estimators $\{\Vhat_{\numobsnew}\}$,
\begin{align*}
\Exp_{\pibehave \otimes \RewardDistPl}[(\Vhat_{\numobs'} - 
\Vphi(\pitarget))^{2}] &
= \sum_{\numobsnew \geq 0} \Exp_{\pibehave \otimes
  \RewardDistPl} [(\Vhat_{\numobsnew} - \Vphi(\pitarget))^{2} \mid
  \numobs' = \numobsnew] \Prob [\numobs' = \numobsnew] \\
& \geq \sum_{\numobsnew = 0}^{\numobs} \Exp_{\pibehave \otimes
  \RewardDistPl}[(\Vhat_{\numobsnew} - \Vphi(\pitarget))^{2}]
  \Prob [\numobs' = \numobsnew].
\end{align*}
Taking the infimum on both sides yields
\begin{align*}
\inf_{\{ \Vhat_{\numobsnew} \}} \Exp_{\pibehave \otimes \RewardDistPl}
    [(\Vhat_{\numobs'} - \Vphi(\pitarget))^{2}] \geq 
    \sum_{\numobsnew = 0}^{\numobs}
    \inf_{\Vhat_{\numobsnew}} \Exp_{\pibehave \otimes \RewardDistPl}
        [(\Vhat_{\numobsnew} - \Vphi(\pitarget))^{2}]
        \Prob [\numobs' = \numobsnew]
\end{align*}
Observe that for any fixed prior $\BayesPrior$, the mapping 
$\numobsnew \mapsto \inf_{\Vhat_{\numobs}}
\Exp_{\pibehave \otimes \RewardDistPl} [( \Vhat_{\numobsnew} - 
\Vphi(\pitarget))^{2}]$ is
decreasing in $\numobsnew$, and hence
\begin{align*}
\inf_{\{ \Vhat_{\numobsnew} \}} \Exp_{\pibehave \otimes
  \RewardDistPl} [(\Vhat_{\numobs'} - \Vphi(\pitarget))^{2}] & \geq
\sum_{\numobsnew = 0}^{\numobs} \inf_{\Vhat_{\numobs}} 
\Exp_{\pibehave \otimes
  \RewardDistPl}[(\Vhat_{\numobs} - \Vphi(\pitarget))^{2}]
  \Prob [\numobs' = \numobsnew]\\ & =
\inf_{\Vhat_{\numobs}} \Exp_{\pibehave \otimes
  \RewardDistPl}[(\Vhat_{\numobs} - \Vphi(\pitarget))^{2}]
   \Prob (\numobs' \leq \numobs) \\
& \geq \inf_{\Vhat_{\numobs}}\Exp_{\pibehave \otimes \RewardDistPl}
[(\Vhat_{\numobs} - \Vphi(\pitarget))^{2}] 
(1 - \exp(- \numobs \betapar^{2}/2)).
\end{align*}
Taking the supremum over all possible priors on both sides and
invoking the minimax theorem (cf.~Theorem 46.5 in the
book~\cite{strasser2011mathematical}) conclude the proof.


\section{Auxiliary results underlying Proposition~\ref{prop:optimal-subset}}

In this section, we prove various auxiliary results that underlie the
proof of Proposition~\ref{prop:optimal-subset}, including
Lemma~\ref{lemma:minimum-value}, used in the proof itself, as well as
Lemma~\ref{lemma:property-of-opt}, which is used to prove
Lemma~\ref{lemma:minimum-value}.

\subsection{Proof of Lemma~\ref{lemma:minimum-value}}
\label{sec:proof-of-lemma-min-value}

To begin with, we make a few simple observations regarding the
optimization problem~\eqref{eq:key-opt-replicate} and the desired
equivalence~\eqref{eq:intermediate-goal}.

\begin{itemize}
\item First, for any action $\pitarget(\action)=0$, one must have $v^{\star}(\action)=0$.
At the same time, $\pitarget(\action)=0$ implies $\likerat(\action)=0$. Therefore on both
sides of the equation~\eqref{eq:intermediate-goal}, the contributions from actions
$a$ with $\pitarget(\action) = 0$ are zero. Consequently, without loss of generality,
we assume that $\pitarget(\action) > 0$ for all $\action \in \ActionSpace$. 
\item Second, if $\pibehave(\action) = 0$ for some $\action
  \in \ActionSpace$, then one must have $v^{\star}(\action) =
  \pitarget(\action) > 0$, which further implies $\action \in
  \SetStar$. As a result, the action $\action$ contributes
  $\pitarget(\action)$ to both sides of the
  equation~\eqref{eq:intermediate-goal}. Consequently, we assume
  without loss of generality that $\pibehave(\action) > 0$ for all
  $\action \in \ActionSpace$.
\item Last but not least, it is straightforward to check that $\zeros
  \leq \vb^{\star} \leq \bpitarget$.
\end{itemize}

\noindent
In what follows, we separate the proof into three cases: (1)
$\vb^{\star} = \bpitarget$, (2) $\vb^{\star} = \zeros$, and (3)
$\zeros \neq \vb^{\star} \neq \bpitarget$.  The desired
equivalence~\eqref{eq:intermediate-goal} is easy to obtain for the
first two cases, while it requires more effort for the last one.

Let us start with the easy cases.

\paragraph{Case 1.}
If $\vb^{\star} = \bpitarget$, then $\SetStar=\ActionSpace$, and
\begin{align*}
\min_{\vb\in\real^{\numaction}}\quad\sqrt{\frac{1}{8 \numobs}\sum_{\action \in \ActionSpace}
\frac{[\pitarget(\action)-v(\action)]^{2}}{\pibehave(\action)}} + 
\frac{1}{2} \sum_{\action \in \ActionSpace} |v(\action)|
= \frac{1}{2} \sum_{\action \in \ActionSpace} |\pitarget(\action)| 
= \frac{1}{2}.
\end{align*}
At the same time, the right hand side of~\eqref{eq:intermediate-goal} reads 
\begin{align*}
\pitarget(\SetStar) + \sqrt{ \frac{ \sum_{\action \notin \SetStar} 
\pibehave(\action) \likerat^{2}(\action)}{\numobs}}
= \pitarget(\SetStar) = 1.
\end{align*}
This establishes the claim for the case when $\vb^{\star} = \bpitarget$.

\paragraph{Case 2.}

If $\vb^{\star}=\zeros$, then
  $\SetStar = \emptyset$, and hence
\begin{align*}
\min_{\vb\in\real^{\numaction}}\quad\sqrt{\frac{1}{8 \numobs}\sum_{\action \in \ActionSpace}
\frac{[\pitarget(\action)-v(\action)]^{2}}{\pibehave(\action)}} + 
\frac{1}{2} \sum_{\action \in \ActionSpace} |v(\action)| 
= \sqrt{\frac{1}{8 \numobs} \sum_{\action \in \ActionSpace}
  \frac{[\pitarget(\action)]^{2}}{\pibehave(\action)}}.
\end{align*}
On the other hand, $\SetStar = \emptyset$ implies
\begin{align*}
\pitarget(\SetStar) + \sqrt{ \frac{ \sum_{\action \notin \SetStar} 
\pibehave(\action) \likerat^{2}(\action)}{\numobs}}
    = \sqrt{\frac{1}{\numobs} \sum_{\action \in \ActionSpace}
  \frac{[\pitarget(\action)]^{2}}{\pibehave(\action)}},
\end{align*}
which matches desired the equivalence~\eqref{eq:intermediate-goal}.

\paragraph{Case 3.}
In the end, we focus on the more challenging case when 
$\zeros \neq \vb^{\star} \neq \bpitarget$.
In view of the optimality condition of the optimization 
problem~\eqref{eq:key-opt-replicate}, we know that
\begin{subequations}
\label{subeq:optimality} 
\begin{align}
\left[\likerat(\action)-\frac{v^{\star}(\action)}{\pibehave(\action)}\right]^{2}
& =2 \numobs \sum_{\action
  \in \ActionSpace}\frac{[\pitarget(\action)-v^{\star}(\action)]^{2}}{\pibehave(\action)},\qquad\text{for
}\action \in \SetStar;\label{eq:KKT-1}\\ (\likerat(\action))^{2} &
\leq2 \numobs\sum_{\action
  \in \ActionSpace}\frac{[\pitarget(\action)-v^{\star}(\action)]^{2}}{\pibehave(\action)},\qquad\text{for
}\action \notin \SetStar.\label{eq:KKT-2}
\end{align}
\end{subequations}To simplify the notation hereafter, we denote 
\begin{align*}
\Term_{1}\coloneqq\sum_{\action \in \SetStar}\frac{[\pitarget(\action)-v^{\star}(\action)]^{2}}{\pibehave(\action)},\qquad\text{and}\qquad \Term_{2}\coloneqq\sum_{\action \notin \SetStar}\frac{[\pitarget(\action)-v^{\star}(\action)]^{2}}{\pibehave(\action)}=\sum_{\action \notin \SetStar}\pibehave(\action)\likerat^{2}(\action).
\end{align*}
A few immediate consequences of the optimality condition is summarized
in the following claim, whose proof is deferred to the end of this
section. 

\begin{lemma}
\label{lemma:property-of-opt}
Suppose that $\zeros \neq \vb^{\star} \neq \bpitarget$ is the
minimizer of the optimization problem~\eqref{eq:key-opt-replicate}.
Then the following conclusions hold:
\begin{enumerate}[label=(\alph*)]
\item There exists some quantity $\varepsilon \in (0,1)$ such that
  $\pibehave(\SetStar) = (1-\varepsilon) /(2 \numobs)$.
\item We have the relation
  $\Term_{1}=\frac{1-\varepsilon}{\varepsilon}\Term_{2}$.
\item The convex program~\eqref{eq:key-opt-replicate} has optimal
  value $\tfrac{1}{2} \pitarget(\SetStar) + \sqrt{\tfrac{\Term_{2}}{8
      \numobs}} \, \varepsilon$.
\end{enumerate}
\end{lemma}
\noindent See Appendix~\ref{SecLemProof:property-of-opt} for the proof
of this claim. \\

We now use Lemma~\ref{lemma:property-of-opt} to establish the
equivalence~\eqref{eq:intermediate-goal} for the third case $\zeros \neq
\vb^{\star} \neq \bpitarget$.  Part (c) of
Lemma~\ref{lemma:property-of-opt} guarantees that then the desired
equivalence~\eqref{eq:intermediate-goal} holds for any $\varepsilon \in
[1/2,1)$.  Therefore, the remainder of our analysis is devoted to the
  case when $\varepsilon \in (0,1/2)$, meaning that $\frac{1}{4 \numobs}
  < \pibehave(\SetStar) < \frac{1}{2 \numobs}$.

Without loss of generality, we assume that the actions are ordered
according to their likelihood ratios---that is, $\likerat(1) \leq
\likerat(2) \leq \cdots \leq \likerat(\numaction)$.  In view of the
optimality condition~\eqref{subeq:optimality} and the restriction
$0<\frac{1}{4 \numobs}\leq\pibehave(\SetStar)\leq\frac{1}{2
  \numobs}<1$, the subset $(\SetStar)^c$ must be of the form $\{1,2,
\ldots, t\}$ for some $t \in [\numaction-1]$, and hence
$\SetStar=\{t+1, \ldots, \numaction\}$.  In words, the support set
$\SetStar$ contains the actions with larger likelihood ratios
$\likerat(\action)$; see the optimality
condition~\eqref{subeq:optimality}.  By applying the first optimality
condition~\eqref{eq:KKT-1}, we find that
\begin{align*}
\likerat^{2}(t+1) \geq \left[\likerat(t+1) -
  \frac{u^{\star}(t+1)}{\pibehave(t+1)} \right]^{2} = 2 \numobs
(\Term_{1} + \Term_{2}) = \frac{ 2 \numobs \Term_{2}}{\varepsilon},
\end{align*}
where the first relation follows from observation that $0\leq
u^{\star}(t+1)\leq\pitarget(t+1)$ and the last equality arises from
Lemma~\ref{lemma:property-of-opt}(b).  In addition, note that
\begin{align*}
\frac{\pitarget(\SetStar)}{\pibehave(\SetStar)} = 
\frac{\sum_{a=t+1}^{\numaction}\pitarget(\action)}
{\sum_{a=t+1}^{\numaction}\pibehave(\action)}\geq
\frac{\pitarget(t+1)}{\pibehave(t+1)}=\likerat(t+1).
\end{align*}
Combining the previous two bounds yields
\begin{align*}
\frac{\pitarget(\SetStar)}{\pibehave(\SetStar)}\geq\likerat(t+1)\geq\sqrt{\frac{2
    \numobs \Term_{2}}{\varepsilon}}.
\end{align*}
This inequality, together with the assumption that
$\pibehave(\SetStar)\geq1/(4 \numobs)$, guarantees that
\begin{align*}
\pitarget(\SetStar)\geq\sqrt{\frac{2
    \numobs}{\varepsilon}\Term_{2}}\pibehave(\SetStar)\geq\sqrt{\frac{\Term_{2}}{8
    \numobs\varepsilon}}\geq\sqrt{\frac{\Term_{2}}{4
    \numobs}}\geq\sqrt{\frac{\Term_{2}}{4 \numobs}\varepsilon}.
\end{align*}
Here the assumption that $\varepsilon\in (0,1/2)$ is repeatedly used.
This together with Lemma~\ref{lemma:property-of-opt}(c) leads to the
conclusion that
\begin{align*}
\sqrt{\frac{1}{8 \numobs}\sum_{\action \in \ActionSpace}
\frac{[\pitarget(\action)-v^{\star}(\action)]^{2}}{\pibehave(\action)}} 
+ \frac{1}{2}\sum_{\action \in \ActionSpace}
|v^{\star}(\action)| = \frac{1}{2} \pitarget(\SetStar) +
\sqrt{\frac{\Term_{2}}{8\numobs}} \varepsilon \asymp \pitarget(\SetStar)
\asymp \pitarget(\SetStar) + \sqrt{\frac{\Term_{2}}{\numobs}}.
\end{align*}
As a result, in the case when $\varepsilon \in (0,1/2)$, the target
equivalence~\eqref{eq:intermediate-goal} follows.


\subsection{Proof of Lemma~\ref{lemma:property-of-opt}}
\label{SecLemProof:property-of-opt}
Summing the first optimality condition~\eqref{eq:KKT-1} over actions
in $\SetStar$ yields
\begin{align}
\label{eq:key-equality}  
\Term_{1} = \sum_{\action \in \SetStar} \pibehave(\action) \left[
  \likerat(\action) - \frac{v^{\star}(\action)}{\pibehave(\action)}
  \right]^{2} & =2 \numobs \pibehave(\SetStar) \sum_{\action
  \in \ActionSpace} \frac{[\pitarget(\action)-
    v^{\star}(\action)]^{2}}{\pibehave(\action)} = 2 \numobs
\pibehave(\SetStar) (\Term_{1} + \Term_{2}),
\end{align}
which implies $(\SetStar)^c\neq\emptyset$. To see this,
assume for the moment that $(\SetStar)^c=\emptyset$ and hence
$\SetStar=\ActionSpace$, $\Term_{2}=0$. The relation above then reduces
to
\begin{align*}
\Term_{1}=2 \numobs \Term_{1},
\end{align*}
which requires $\Term_{1}=0$ and hence $\vb^{\star}=\bpitarget$. This
contradicts the assumption that $\vb^{\star}\neq\bpitarget$. Since
$(\SetStar)^c$ is non-empty and $\likerat(\action)>0$, one must have $\Term_{2}>0$.
In addition, since $\vb^{\star}\neq\zeros$, $\SetStar$
is nonempty ($\pibehave(\SetStar)>0$), which together
with the identity~\eqref{eq:key-equality} reveals that 
\begin{align*}
\Term_{1} > 2 \numobs \pibehave(\SetStar)\Term_{1}.
\end{align*}
This readily gives the first claim that 
\begin{align*}
\pibehave(\SetStar) = (1-\varepsilon) \tfrac{1}{2 \numobs}
\end{align*}
for some $\varepsilon \in (0,1)$. With this representation in place, we
can also deduce from the equation~\eqref{eq:key-equality} that
\begin{align*}
\Term_{1} = \frac{\pibehave(\SetStar)}{\frac{1}{2 \numobs} -
  \pibehave(\SetStar)} \Term_{2} = \frac{1-\varepsilon}{\varepsilon}
\Term_{2},
\end{align*}
which is the second claim. Regarding the last claim, applying the
first optimality condition~\eqref{eq:KKT-1} ensures that for $\action
\in \SetStar$, we have
\begin{align*}
v^{\star}(\action) = \pitarget(\action) - \pibehave(\action) \sqrt{2
  \numobs (\Term_{1} + \Term_{2})}.
\end{align*}
As a result, the minimum value obeys
\begin{align*}
\sqrt{\frac{1}{8 \numobs} \sum_{\action \in \ActionSpace}
  \frac{[\pitarget(\action)-v^{\star}(\action)]^{2}}{\pibehave(\action)}}
+ \frac{1}{2} \sum_{\action \in \ActionSpace} | v^{\star}(\action)| &
= \sqrt{\frac{1}{8 \numobs}}\sqrt{\Term_{1} + \Term_{2}} + \frac{1}{2}
\sum_{\action \in \SetStar} v^{\star}(\action) \\
& = \sqrt{\frac{1}{8 \numobs}} \sqrt{\Term_{1} + \Term_{2}} +
\frac{1}{2} \left(\pitarget(\SetStar) - \pibehave(\SetStar) \sqrt{2
  \numobs (\Term_{1} + \Term_{2})} \right) \\
& = \frac{1}{2}\pitarget(\SetStar) + \sqrt{\frac{1}{8 \numobs}}
\sqrt{\Term_{1} + \Term_{2}} \left(1 - 2 \numobs \pibehave(\SetStar)
\right).
\end{align*}
Use the first two claims $\Term_{1} + \Term_{2} = \Term_{2}/\varepsilon$
and $1 - 2 \numobs\pibehave(\SetStar) = \varepsilon$ to finish the proof.


\section{Proof of Lemma~\ref{lemma:duality-lower-bound}}
\label{sec:proof-duality-lower-bound}

In this appendix, we derive the dual formulation of the primal
problem~\eqref{eq:lower-bound-opt-primal}.  First, note that we may
assume without loss of generality that $\pibehave(\action) >
0$. Indeed, if $\pibehave(\action) = 0$ for some action $\action$,
then the optimal primal variable $\delpar(\action)$ in the primal
problem~\eqref{eq:lower-bound-opt-primal} should be set to $1/2$,
while the optimal dual variable $v(\action)$ should be
$\pitarget(\action)$ in the dual
formulation~\eqref{eq:lower-bound-opt-dual}.  Both contribute
$\tfrac{1}{2}\pitarget(\action)$ to the objective values.  For a
scalar $\lambda \geq 0$ and vector $\vb \geq 0$, the Lagrangian of the
primal problem~\eqref{eq:lower-bound-opt-primal} is given by
\begin{align*}
\LagFun (\bm{\delpar},\lambda,\vb) & \coloneqq - \sum_{\action
  \in \ActionSpace} \pitarget(\action) \delpar(\action) + \lambda
\Big( \sum_{\action \in \ActionSpace} \pibehave(\action)
\delpar^{2}(\action) - \frac{1}{8 \numobs} \Big) + \sum_{\action
  \in \ActionSpace} v(\action) (\delpar(\action) - \frac{1}{2}) \\ & =
- \frac{\lambda}{8 \numobs} + \sum_{\action \in \ActionSpace} \left \{
\lambda \pibehave(\action) \delpar^{2}(\action) + [v(\action) -
  \pitarget(\action)] \delpar(\action) - \frac{1}{2} v(\action) \right
\} .
\end{align*}
We now compute the dual function $g(\lambda, \vb) = \inf_{\bm{\delpar}
  \in \real^\numaction} \LagFun( \bm{\delpar}, \lambda, \vb)$, and
find that
\begin{align*}
  g(\lambda,\vb) & = \begin{cases} -\frac{\lambda}{8 \numobs} -
    \frac{1}{4\lambda} \sum_{\action \in \ActionSpace}
    \frac{[\pitarget(\action) - v(\action)]^{2}} {\pibehave(\action)}
    - \frac{1}{2} \sum_{\action \in \ActionSpace}v(\action) & \mbox{if
      $\lambda>0$,} \\
       -\frac{1}{2} & \mbox{if $\lambda=0$ and $\vb=\bpitarget$, and}
       \\
       -\infty & \mbox{otherwise.}
\end{cases}
\end{align*}
The value in the first case follows by choosing the optimal
$\delpar^{\star}(\action) = \frac{\pitarget(\action) - v(\action)}{2
  \lambda \pibehave(\action)}$.

Since the primal problem~\eqref{eq:lower-bound-opt-primal} satisfies
Slater's condition, strong duality holds and hence
\begin{align*}
-\sum_{\action \in \ActionSpace} \pitarget(\action)
\delpar^{\star}(\action) & = \max_{\lambda\geq0,\vb\geq \zeros}
g(\lambda,\vb) \; = \; \max_{\vb\geq \zeros} \left \{
-\sqrt{\frac{1}{8 \numobs}\sum_{\action \in \ActionSpace}
  \frac{[\pitarget(\action) - v(\action)]^{2}}{\pibehave(\action)}} -
\frac{1}{2} \sum_{\action \in \ActionSpace} v(\action) \right \} \\
& = - \min_{\vb \geq 0} \left \{ \sqrt{\frac{1}{8 \numobs}
  \sum_{\action \in \ActionSpace} \frac{[\pitarget(\action) -
      v(\action)]^{2}}{\pibehave(\action)}} + \frac{1}{2}
\sum_{\action \in \ActionSpace} v(\action) \right \}.
\end{align*}
The constrained optimization problem on the right hand side is
equivalent to the unconstrained one~\eqref{eq:lower-bound-opt-dual},
which completes the proof.


\section{Proof of auxiliary lemmas for Theorem~\ref{thm:ratio-lower-bound}}
\label{subsec:Proof-of-auxiliary-lower-bound-eta}

In this section, we collect the proofs of various auxiliary lemmas
used in the proof of Theorem~\ref{thm:ratio-lower-bound}.

\subsection{Proof of Lemma~\ref{lemma:risk-equivalence-Poisson-1}}

For each positive integer $\ell$, let $\Vhat_\ell$ be the optimal
estimator under the multinomial sampling model based on $\ell$
samples, one that achieves the minimax risk
$\Risk_{\mathsf{M}}^{\star} \left(\pitarget, \ell, \plower /(1 +
\varepsilon) \right)$.  Now we define a near-optimal estimator in the
Poisson sampling model.  Let $T$ be the total number of rewards
observed in the Poissonized model; by construction, the random
variable $T$ follows a $\Poi(\numobs \sum_{\action =
  1}^{\numaction}\pibehave(\action))$ distribution.

Now consider the estimator $\Vhat_{T}$---that is, the minimax optimal
estimator based on $T$ samples.  This choice yields an upper bound on
the Poissonized risk, namely
\begin{align*}
\Risk_{\mathsf{P}}^{\star} \left(\pitarget, \numobs, \plower,
\varepsilon \right) \leq \sup_{\pibehave \in \LocalParaSpace(\plower
  ,\varepsilon), \RewardDistPl \in \RewardFamily} \Exp_{\pibehave
  \otimes \RewardDistPl} \Big[ (\Vhat_{T} - \sum_{\action
    \in \ActionSpace} \pitarget(\action) \reward(\action))^{2} \Big].
\end{align*}
Further note that for any behavior policy $\pibehave \in
\Theta(\plower, \varepsilon)$, the distribution
$\tilde{\pi}_{\mathsf{b}} = \pibehave/\|\pibehave\|_{1}$ satisfies the
lower bound \mbox{$\pibehave \geq \plower /(1+\varepsilon)$.} In
addition, conditional on the realization of $T$, the Poisson model is
equivalent to the original multinomial model with a discrete
distribution $\tilde{\pi}_{\mathsf{b}}$.  Combining these two facts
together guarantees that, for any \mbox{$\pibehave
  \in \LocalParaSpace(\plower, \varepsilon)$,} we have the
decomposition
\begin{align*}
\Exp_{\pibehave \otimes \RewardDistPl} \Big[ (\Vhat_{T}-\sum_{\action
    \in
    \ActionSpace} \pitarget(\action) \reward(\action))^{2} \Big] & =
\sum_{\ell=0}^{\infty} \Exp_{\pibehave \otimes \RewardDistPl} \Big[
  (\Vhat_{T}-\sum_{\action = 1}^{\numaction} \pitarget(\action)
  \reward(\action))^{2} \,\mid \, T = \ell \Big] \cdot \Prob(T = \ell)
\\
& = \sum_{\ell=0}^{\infty} \Exp_{\tilde{\pi}_{\mathsf{b}}
  \otimes\RewardDistPl} \Big[(\Vhat_{\ell}-\sum_{\action
    \in \ActionSpace} \pitarget(\action) \reward(\action))^{2} \mid
  T=\ell \Big] \cdot \Prob(T=\ell) \\
& \leq \sum_{\ell=0}^{\infty} \Risk_{\mathsf{M}}^{\star} \left(
\pitarget, \ell, \tfrac{\plower}{1 + \varepsilon} \right) \cdot
\Prob(T=\ell),
\end{align*}
where the last line uses the fact that $\Vhat_\ell$ is minimax
optimal.  Since the function $\ell \mapsto
\Risk_{\mathsf{M}}^{\star}\left(\pitarget, \ell,
\tfrac{\plower}{1+\varepsilon} \right)$ is non-increasing, we can
write
\begin{align*}
\Exp_{\pibehave \otimes \RewardDistPl} [(\Vhat_{T} - \sum_{\action
    \in \ActionSpace} \pitarget(\action) r(\action))^{2}] & \leq
\Risk_{\mathsf{M}}^{\star} \left(\pitarget, 0,
\tfrac{\plower}{1+\varepsilon} \right) \Prob(T < \tfrac{\numobs}{2}) +
\Risk_{\mathsf{M}}^{\star} \left( \pitarget, \tfrac{\numobs}{2},
\tfrac{\plower}{1+\varepsilon} \right) \\
& \stackrel{(i)}{\leq} \Rmax^{2} \cdot \Prob(T < \tfrac{\numobs}{2}) +
\Risk_{\mathsf{M}}^{\star} \left (\pitarget, \tfrac{\numobs}{2},
\tfrac{\plower}{1 + \varepsilon} \right) \\
& \stackrel{(ii)}{\leq} \Rmax^{2} \cdot e^{-3 \numobs/72} +
\Risk_{\mathsf{M}}^{\star} \left(\pitarget, \tfrac{\numobs}{2},
\tfrac{\plower}{1+\varepsilon} \right),
\end{align*}
where step (i) follows from the inequality $\Risk_{\mathsf{M}}^{\star}
\left( \pitarget, 0, \tfrac{\plower}{1+\varepsilon} \right) \leq
\Rmax^{2}$, whereas step (ii) follows from a standard tail bound for
Poisson random variables; see e.g., Lemma 5 of the 
paper~\cite{han2015minimax}. 


\subsection{Proof of Lemma~\ref{lemma:existence-X-X'}}

Lemmas~\ref{lemma:error-1/x} and~\ref{lemma:duality} guarantee the
existence of random variables $\Uvar$ and $\Uvar'$ supported on the
interval $[\tfrac{\numobs \plower }{4\log \numaction}, \; 1]$ such
that
\begin{align*}
\Exp [\tfrac{1}{\Uvar}]- \Exp [\tfrac{1}{\Uvar'}] & \geq \frac{4\log
  \numaction}{\numobs \plower } \cdot (1-2 \sqrt{\frac{\numobs \plower
  }{4 \log \numaction}})^{\lceil 48 \log \numaction \rceil} \\
& \geq\frac{4\log \numaction}{\numobs \plower} \cdot \exp(-96
\sqrt{\numobs \plower \log \numobs}) \\
\Exp[\Uvar^{j}] & = \Exp[(\Uvar')^{j}] \quad \mbox{for all $j= 0, 1,
  \ldots,\lceil 48 \log \numaction\rceil -1$.}
\end{align*}
Here we used the fact that $(1-x)^{\numobs} \geq e^{-2 \numobs x}$ for
$x \in [0, \tfrac{1}{2}]$.

Define the shifted random variables $\Vvar \coloneqq \Uvar -
\frac{\numobs \plower }{8\log \numaction}$ and $\Vvar' \coloneqq
\Uvar'-\frac{\numobs \plower }{8\log \numaction}$.  With these
definitions, it is straightforward to check that $\Vvar$ and $\Vvar'$
take values in the interval $[\frac{\numobs \plower }{8\log
    \numaction},1]$ and they satisfy the bounds
\begin{align}
\label{eq:gap-V}   
\Exp\left[\frac{1}{\Vvar + \frac{\numobs \plower }{8 \log \numaction}}
  \right] - \Exp \left[ \frac{1}{\Vvar' + \frac{\numobs \plower }{8\log
      \numaction}} \right] & \geq\frac{4 \log \numaction}{\numobs
  \plower } \cdot \exp(-96 \sqrt{\numobs \plower \log \numaction}),
\quad \mbox{and} \\
\Exp[\Vvar^{j}] & = \Exp[(\Vvar')^{j}], \mbox{for any $j= 0, 1, \ldots, \lceil
  48 \log \numaction \rceil-1$.} \nonumber
\end{align}

Finally, we construct the desired random variables $\Xvar, \Xvar'$
 via changes
of variables on $\Vvar, \Vvar'$. More specifically, the probability density
functions of $\Xvar, \Xvar'$ are given by
\begin{align*}
P_{\Xvar}(\mathsf{d}\xvar) & \coloneqq\left(1-\Exp\left[\frac{\numobs \plower/
    (8\log
    \numaction)}{\Vvar}\right]\right)\delta_{0}(\mathsf{d}\xvar)+\frac{\numobs
  \plower /(8\log \numaction)}{\xvar} P_{\Vvar}(\mathsf{d}\xvar), \quad \mbox{and} \\
P_{\Xvar'}(\mathsf{d}\xvar) & \coloneqq \left( 1 - \Exp\left[\frac{\numobs
    \plower /(8\log \numaction)}{\Vvar'}\right]\right)
\delta_{0}(\mathsf{d}\xvar)+\frac{\numobs \plower /(8\log
  \numaction)}{\xvar}P_{\Vvar'}(\mathsf{d}\xvar).
\end{align*}
This is a valid construction since $\Vvar, \Vvar'\geq \numobs \plower /(8\log
\numaction)$. In addition, we see that $\Xvar, \Xvar' \in [0,1]$, and for any
$j = 1, \ldots, \lceil 48 \log \numaction \rceil$,
\begin{align*}
\Exp[\Xvar^{j}] & = \int_{\frac{\numobs \plower }{8\log
    \numaction}}^{1}\xvar^{j-1} \cdot\frac{\numobs \plower }{8\log
  \numaction}\cdot P_{\Vvar}(\mathsf{d}\xvar) = \frac{\numobs \plower
  \Exp[\Vvar^{j-1}]}{8\log \numaction} = \frac{\numobs \plower
  \Exp[(\Vvar')^{j-1}]}{8\log \numaction} = \Exp[(\Xvar')^{j}],
\end{align*}
where we have used the fact that $\Exp[\Vvar^{j}]=\Exp[(\Vvar')^{j}]$ for all
\mbox{$j = 0, 1, \ldots, \lceil 48 \log \numaction\rceil - 1$.} The
above formula also tell us that $\Exp[\Xvar] = \Exp[\Xvar'] = \frac{\numobs
  \plower }{8\log \numaction}\cdot\Prob (\Vvar > 0)=\frac{\numobs \plower
}{8\log \numaction}$.

Furthermore, we have
\begin{align*}
\Exp \left [\frac{\Xvar}{ \Xvar + \frac{\numobs \plower }{8\log \numaction}}
  \right ] =\int_{\frac{\numobs \plower }{8\log \numaction}}^{1}
\frac{1}{\xvar + \frac{\numobs \plower }{8\log \numaction}} \frac{\numobs
  \plower }{8 \log \numaction} \cdot P_{\Vvar}(\mathsf{d}\xvar)=\frac{\numobs
  \plower }{8\log \numaction} \Exp \left[ \frac{1}{\Vvar + \frac{\numobs
      \plower }{8 \log \numaction}} \right],
\end{align*}
Together with equation~\eqref{eq:gap-V}, this relation implies that
\begin{align*}
\Exp \left[\frac{\Xvar}{ \Xvar + \frac{\numobs \plower }{8\log \numaction}}
  \right] - \Exp \left[ \frac{\Xvar'}{\Xvar'+\frac{\numobs \plower}{8 \log
      \numaction}} \right] \geq \frac{1}{2} \exp(-96 \sqrt{\numobs
  \plower \log \numaction}),
\end{align*}
which concludes the proof.

\section{Auxiliary results underlying Theorem~\ref{thm:upper-bound-eta}} \label{sec:aux-upper-bound-eta}

In this section, we collect the proofs of some useful results for establishing Theorem~\ref{thm:upper-bound-eta}.

\subsection{Proof of Lemma~\ref{lemma:bound-chebyshev}}
\label{subseq:proof-of-lemma-cheby-bound}

In order to simplify notation, let us introduce the shorthand
\begin{align}
\label{defn:ratio}  
\ratio \coloneqq \frac{\myleft}{\myright} = \frac{\plower
}{ \frac{\ccon_{1}\log \numaction}{\numobs}}.
\end{align}
We split the proof into two cases, depending on whether
$\pibehave(\action) \in [\myleft, \myright]$, or $\pibehave(\action) >
\myright$.

\paragraph{Case 1: $\pibehave(\action) \in [\myleft, \myright]$.}
When $\pibehave(\action) \in [\myleft, \myright]$, we have
$|\ChebPoly( \frac{ 2 \pibehave(\action) - \myright - \myleft}
{\myright - \myleft})| \leq 1$, and hence
\begin{align*}
|\ScaledCheby_{\cdeg} ( \pibehave(\action) ) | & \leq 
\frac{1}{\big| \ChebPoly
  \big(\frac{ - \myright - \myleft}{\myright - \myleft} \big) \big |} =
\big| \frac{1}{\ChebPoly \big(-\frac{1+\ratio}{1-\ratio} \big)} \big|,
\end{align*}
which implies that $\left |e^{-\numobs \pibehave(\action)}
\ScaledCheby_{\cdeg}(\pibehave(\action)) \right | \leq \frac{1}{\left
  | \ChebPoly\left(-\frac{1+\ratio}{1-\ratio}\right)\right |}$.

\paragraph{Case 2: $\pibehave(\action) > \myright$.}
When $\pibehave(\action) > \myright$, we know that
\begin{align*}
\left |e^{-\numobs \pibehave(\action)}
\ScaledCheby_{\cdeg}(\pibehave(\action)) \right | \leq \max_{x \in
  (\myright, 1]} e^{-\numobs x} | \ScaledCheby_{\cdeg}(x)| \leq
\max_{1< y \leq \frac{2-\myright - \myleft}{\myright - \myleft}}
\exp\left(- \numobs \myright y ( 1 - \ratio ) / 2 \right )
\ChebPoly(y) \frac{ \exp( - \numobs \myright (1 + \ratio) / 2 )}
         {\left | \ChebPoly \left (- \frac{1 + \ratio}{ 1 -
             \ratio}\right)\right |}.
\end{align*}
Lemma 4 from Wu et al.~\cite{wu2019chebyshev} guarantees that if
$\LocalBeta = O(\cdeg)$, then
\begin{align*}
\sup_{y \geq 1} \left \{ e^{-\LocalBeta y} \ChebPoly(y) \right \} =
\frac{1}{2} \left ( \frac{\LocalAlpha + \sqrt{\LocalAlpha^{2} +
    1}}{e^{\sqrt{ 1 + 1/\LocalAlpha^{2}}}} \left( 1 + o_{\cdeg}(1)
\right) \right)^{\cdeg}, \qquad \mbox{as $\cdeg \to \infty$,}
\end{align*}
where $\LocalAlpha \coloneqq \cdeg / \LocalBeta$. We apply this
identity with the choices
\begin{align*}
\LocalBeta = \numobs \myright (1 - \ratio ) / 2 \leq
\ccon_{1}\log \numaction, \qquad \LocalAlpha= 2 \LocalLambda \coloneqq
2\frac{\plaincon_{0}}{\plaincon_{1}(1-\ratio)},
\end{align*}
thereby obtaining the inequality
\begin{align*}
\left | e^{ - \numobs \pibehave(\action)} \ScaledCheby_{\cdeg}(
\pibehave(\action) ) \right | & \leq \frac{1}{2} \left ( \frac{2
  \LocalLambda +\sqrt{ 4 \LocalLambda^{2} + 1}} {e^{\sqrt{ 1 + 1/(4
      \LocalLambda^{2} ) } } } \left(1 + o_{\numaction}(1) \right )
\right )^{\cdeg} \frac{ \exp( -\numobs \myright (1 + \ratio ) / 2 ) }
       {\left | \ChebPoly \left( -\frac{1 + \ratio}{1-\ratio} \right)
         \right |} \\
& = \frac{1}{2} \left(\frac{2 \LocalLambda + \sqrt{4
           \LocalLambda^{2}+1}}{e^{\sqrt{1 + 1/(4\LocalLambda^{2})} +
           1/(2\LocalLambda)}} \left( 1 + o_{\numaction}(1) +
       o_{\ratio}(1) \right) \right)^{\cdeg} \frac{1}{\left |
         \ChebPoly \left( -\frac{1+\ratio}{1-\ratio} \right) \right
         |}.
\end{align*}
\newcommand{\plainconzero}{\ensuremath{\plaincon_0}}
\newcommand{\plainconone}{\ensuremath{\plaincon_1}}

By a suitable choice of the universal constants $(\plainconzero,
\plainconone)$---in particular, by taking $\plainconzero \ll
\plainconone$--- we can make $\LocalLambda$ as small as we please.
This freedom allows us to guarantee that for all $\pibehave(\action)
\geq \myright$, we have
\begin{align*}
\left |e^{-\numobs \pibehave(\action)} \ScaledCheby_{\cdeg}(
\pibehave(\action) ) \right | \leq \frac{\left(1 + o_{\numaction}(1) +
  o_{\ratio}(1) \right)}{\left | \ChebPoly\left(\frac{-\myright-
    \myleft}{\myright- \myleft} \right) \right |} \leq \frac{2}{\left
  | \ChebPoly\left(-\frac{1+\ratio}{1-\ratio} \right) \right |},
\end{align*}
as long as $\numaction$ and $\plaincon_{1}$ are both sufficiently large.

\paragraph{Putting pieces together.}
By combining the previous two cases together, we find that for any
$\pibehave(\action) \geq \myleft$, we have
\begin{align*}
\left |e^{-\numobs \pibehave(\action)} \ScaledCheby_{\cdeg}(
\pibehave(\action) ) \right | \leq \frac{2}{\left |
  \ChebPoly\left(-\frac{1+\ratio}{1-\ratio} \right) \right |} \leq 4
\left ( 1 - \frac{2 \sqrt{\ratio}}{1 + \sqrt{\ratio}}\right )^{\cdeg}.
\end{align*}
In this argument, the final inequality exploits a basic fact about
Chebyshev polynomials, namely that
\begin{align*}
\left | \ChebPoly\left(-\frac{1+\ratio}{1-\ratio} \right) \right |
\geq \frac{1}{2} \left ( 1 - \frac{2 \sqrt{\ratio}}{1 + \sqrt{\ratio}}
\right )^{- \cdeg}.
\end{align*}
To conclude, we make note of the elementary inequalities
$(1-x)^{\cdeg} \leq \exp(-x\cdeg)$ for $x \in (0,1)$, and $\frac{2
  \sqrt{\ratio}}{1 + \sqrt{\ratio}} \geq \sqrt{\ratio}$.  Substituting
these bounds yields the claimed result---viz.
\begin{align*}
\left |e^{-\numobs \pibehave(\action)} \ScaledCheby_{\cdeg}(
\pibehave(\action) ) \right | \leq 4 \exp \left( -\frac{2 \cdeg
  \sqrt{\ratio}}{1 + \sqrt{\ratio}} \right ) \leq 4 \exp \left( -
\cdeg \sqrt{\ratio} \right ).
\end{align*}

\subsection{Proof of the bounds~\eqref{subeq:variance-alpha}}
\label{sec:proof-of-bounds-alpha}

We prove each of the two bounds~\eqref{eq:variance-alpha-1}
and~\eqref{eq:variance-alpha-2} in turn. 

\paragraph{Proof of the inequality~\eqref{eq:variance-alpha-1}.}

When it comes to $\TermAlpha_{1}$, given that $g_{\cdeg}(0)=0$, one has
\begin{align*}
\Exp \left [ \rhat(\action) g_{\cdeg}(\numobs(\action)) \mid 
\numobs(\action) \right ] = \reward(\action) g_{\cdeg} (\numobs(\action)),
\end{align*}
and hence
\begin{align*}
\TermAlpha_{1} = \Var \left ( \reward(\action) g_{\cdeg}(\numobs(\action)) 
\right ) \leq \Rmax^2 \Var \left ( g_{\cdeg}(\numobs(\action)) \right ) 
= \Rmax^2 \Var \left( g_{\cdeg}(\numobs(\action)) - 1 \right).
\end{align*}
Here, the inequality arises from the fact that 
$|\reward(\action)| \leq \Rmax$, and the last identity uses the 
translation invariance of the variance. 
Splitting $g_{\cdeg}(\numobs(\action)) - 1$ into $(g_{\cdeg}(\numobs(\action)) - 1) \indicator \{ \numobs(\action) \leq \cdeg \}$
and $(g_{\cdeg}(\numobs(\action)) - 1) \indicator \{ \numobs(\action) > \cdeg \}$ 
and using the elementary inequality $\Var(X+Y)\le2\Var(X)+2\Var(Y)$,
we can obtain
\begin{align*}
\tfrac{\TermAlpha_{1}}{\Rmax^2} & \leq 2 \Var \left ( \left \{ 
g_{\cdeg}(\numobs(\action)) - 1\right \} \indicator \{ \numobs(\action) \leq \cdeg \}
\right ) + 2 \Var \left ( \left \{ g_{\cdeg}(\numobs(\action)) - 1 \right \} 
\indicator \{ \numobs(\action) > \cdeg \} \right ) \\
 & = 2 \Var \left( \left \{ g_{\cdeg}(\numobs(\action)) - 1 \right \}  
 \indicator \{ \numobs(\action)\leq \cdeg \} \right)\\
 & \leq 2 \Exp \left ( \left \{ g_{\cdeg}(\numobs(\action)) - 1 \right \}^{2}
 \indicator \{ \numobs(\action) \leq \cdeg \} \right).
\end{align*}
Here, the equality is due to the fact that $g_{\cdeg}(\numobs(\action)) = 1$ for $\numobs(\action) > \cdeg$. Substitute in the definition of 
$g_{\cdeg}(\numobs(\action))$ to see that
\begin{align}
\frac{\TermAlpha_{1}}{\Rmax^2} & \leq 2 \sum_{j=0}^{\cdeg} 
e ^ { - \numobs \pibehave(\action) } \frac{[ \numobs \pibehave(\action) ]^{j}} {j!} (\coeff_{j} j! / \numobs^{j}) ^ {2} = 
2 e^{ -\numobs \pibehave(\action)} \sum_{j=0}^{\cdeg} \coeff_{j}^{2}j! 
[ \frac{ \pibehave(\action)}{\numobs}]^{j}. \label{eq:bound-alpha-1}
\end{align}
It has been shown in equation~(47) in Section~6.1 in the paper~\cite{wu2019chebyshev} that for $j=1,2,\ldots \cdeg$, 
\begin{align}
\label{eq:chebyshev-coefficient}  
|\coeff_{j}| & \leq \frac{1}{2}
(\frac{4}{\myright})^{j} \exp \left( (\cdeg+j) \entropy (\frac{2j}{\cdeg + j}) \right ) 
\leq \frac{1}{2} ( \frac{4}{\myright})^{j}
\numaction^{2\plaincon_{0}},
\end{align}
where $ \entropy(\lambda) \coloneqq -\lambda \log\lambda - (1-\lambda) \log (1-\lambda) $ denotes the binary entropy function. The last inequality holds true since $j\leq \cdeg = \plaincon_{0}\log \numaction$ and $\entropy(\lambda) \leq 1$. Combine the previous two bounds~\eqref{eq:bound-alpha-1} and~\eqref{eq:chebyshev-coefficient} together to see
\begin{align*}
\frac{\TermAlpha_{1}}{\Rmax^2} & \leq 2 e^{-\numobs \pibehave(\action)}\left(\coeff_{0}^{2} + \sum_{j=1}^{\cdeg} \coeff_{j}^{2} j! [\frac{\pibehave(\action)}{\numobs}]^{j} \right) \\
 & \leq 2 e^{-\numobs \pibehave(\action)} \left( 1 + \frac{1}{4}\sum_{j=1}^{\cdeg} ( \frac{ 16 \pibehave(\action) }{\numobs \myright^{2}})^{j}\numaction^{4\plaincon_{0}} j! \right) \\
 & \leq 2 e^{-\numobs \pibehave(\action)} + \frac{1}{2} e^{-\numobs 
 \pibehave(\action)} \sum_{j=1}^{\cdeg} (\frac{16 \cdeg \pibehave(\action)}{\numobs \myright^{2}})^{j} \numaction^{4\plaincon_{0}},
\end{align*}
where the last line follows from the elementary inequality $j!\leq j^{j}$ and the fact that $j\leq \cdeg$. To further upper bound $\TermAlpha_{1}$, we consider two separate cases. First, when $ 16 \cdeg \pibehave(\action)\leq \numobs \myright^{2}$, one clearly has 
\begin{align*}
e^{- \numobs \pibehave(\action) } \sum_{j=1}^{\cdeg} ( 
\frac{ 16 \cdeg \pibehave(\action) } { \numobs \myright^{2} } )^{j} 
\numaction^{4 \ccon_{0}}
\leq \cdeg \numaction^{4 \ccon_{0}}.
\end{align*}
On the other hand, when $ 16 \cdeg \pibehave(\action)\geq \numobs \myright^{2}$, we have
\begin{align*}
e^{- \numobs \pibehave(\action)} \sum_{j=1}^{\cdeg} (\frac{16 \cdeg
  \pibehave(\action)}{\numobs \myright^{2}})^{j}
\numaction^{4\plaincon_{0}} & \leq \cdeg \numaction^{4\plaincon_{0}}
e^{-\numobs \pibehave(\action)} \left(\frac{16 \cdeg
  \pibehave(\action)}{\numobs \myright^{2}} \right)^{\cdeg} \\
& = \cdeg \numaction^{ 4\plaincon_{0}} \exp \left(- \numobs
\pibehave(\action) + \cdeg \log \left(\frac{16\plaincon_{0} \numobs
  \pibehave(\action)}{\plaincon_{1}^{2} \log \numaction}\right)\right)
\\
& \leq \cdeg \numaction^{4\plaincon_{0}}
\end{align*}
as long as $\plaincon_{1}^{2} \geq 32 \plaincon_{0}^{2}$. In sum,
using the definition $\cdeg = \plaincon_{0} \log \numaction $, we
arrive at the conclusion that
\begin{align*}
\TermAlpha_{1} \leq \Rmax^2 \left \{ 2 e^{-\numobs \pibehave(\action)
} + \frac{1}{2} \plaincon_{0} \log \numaction \cdot
\numaction^{4\plaincon_{0}} \right \}.
\end{align*}

\paragraph{Proof of the inequality~\eqref{eq:variance-alpha-2}.}

Our next step is to upper bound the second term $\TermAlpha_{2}$.  We
begin by observing that
\begin{align*}
\Var \left( \rhat(\action) g_{\cdeg}(\numobs(\action)) \mid 
\numobs(\action) \right ) = g_{\cdeg}^{2} (\numobs(\action)) \frac {
\sigma_{\RewardDistPl}^{2} (\action)} {\numobs(\action)}
\indicator \{ \numobs(\action) > 0\},
\end{align*}
which further implies 
\begin{align*}
\TermAlpha_{2} & \leq \Rmax^2 \Exp \left [ \frac{ g_{\cdeg}^{2}
(\numobs(\action))}{\numobs(\action)} \indicator \{ \numobs(\action) > 0 \} \right ] \\
 & = \Rmax^2 \Exp \left[ \frac{g_{\cdeg}^{2}(\numobs(\action))} {\numobs(\action)} \indicator \{ 0 < \numobs(\action) \leq \cdeg \} \right ]
 + \Rmax^2 \Exp \left[ \frac{1}{\numobs(\action)} \indicator \{\numobs(\action) > \cdeg \} \right ].
\end{align*}
Here the last identity uses the fact that $g_{\cdeg}(\numobs(\action)) = 1$ 
for $\numobs(\action) > \cdeg$.
Using the inequality $(x+y)^{2} \leq 2x^{2} + 2y^{2}$, we can decompose
the first term into
\begin{align*}
\Exp \left [ \frac{ g_{\cdeg}^{2}(\numobs(\action))}{\numobs(\action)} \indicator 
\{ 0 < \numobs(\action) \leq \cdeg \} \right ] & \leq 2 \Exp \left [
\frac{[g_{\cdeg}(\numobs(\action)) - 1]^{2}}{\numobs(\action)} \indicator 
\{0 < \numobs(\action) \leq \cdeg \} \right ] + 2 \Exp \left [ 
\frac{1}{\numobs(\action)} \indicator \{0 < \numobs(\action) \leq \cdeg \} \right],
\end{align*}
which results in 
\begin{align*}
\frac{ \TermAlpha_{2} }{ \Rmax^2 } \leq 2 \Exp \left [ 
\frac{[ g_{\cdeg}(\numobs(\action)) - 1]^{2}} {\numobs(\action)} 
\indicator \{0 < \numobs(\action) \leq \cdeg \} \right] + 
2 \Exp \left[\frac{1}{\numobs(\action)} \indicator \{\numobs(\action)>0\}\right].
\end{align*}
Note however the first term has been controlled in the analysis of
$\TermAlpha_{1}$: 
\begin{align*}
2 \Exp \left [ \frac{ [g_{\cdeg}(\numobs(\action)) - 1 ]^{2}}{\numobs(\action)} 
\indicator \{ 0 < \numobs(\action) \leq \cdeg \}\right] & \leq 2 \Exp \left [[g_{\cdeg}(\numobs(\action))-1]^{2} \indicator \{ 0 < \numobs(\action) \leq \cdeg \} \right ] \\
 & \leq 2 \Exp \left[ [g_{\cdeg}(\numobs(\action))-1]^{2} \indicator \{\numobs(\action)\leq \cdeg\}\right]\\
 & \leq 2 e^{ - \numobs \pibehave(\action)} + 
 \frac{1}{2}\plaincon_{0} \log \numaction \cdot \numaction^{4\plaincon_{0}}.
\end{align*}
Regarding the second term, Lemma 1 of the paper~\cite{li2015toward} tells us that\footnote{Though Lemma 1 of the paper~\cite{li2015toward} deals with the case when $\numobs(\action)$ is a binomial random variable, the same proof works for the case with Poisson random variables since the multiplicative Chernoff bound used therein also holds for the Poisson case.}  
\begin{align}
\Exp \left[ \frac{1}{ \numobs(\action) } \indicator \{ \numobs(\action) > 0 \} \right ]
\leq \min \left \{ 1, \frac{5}{\numobs \pibehave(\action) } \right \}. 
\end{align}
Combining the preceding bounds yields the stated
conclusion~\eqref{eq:variance-alpha-2}.


\section{Some auxiliary results}
\label{SecAuxiliary}

This section gathers some known auxiliary results that
are used in our analysis.

\subsection{Best polynomial approximation}

Given an interval $I \coloneqq [\myleft,\myright]$ with $\myleft > 0$,
a positive integer $\cdeg > 0$ and a continuous function $\phi$ on
$I$, let
\begin{align*}
\label{eq:defn-best-approx}  
\PolyApproxError_{\cdeg}(\phi;I) \coloneqq \inf_{\{ \coeff_{i} \}}
\sup_{x \in I} \left| \sum_{i=0}^{\cdeg} \coeff_{i} x^{i} - \phi(x) 
\right|
\end{align*}
denote the best uniform approximation error of $\phi$ on $I$ by
degree-$\cdeg$ polynomials.

In particular, for the function $\phi(x)=1/x$, the following
lemma, proved in Section 2.11.1 of the book~\cite{timan2014theory},
provides a precise characterization of $\PolyApproxError_{\cdeg}
(1/x;[\myleft,\myright])$.

\begin{lemma}
\label{lemma:error-1/x}
Fix any $\myright > \myleft >0$ and any positive integer
$\cdeg$. Denoting $\ratio \coloneqq \myleft / \myright$, we have
\begin{align*}
2 \PolyApproxError_{\cdeg} \left(\tfrac{1}{x}, [\myleft,\myright]
\right) = \frac{(1+\sqrt{\ratio})^{2}}{\myleft} \left(1 -
\frac{2\sqrt{\ratio}}{1 + \sqrt{\ratio}} \right)^{\cdeg + 1}.
\end{align*}
\end{lemma}

In fact, the problem of best polynomial approximation is closely
related to the problem of moment matching, as shown in the following
lemma (cf. Appendix E of the paper~\cite{wu2016minimax}).
\begin{lemma}
\label{lemma:duality}
The following identity holds:
\begin{align*}
2 \PolyApproxError_{\cdeg}(\phi;I) & = \max \quad \Exp_{\Xvar \sim 
\LocalDist_{1}}[ \phi(\Xvar) ] -
\Exp_{\Xvar \sim \LocalDist_{0}}[\phi(\Xvar)]\\ & \quad \mathrm{subject}
\,\text{\ensuremath{\mathrm{to}}}\quad\Exp_{\Xvar \sim
  \LocalDist_{1}}[\Xvar^{l}] = \Exp_{\Xvar \sim \LocalDist_{0}}[\Xvar^{l}],\qquad l=0,1,\ldots,\cdeg,
\end{align*}
where the maximum is taken over pairs of distributions
$\LocalDist_{0}, \LocalDist_{1}$ supported on the interval $I$.
\end{lemma}


\subsection{Minimax lower bound via Le Cam's method}

Here we state a version of Le Cam's method for lower bounds based on
mixture distributions.  Consider a class of distributions
$\{\Prob_{\LocalPara} \mid \LocalPara \in \LocalParaSpace \}$, and a
target function $\Target(\LocalPara)$ of the parameter $\LocalPara$.
Let $\LocalObs$ be a random vector drawn according to some
distribution $\Prob_{\LocalPara}$, and $\TargetHat(\LocalObs)$ be an
arbitrary estimator of the target $\Target(\LocalPara)$ based on the
data $\LocalObs$.

Let $\Prior{0}, \Prior{1}$ be two priors on the parameter space
$\LocalParaSpace$. Correspondingly, let $\LocalMarginal{i}$ denote the
marginal distribution of the observation $\LocalObs$ under the prior
$\Prior{i}$, for $i = 0, 1$.  We then have:
\begin{lemma}
\label{lemma:fuzzy-tsybakov-1}
Suppose that there exist some quantities $\xi \in \real, s > 0, 0 \leq
\beta_{0}, \beta_{1}<1$ such that
\begin{align*}
\Prior{0}(\LocalPara : \Target(\LocalPara) \leq \xi-s) & \geq 1 -
\beta_{0}; \\ \Prior{1}(\LocalPara : \Target(\LocalPara) \geq \xi+s) &
\geq 1 - \beta_{1}.
\end{align*}
If $\TVdist(\LocalMarginal{1},\LocalMarginal{0})\leq \plower < 1$,
then
\begin{align*}
\inf_{\TargetHat} \sup_{\LocalPara \in \LocalParaSpace} \Prob
_{\LocalPara} \left( | \TargetHat(\LocalObs) - \Target(\LocalPara) |
\geq s \right) \geq \frac{1- \plower -\beta_{0}-\beta_{1}}{2}.
\end{align*}
\end{lemma}


\subsection{Divergence between mixtures of Poisson distributions}

Given a nonnegative random variable $\Xvar$, denote by $\Exp[
  \Poi(\Xvar) ]$ the Poisson mixture with respect to the variable
$\Xvar$.  We then have the following bound, proved as Lemma 3 in the
paper~\cite{wu2016minimax}, on the TV distance between two such
Poisson mixtures.
\begin{lemma}
\label{lemma:Poisson-moments}
Let $\Xvar, \Xvar'$ be random variables supported on $[0,b]$ such that
$\Exp[\Xvar^{j}] = \Exp[(\Xvar')^{j}]$ for $j=1,2,\ldots, \cdeg$ for
some $\cdeg > 2 e b $.  Then the TV distance is bounded as
\begin{align}
\TVdist \left (\Exp[\Poi(\Xvar)],\Exp[\Poi(\Xvar')] \right) & \leq
\left( \frac{2 e b}{\cdeg} \right)^{\cdeg}.
\end{align}
\end{lemma}

\end{document}